\documentclass[10pt,twocolumn,letterpaper]{article}

\usepackage[pagenumbers]{wacv} 

\definecolor{wacvblue}{rgb}{0.21,0.49,0.74}
\usepackage[pagebackref,breaklinks,colorlinks,allcolors=wacvblue]{hyperref}
\usepackage{tabularx}
\newtheorem{definition}{Definition}
\usepackage{amsmath}
\usepackage{mathtools}
\DeclarePairedDelimiter\abs{\lvert}{\rvert}%
\newcommand{\appendixhead}
    {{\textbf{\Large ClimateVID - Social Media Videos Analysis and Challenges Involved}\vspace{0.25in}\par}}

\def\wacvPaperID{1879} 
\def\confName{WACV}
\def\confYear{2026}

\title{ClimateVID - Social Media Videos Analysis and Challenges Involved}

\author{Shiqi Xu\thanks{Equal Contributions}\\
University of Mannheim \\
\and
Moritz Burmester\footnotemark[1]\\
University of Mannheim \\
\and
Katharina Prasse\\
University of Mannheim\\
{\tt\small katharina.prasse@uni-mannheim.de}
\and
Isaac Bravo\\
Technical University of Munich\\
\and
Stefanie Walter\\
Technical University of Munich\\
\and
Margret Keuper\\
University of Mannheim, \\
Max-Planck-Institute for Informatics, Saarland Informatics Campus
}

\begin{document}
\maketitle
\begin{abstract}
The pervasive growth of digital content, specifically short videos on social media platforms, has significantly altered how topics are discussed and understood in public discourse. 
In this work, we advance automated visual theme detection by assessing zero-shot and clustering capabilities on social media data.
(1) We evaluated the capabilities of notable VLMs such as VideoChatGPT, PandaGPT, and VideoLLava using zero-shot image classification and compared their performance to the baseline provided by frame-wise CLIP image classification. 
(2) By treating clustering as a minimum cost multicut problem, we aim to uncover insightful patterns in an unsupervised manner.
For both analysis strategies, we provide extensive evaluations and practical guidance to practitioners.
While VLMs are currently not able to detect climate change specific classes, the clustering results are distinct visual frames.
We find that both ConvNeXt V2 and DINOv2 produce meaningful clusters, with DINOv2 focusing more on style differences and abstract categories, while ConvNeXt V2 clusters differ in more fine-grained ways.
Code available at \url{https://github.com/KathPra/ClimateVID.git}.
\end{abstract}
    
\section{Introduction}
\label{sec:intro}
The explosion of digital content on social media platforms has dramatically transformed the landscape of communication and information sharing. 
Understanding how a topic is portrayed in the media landscape is essential for visual media research and a pre-requisite for effectively shaping public understanding and engagement \cite{seelig_frame_2022}.
In the context of climate change, the identification of visual themes and patterns is an active field of research \cite{gardam2025, bravo2025viral, hayes2024transformative, hayes2025visual, gardam2025, blewett2025beyond, yan2025multimodal, zeng2024understanding}.
This includes both the content of climate visuals \cite{bravo2025viral, mooseder_social_2023}, but also the visual representation of a certain event \cite{o2023visual, mcgarry2025fire,ko2024experience}.
Especially during the evolution of climate change denialism and skepticism \cite{kahl2025climate,mendy2024counteracting, luke2024climate, nwokolo2025climate, gardam2025}, understanding the visual discourse is key in order to effectively communicate facts and shape public understanding.
Visual theme analysis allows researchers to trace how specific themes and perspectives circulate within public discourse \cite{wardekker_visual_2019,rebich-hespanha_dominant_2016, wozniak_who_2017, mooseder_social_2023}. 
Among the most dynamic types of content are short videos, which are widely used to convey everything from entertainment to political messaging and social activism.
Especially when researching global topics, automated analysis is essential for the generation of insights.
At the same time, advances in vision models open up new possibilities for automating video understanding at scale.

In this work, we evaluate VLM capabilities in zero-shot image classification for VideoChatGPT \cite{maaz2023video}, PandaGPT \cite{su2023pandagpt}, and VideoLLava \cite{lin2024video} and compare them to a baseline of frame-wise CLIP \cite{radford2021learning}. 
In our evaluations, we focus exclusively on open-source and open-weights methods.
Given large deficits in VLM on social media images, we recommend frame-by-frame analysis for social media video understanding with pre-defined categories.
We further discuss how to best handle meme content in social media videos.

Additionally, we follow-up on work by \citet{prasse_i_2025} and \citet{mooseder_social_2023} by evaluating clustering with the goal of discovering topics in an unsupervised manner.
Clustering video data instead of image data is a more challenging task given that videos have temporal redundancy, changing viewpoints, and multiple distinct scenes \cite{truong_video_2007,gygli_creating_2014}. 
This makes it necessary to decide not only which frames to extract, but also how to combine them into a single, representative video-level embedding for clustering. 
Suboptimal choices in these steps risk over-representing redundant content or overlooking short but important moments, which limits the ability to uncover meaningful visual themes.
We evaluate several frame selection strategies and find their effect to be benign.
Given that CLIP embeddings have returned coarse clusters \cite{prasse_i_2025}, we evaluate DinoV2 \cite{oquab2023dinov2} and ConvNeXt V2 \cite{woo2023convnext} embedding spaces for this task.
Based on our evaluations, we recommend using ConvNeXt V2 embeddings for video clustering.
Phrasing the clustering task as a minimum cost multicut problem yields clusters of visual themes, confirming previous findings \cite{prasse_i_2025, andres_probabilistic_2011}.
The identification of non-dominant visual themes opens the door for in-depth social science research such as  \cite{blewett2025beyond}.

\noindent Our main contributions are as follows:
\begin{itemize}
    \item Evaluation of VLMs (VideoChatGPT, PandaGPT, and VideoLLava), using zero-shot image classification in the context of social media videos. 
    \item Extension of minimum cost multicut clustering to social media videos including methodological ablations for video processing.
    \item Practical recommendations for social media video analysis with the goal of assisting researchers and practitioners in the field.

\end{itemize}

\section{Data}
\label{sec:dataset}
We analysed Twitter videos in line with X's terms of services, developer agreement, developer policy, and privacy agreement.
For both methods assessed, we have created a representative subset of all available videos to generate insights in a resource-efficient manner.

\subsection{Collection}
\label{ssec:coll}
The videos were collected using the X Academic API in 2023, before the closure of the API for free academic access.
The videos were selected based on the presence of either \textit{climatechange}, \textit{climate change}, or \textit{\#climatechange} in the tweet text. 
Tweets without videos are not part of this study.
We have only included videos for which we have a full year of content available in order to allow year-wise comparison.

\subsection{Licence}
\label{ssec:lic}
According to X's terms of services, the users retain all rights to their content submitted to X.
By uploading content, users grant X a worldwide, non-exclusive, royalty-free license to share their data. 
Researchers are allowed to access X data through the API and conduct analysis while respecting the users' reasonable expectation of privacy and refraining from investigating sensitive user information.
Given that individuals' perception of climate change may be related to their political affiliations and beliefs, we conduct our analysis exclusively on the videos and disregard who shared the video.
Moreover, we report results either as a yearly aggregate or on the entire dataset.

To make our study transparent, we share the list of tweet ids used for each task.
Within this list, we indicate the year of posting and mark duplicate videos.
Sharing X post ids complies with user post deletions and content removal.

\subsection{Duplicate Detection}
\label{ssec: dups}
The prevalence of reposts and direct re-uploads on social media platforms frequently results in numerous identical video instances, making deduplication necessary.
We calculated hash values for the first and last non-black frame of each video and evaluated duplicates based thereon.
The first video encountered was designated as the ``original", and subsequent videos were labeled as ``duplicates".
Hashing was chosen primarily for its computational efficiency. 
Out of 129,955 videos, 51,505 videos were identified as duplicates (39.6\%) and 78,450 unique videos (60.4\%).

\subsection{Classification subset}
We assessed the zero-shot classification efficacy on 44,927 videos out of 78,450 unique videos collected ($57\%$).  
We randomly selected 12,000 videos per year (2019 - 2021), while in 2022 only 9,399 unique videos were available, and thus all were selected. 
Out of these 45,399 videos, 472 videos ($0.6\%$) could not be processed by at least one model and were thus removed from the sample.

\subsection{Clustering subset}
Clustering performance was evaluated on 2,975 videos ($4\%$).
Under the assumption that longer videos are more difficult to analyse, we removed video of length $< 5 s$.
We sample videos from each month proportionally to the number of unique videos.
\section{Methods}
\label{sec:methods}
Our purpose is to critically assess methods for social media video analysis to guide future research on method selection.
We evaluate several methods for zero-shot classification using foundation models \cite{radford2021learning, lin2024video, su2023pandagpt, maaz2023video}; we exclusively considered open-source and open-weights models.
Additionally, we employ Minimum-Cost Multicut clustering which is highly suitable for social media image clustering \cite{prasse_i_2025}.
Building on \citet{prasse_i_2025}, we evaluate the two most recommended embedding spaces, \ie DinoV2 \cite{oquab2023dinov2} and ConvNeXt v2 \cite{woo2023convnext}.

\subsection{Classification}
\label{ssec:class}
We evaluated the capacity of Video-LLaVA \cite{lin2024video}, Video-ChatGPT \cite{maaz2023video}, PandaGPT \cite{su2023pandagpt} and CLIP \cite{radford2021learning} to classify social media video content. 
Our comparison of three video frontier models and an image model is intended to highlight differences between the models.

\textbf{Video-LLaVA} \cite{lin2024video} consists of LanguageBind's \cite{zhu2023languagebind} visual encoders that map images and videos into a shared language-centric feature space.
The vision encoder is initialized as OpenCLIP-ViT-L\/14 \cite{ilharco2021openclip}. 
A 2-layer fully connected network with GeLU transforms the output into the LLM inputs, \ie Vicuna-7B v1.5 \cite{chiang2023vicuna}.
The model is trained in two stages: (1) ``understanding pretraining,” uses 558k image-text pairs (from LAION-CC-SBU \cite{sharma2018conceptual}) and 702k video-text pairs (from Valley \cite{luo2023valley} and WebVid \cite{bain2021frozen}). 
In this stage, only the shared projection after the vision encoder and before the LLM is trained.
(2) ``instruction tuning,” employs a 665k image-text instruction dataset (from LLaVA 1.5 \cite{liu2024improved}) and a 100k video-text instruction dataset (from Video-ChatGPT \cite{maaz2023video}). 
Besides the shared projection, the LLM is now fine-tuned.
This entire training is done using an auto-regressive loss.

\textbf{Video-ChatGPT} \cite{maaz2023video} is designed for detailed video understanding and dialogue capabilities.
In contrast to Video-LLava, it is fine-tuned exclusively on video-caption pairs, which aids temporal understanding.
Its architecture uses a pre-trained visual component, CLIP ViT-L/14 \cite{radford2021learning} taken from LLaVa \cite{liu2023visual}.
Spatiotemporal relations are captured by average-pooling the individual frame representations, once over the temporal dimension and once over the spatial dimension.
The video representation is sent through a linear projection layer to the LLM Vicuna-7B v1.1 \cite{chiang2023vicuna}. 
The projection layer is fine-tuned while the visual encoder and the LLM remain frozen.
The model is instruction-tuned on a dataset comprising 100,000 video-instruction pairs. 
Training proceeds with an autoregressive objective.

\textbf{PandaGPT} \cite{su2023pandagpt} is a multimodal instruction-following model that combines ImageBind \cite{girdhar2023imagebind} with the Vicuna-13B v.0 \cite{chiang2023vicuna} LLM, enabling understanding and reasoning across modalities.
The two components are connected using a trainable linear projection and LoRA \cite{hu2022lora} based adaptator following the LLM. 
PandaGPT is trained by using 160k image-text pairs released by \citet{liu2023visual} and \citet{zhu2023minigpt} in a multi-turn dialogue format as proposed by \cite{liu2023visual, zhu2023minigpt}.
During training, the projection and LoRA weights are updated, while the encoder and LLM remain frozen. 

\textbf{CLIP} \cite{radford2021learning} is a vision-language aligned model pre-trained on 400 million image–text pairs. 
The two modalities are aligned using a contrastive loss, which moves matching data points  closer while moving non-matching ones  apart.

To classify each selected video, a total of 59 distinct categories are defined in accordance with the relevant climate change literature \cite{bravo2025viral}, which have been shown to be relevant by previous case studies \cite{prasse2023towards, mooseder_social_2023}.
These categories are divided into five main groups: \textit{animals}, \textit{consequences}, \textit{climate action}, \textit{setting} and \textit{type}. 
A detailed overview of all categories in their respective group is provided in \cref{tab:categories}. 
We refrain from calling the categories classes, as they do not align with traditional machine learning classes.
However, given the use case, we agree with previous work that social media analysis requires task-dependent categories.

\begin{table}
  \centering\caption{Categories in their groups.}
   \footnotesize
    \begin{tabular}{p{0.25\columnwidth}p{0.65\columnwidth}}   
    \toprule
    \textbf{Group} & \textbf{Categories} \\
    \midrule
    Animals           & Pets, Farm Animals, Polar Bears, Land Mammals, Sea Mammals, Fish, Amphibians, Reptiles, Invertebrates, Birds, Insects, Other Animals \vspace{0.2cm}\\ 
    Consequences      & Floods, Drought, Wildfires, Rising temperature, Other Extreme Weather Events, Melting Ice, Sea Level Rise, Human Rights, Economic Consequences, Biodiversity Loss, Covid, Health, Other Consequence  \vspace{0.2cm}\\ 
    Climate Action   & Politics, Protests, Solar Energy, Wind Energy, Hydropower, Bioenergy, Coal, Oil, Natural Gas, Other Climate Action     \vspace{0.2cm}\\ 
    Setting           & No Setting, Residential Area, Industrial Area, Commercial Area, Agricultural, Rural, Indoor Space, Arctic or Antarctica, Ocean, Coastal, Desert, Forest or Jungle, Other Nature, Outer Space, Other Setting  \vspace{0.2cm}\\ 
    Type              & Event Invitations, Meme, Infographic, Data Visualization, Illustration, Screenshot, Single Photo, Photo Collage, Other Type \\     
    \bottomrule
  \end{tabular}
\label{tab:categories}
\end{table}

While Video-LLaVA, PandaGPT, and Video-ChatGPT process the entire video in combination with a prompt and generate text-based responses, CLIP analyzes individual frames and returns probability distributions across the predefined categories. 
It is important to note that PandaGPT is only a research prototype and a not fully trained version was used. 
Given the use case, multi-label outputs were accepted.
We evaluate the adherence of the model to the required output format by comparing the returned category against the output text to the categories directly.
The word matching first removes terms which are negated, \ie ``This video contains \underline{no} pets, but farm animals" returns \textit{farm animals}.
All prompts can be found in \cref{app:promts}.

Unlike VLMs, CLIP requires a list of objects as input and returns probability scores indicating the likelihood of each object appearing in a frame. 
Any category exceeding a 50\% probability threshold is included in the final classification. 
If no category can be found with a likelihood of at least 50\%, the highest-scoring category will be chosen for the \textit{setting} and \textit{type} groups, while \textit{No} will be selected for the \textit{animals}, \textit{consequence} and \textit{climate action} groups.
Since CLIP processes individual frames rather than whole videos, each video was uniformly sampled into eleven frames. 
For videos with fewer than eleven frames, all available frames were utilized. 
The classification results are aggregated frame predictions.

We evaluate zero-shot classification by sanity-checking the results on semantic coherence given the climate change literature.
To this end, we compare the returned class labels per model and,  if this check is successful, between models.
Moreover, we validate the results by manually inspecting the images corresponding to the given labels.
Our evaluation has the goal of enabling low-resource, large-scale research into climate change videos on social media.


\subsection{Model Prompting}
For all VLMs, we have used the same prompts per category; we provide the prompts in \cref{app:promts}.
To optimize computational efficiency, a hierarchical prompting strategy was used. 
The system first conducts an initial screening using three simple binary, \textit{yes} or \textit{no}, prompts to detect the presence of the category groups \textit{animals}, \textit{consequences} and \textit{climate action} in a video. 
Each prompt is presented without shared context, allowing the model to evaluate them independently. 
If a positive response is received for any of the category groups, the system proceeds to a second batch of more detailed classification prompts. 
This phase includes prompts for the relevant category-specific labels as well as the universal \textit{setting} and \textit{type} prompts, which are needed for every video. 
The model is prompted to return the number corresponding to the category.
Both prompt types begin with a clear task instruction for the model. The phrase ``Analyze the video."  ensures that the model connects the prompt and the category list with the video content. 

\subsection{Clustering}
\label{ssec:clust}

We investigate clustering phrased as a Minimum Cost Multicut Problem [MP] for videos, given its effectiveness for social media images \cite{prasse_i_2025}.
This clustering technique has been shown as effective for several vision tasks such as content clustering \cite{prasse_i_2025,ho2020learning,ho2021msm}, segmentation \cite{andres2011probabilistic, jung2022optimizing, Galasso_2014_CVPR,jung2022learning, kardoost2019solving}, and multi-person tracking \cite{tang2016multi, tang2017multiple, ho2020unsupervised, ho2020two, nguyen2022lmgp}.
Further uses are efficient image and mesh graph decomposition \cite{keuper2015efficient} and motion segmentation~\cite{KB15b, K17,KB19,kardoost21,kardoost22}.

\begin{definition}
A finite, undirected graph G = (V,E) with cost $w: E  \rightarrow \mathbb{R} $ assigned to the edges is cut into detached components such that edges with minimal cost are cut
\begin{equation}
    \min_{\textbf{y} \in \{0,1\}^{\abs{E}}} \hspace{1em} c(\textbf{y}) = \textbf{y}^T\textbf{w} = \sum_{e \in E} w_e y_e ,
\end{equation}
where $y$ is the binary edge label indicating whether the edge should be cut. This is subject to the linear constraints
\begin{equation}
    \forall C \in \text{cycles}(G), \forall e \in C: (1-y_e) \leq \sum_{e' \in C \setminus \{e\}} (1-y_{e'}) .
\end{equation}
\end{definition}

The MP can be viewed as a Bayesian network \cite{andres_probabilistic_2011, keuper2015efficient, prasse_i_2025}, where the optimal solution is maximizing the posterior probability, and thus is the optimal solution to the clustering task.
This method has a single hyperparameter $cal$, the calibration term, which transforms the edge weights and thus determines the likelihood of cutting an edge.
When two data points stop being similar highly depends on the embedding space and we thus ablate $cal$.
Due to time and resource constraints, strong solvers have been developed which provide a near optimal solution with a much lower budget \cite{graph_mcmc}.
In this work we use greedy agglomerative edge contraction first, which we then improve further using the Kerninghan-Lin Algorithm with joins \cite{graph_mcmc}.

Our clustering pipeline consists of four steps: frame extraction, embedding generation, frame embedding combination, and clustering.
Firstly, the frame extraction step selects representative frames using either uniform static sampling or visual diversity-based sampling. These frames are passed to a pre-trained foundation model—DINOv2 \cite{oquab2023dinov2} or ConvNeXt V2 \cite{woo2023convnext}—which encodes them into embeddings that capture visual themes.
Then, the frame embeddings are aggregated into a single representation per video using combination method like simple average or weighted average. A similarity graph is constructed by computing pairwise cosine similarities between video embeddings using \citet{prasse_i_2025}'s implementation. This graph is then partitioned using the Minimum Cost Multicut algorithm \cite{andres_probabilistic_2011} to identify clusters of visually similar content. 

\textbf{ConvNeXt V2 \cite{woo2023convnext}} is a fully convolutional masked autoencoder architecture.
This entails the random masking out of 60\% of the training image's pixels and predicting their content.
The encoder uses sparse convolutions during pre-training and standard convolutions during fine-tuning.
The decoder is a standard ConvNeXt block.
The self-supervised training is done using ImageNet-1K for 800 pre-training and 100 fine-tuning epochs.
Additionally, global response normalization is used (GRN).
Mean square differences between the reconstructed and the target imaged is used as a training signal for the masked out image regions.

\textbf{DinoV2} \cite{oquab2023dinov2} is a ViT architecture which is trained using a student teacher setting.
The training objectives are both on image-level and patch-level, i.e. the cross-entropy loss between image features from the student and the teacher network, and between masked student patches and unmasked teacher patches respectively.
Untyping the weights for both objectives further improves performance.
A highly curated dataset is used for pre-training with normalization and the embedding space is regularized by the entropy-based KoLeo estimator.
The training is self-supervised and the image resolution is increased during the end of pretraining. 

Our evaluation includes both qualitative and quantitative checks.
This includes well-established clustering metrics: silhouette score \cite{rousseeuw_silhouettes_1987}, Davies-Bouldin \cite{davies_cluster_1979}, and Calinski-Harabasz \cite{calinski_dendrite_1974} since we strive for cluster cohesion, separation, and density. 
Additionally, we evaluate usefulness for further analysis via cluster balance (Gini coefficient), coverage efficiency (\% of videos in the top-10 largest clusters), and over-fragmentation (singleton ratio).
When comparing 2 clusterings, we report Chi-square to assess temporal independence of clusters (single event or long-term topic) and pairwise cosine similarities between cluster centroids.
Lastly, we sample up to 10 videos randomly from each of the 10 largest clusters and extract 4 frames per video to capture temporal variation within videos.
More evaluation details can be found in \cref{app:clusteval}



\subsection{Frame Selection}
Selecting representative video frames  has been extensively discussed in the video summarization literature \cite{gygli_creating_2014,zhang_video_2016,huynh-lam_cluster-based_2024,de_avila_vsumm_2011,mundur_keyframe-based_2006}. 
To simplify video analysis, key-frame extraction methods reduce the problem to an image-based one by selecting representative frames from each video \cite{janwe_video_2016}. 
They usually rely on detecting visual changes \cite{ejaz_efficient_2013}, or by grouping frames through clustering based on low-level visual features\cite{gygli_creating_2014,huynh-lam_cluster-based_2024,zhuang_adaptive_1998}.  
Early work by \citet{zhuang_adaptive_1998} used k-means clustering on color histogram features to group visually similar frames within video shots. 
\citet{mundur_keyframe-based_2006} advanced this approach by introducing Delaunay Triangulation for clustering with color feature space. 
\citet{janwe_video_2016} proposed a clustering-based key-frame extraction method combining HSV color histograms and GLCM texture features, followed by mutual similarity filtering to remove near-duplicates. 

The clustering-based key-frame extraction typically follows a common approach: extracting visual features from frames, applying clustering algorithms to group similar frames, and selecting cluster centers as representative key frames \cite{yang_key_2005}. 
While these approaches effectively identify representative frames, they primarily focus on low-level visual features (color, texture) and may miss semantically important content. 
Furthermore, most existing work evaluates frame selection in the context of video summarization rather than semantic clustering. This work addresses this gap by comparing frame selection strategies specifically for visual theme discovery using embeddings from foundation models.
\citet{lee_large_2020} show that averaging visual features from sparsely sampled frames (e.g., at 1 fps) results in high-quality video-level embeddings suitable for large-scale video understanding tasks, such as retrieval and classification.
Their results show that simple aggregation over pretrained CNN features can be highly effective.
While temporal modeling approaches capture motion dynamics, they may be unnecessarily complex for discovering recurring visual themes in climate communication, where static visual elements often carry the core message \cite{schafer_frame_2017,mooseder_social_2023}.

\textbf{Static selection} uniformly samples frames at evenly spaced intervals and thus captures the temporal progression of videos.
We fix the frame sampling rate at 1 fps, sample at least 4 frames  ($n_{min}$), and not more than 100 ($n_{max}$) per video.
For a video of length ($t_{\text{video}}$), we sample 
\begin{equation}
n = \mathrm{min}(\mathrm{max}(t_{\text{video}} \times \text{fps}, n_{\text{min}}), n_{\text{max}})
\end{equation}

Thus we sample a varying number $n$ of frames per video 
\begin{equation}
\text{indices}_i = \left\lfloor i \cdot \frac{T - 1}{n - 1} \right\rfloor, \quad i \in \{0, 1, ..., n-1\}
\end{equation}
where $T$ is the total number of frames. This produces $n$ evenly spaced indices from 0 to $T-1$, ensuring that each temporal segment of the video is equally represented.

\textbf{Dynamic selection} identifies and samples distinct frames within each video. 
This approach addresses scenarios where important visual content may be concentrated in brief segments—such as a sign appearing for only a few seconds in a longer video, which static selection might miss. 
The implementation in this work builds upon clustering-based keyframe extraction methods established in prior work \citet{liu_combined_2005,janwe_video_2016,yeung_efficient_1995,zhuang_adaptive_1998}.

\textbf{Frame validation} excludes color-uniform images as proposed by \citet{n_image_2016}.
We remove frames which have grayscale standard deviation $< 10$, frames with mean brightness $<5$ or $>250$, and corrupted frames.
When a frame fails validation, we search for the nearest valid frame within a $\pm$5-frame window. 
If no valid frame exists within the window, the position is skipped.

\textbf{Video representation} are created by transforming the frame representations into a single representation.
An overview of the evaluated combination strategies can be found in \cref{tab:method_props} while details are in \cref{app:frameselection}.
\begin{table}[ht]
\centering
\caption{Comparison of frame combination methods}
\footnotesize
\begin{tabular}{lccc}
\toprule
\textbf{Method} & \textbf{Models} & \textbf{Complexity}  \\
\midrule
Simple Average & Both & $O(N)$  \\
Max Confidence & ConvNeXt V2 & $O(N)$  \\
Weighted Diversity & Both & $O(N^2)$ \\
Weighted Confidence & ConvNeXt V2 & $O(N)$ \\
Temporal Coherence & DINOv2 & $O(Nr)$  \\
\bottomrule
\end{tabular}
\label{tab:method_props}
\end{table}
\section{Analysis: Zero-shot Classification}
\label{sec:classification}

\subsection{Prompt-Engineering \& Formatting}
Without this explicit instruction ``Analyze the video.", models appeared to select categories at random, without demonstrating a reasoning process behind it.
The simpler yes/no format was reliably accepted by all models. 
However, the more detailed format, which required the model to output a category number instead of the category itself was handled inconsistently. 
Video-LLaVA \cite{lin2024video} demonstrated the highest success rate, while Video-ChatGPT \cite{maaz2023video} and PandaGPT \cite{su2023pandagpt} encountered significant formatting challenges.
This finding is consistent across groups for Video-LLaVA and PandaGPT, \ie they differ at most by 3.5\%.
Video-ChatGPT differs more greatly: for \textit{animals} and \textit{consequences} the formatting works quite well, while for the remaining groups, the model outputs diverge more. 
For at least 51.5\% of the videos, the answer can also be parsed using word search.
Simultaneously this indicates that the parsing of the numeric output might be erroneous, as it appears to be a number in a longer text.

\subsection{Results}
Zero-shot classification results are visualized in bar charts showing the relative frequency of a given class on a quarterly basis.
If necessary, a table illustrating the normalized distribution of categories within the analyzed category group is generated for pattern recognition.
Following the method investigation, we inspect the video content over time while re-introducing duplicates using line graphs.
To this end, we assign the same label to the duplicates as the original video has.
We present the main findings and insights in the main paper, all results can be found in \cref{app:ZS}.

\textbf{Video-LLaVA}: The results from Video-LLaVA across all category groups are excluded due to a strong bias for the first category in the multi-option prompt.
The binary prompt (yes/no) does not share this bias, as for all two-stage category groups \textit{no}, \ie the second option, is the most frequent answer.
Further evaluation by changing the order of categories in the prompts both confirms the first label bias and undermines the faith in the binary result.
When comparing the Jaccard similarity of two independent model responses in the binary case, the overlaps are relatively low, \ie 75.81\% for \textit{animals}, 49.89\% for \textit{consequences}, and 63.78\% for \textit{climate action}.
We thus do not further analyse Video-LLaVA here.
See \cref{app:videollava} for more details.
  
\textbf{Video-ChatGPT}: Every category group except for \textit{Type} is getting excluded due to a high frequency of implausible and repetitive label combinations, leading to deficient faithfulness. 
More details can be found in \cref{app:videochatgpt}

\textbf{PandaGPT}: The \textit{animal} and \textit{consequence} groups are excluded due to the model's tendency to classify nearly every video with almost all categories. 
The \textit{climate action} group is excluded due to a bias towards the first category mentioned in the prompt, as well as the more general \textit{other climate action} category. 
The \textit{setting} group is excluded due to a high occurrence of conflicting categories.
The predictions for the \textit{type} categories pass the initial sanity check and are assessed in more detail.
See \cref{app:pandagpt} for more details.

\textbf{CLIP}: The model's results passed the sanity check for all groups and were thus further evaluated. We focused primarily on the distribution of classified videos within each category group and their trends over time. When significant spikes or unexpected distributions were observed, particularly in categories that seemed uncharacteristic, a targeted manual inspection of the underlying video content was conducted to identify potential causes. This methodical approach allowed for a broad evaluation of the trend results and an assessment of the model's classification accuracy. 
This was done by grouping videos within the quarter by their original video and then counting the appearances of the most prominent recurring video. 
This specific video was then manually inspected to verify the assigned category. 
More details can be found in \cref{app:clip}.

\section{Analysis: Clustering}
We aim to detect patterns in climate change videos in order to better understand this visual discourse.
To this end, we discuss frame selection and combination strategies, and embedding space differences.
Moreover, we discuss the efficacy of metrics for determining the optimal setup.
We ablate the clustering calibration term $cal$ in \cref{app:calterm}.

\subsection{Frame Selection and Combination}
We find that frame selection methods do not substantially affect the clustering results. 
For both embedding spaces, there is no clear trend when comparing static to dynamic frame selection across combination methods.
All DINOv2 frame combination strategies perform similarly.
For ConvNeXt V2, the same trend can be observed except for single frame selection, which underperforms in comparison to the other strategies.
This finding follows our intuition, as videos with more than one theme are only partially represented.

\subsection{Visual Theme Analysis}
\begin{figure*}[ht]
        \centering
        \begin{subfigure}[b]{0.31\textwidth}
        \includegraphics[width=\textwidth, trim={0 0 0 1.5cm},clip]{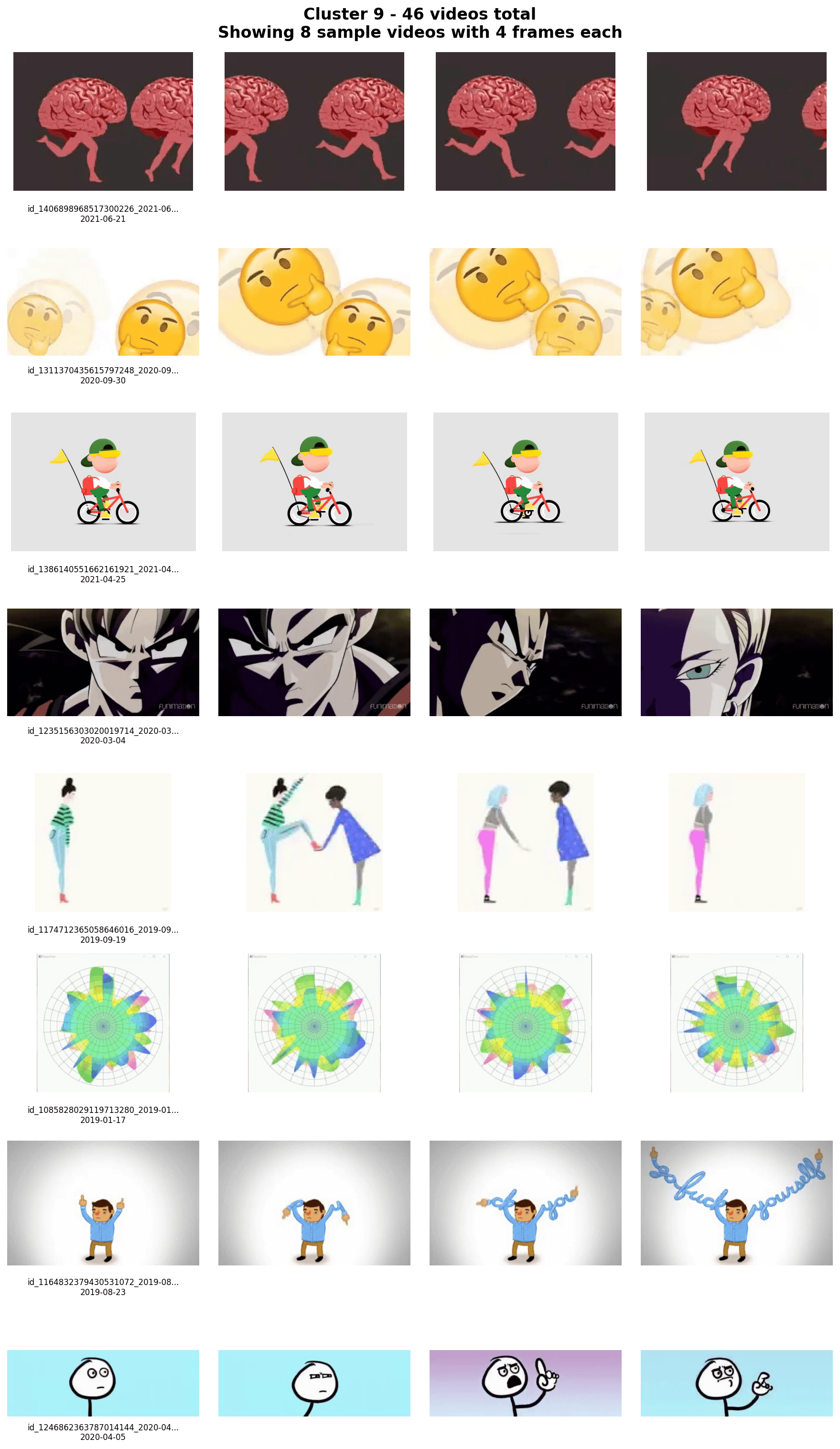}
        \caption{Cluster 9 (49 videos)}
        \end{subfigure}
        \hspace{0.3cm}
        \begin{subfigure}[b]{0.31\textwidth}
        \includegraphics[width=\textwidth,trim={0 0 0 1.5cm},clip]{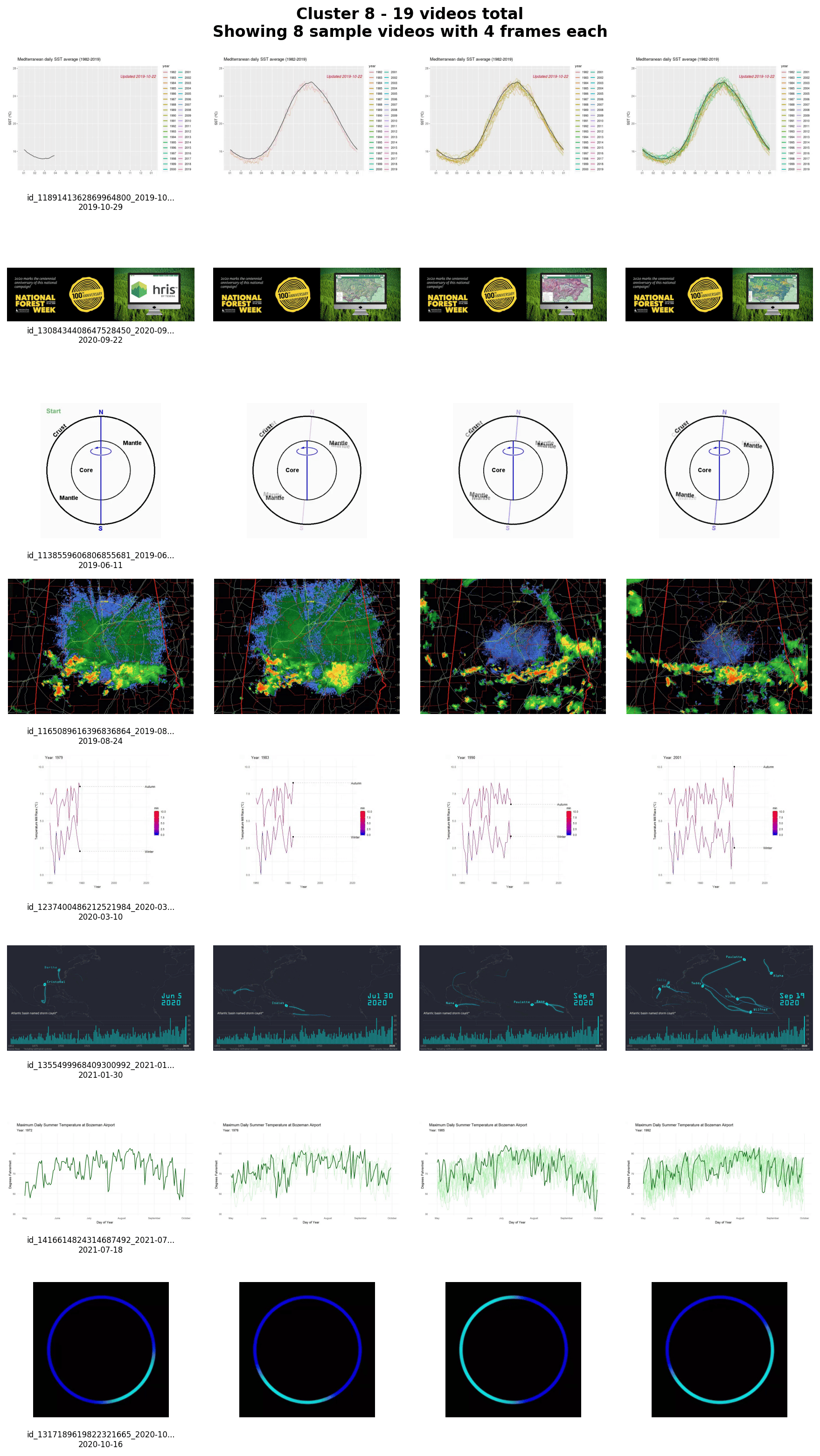}
         \caption{Cluster 8 (19 videos)}
        \end{subfigure}
        \hspace{0.3cm}
        \begin{subfigure}[b]{0.31\textwidth}
        \includegraphics[width=\textwidth,trim={0 0 0 1.5cm},clip]{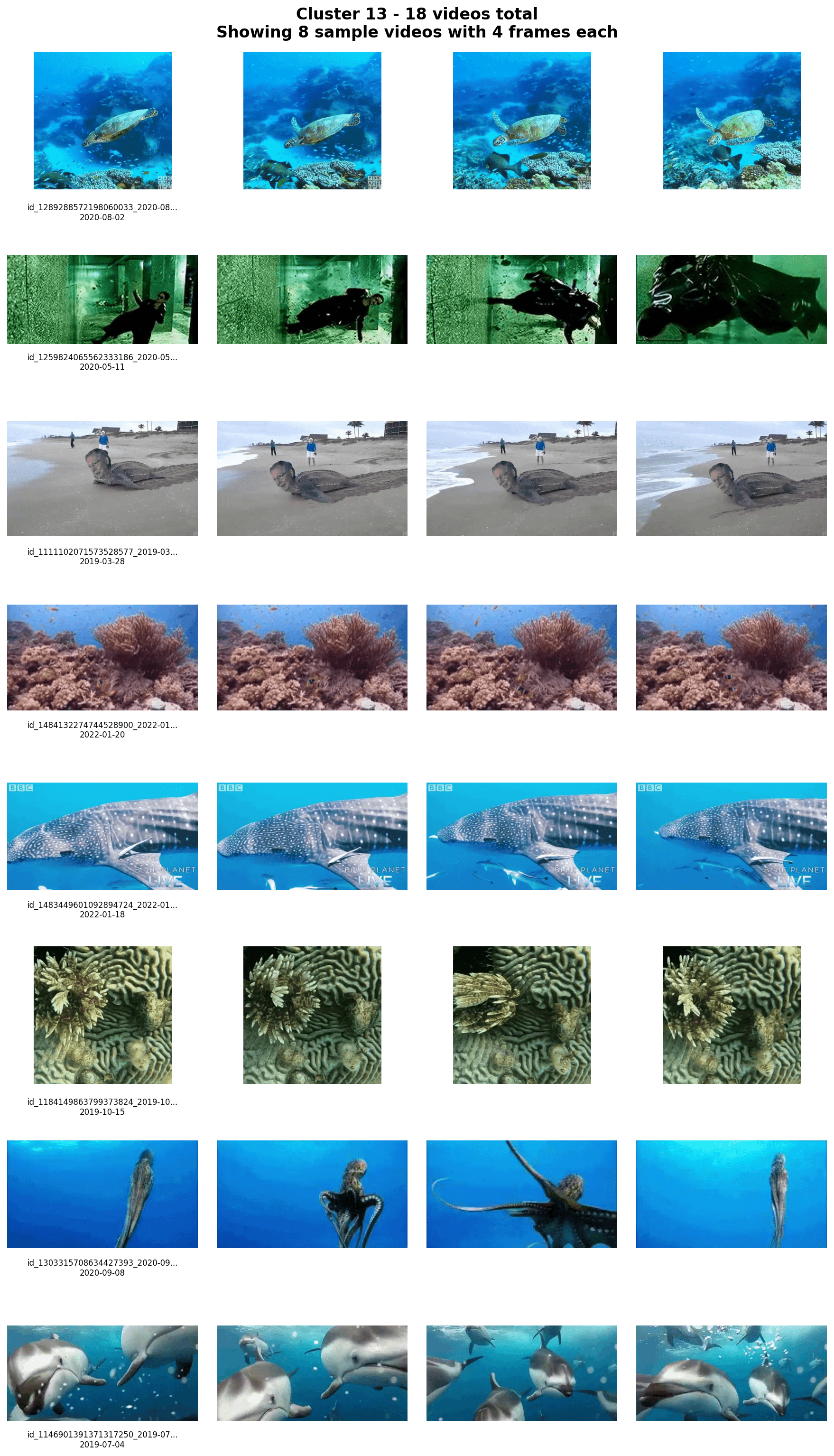}
         \caption{Cluster 13 (18 videos)}
        \end{subfigure}
        \caption{ConvNeXt~V2 clusters contain fine-grained differentiation of visual themes.}
\label{fig:convNext_highlights}
\end{figure*}

Clustering using both embedding spaces returned meaningful clusters for further analysis.
As shown in \cref{fig:convNext_highlights}, both embedding spaces find common trends, with ConvNeXt V2 clusters being more granular compared to DINOv2.

\begin{table}[ht]
    \centering
    \footnotesize
    \caption{Top 10 largest clusters, sorted by cluster size. Semantic similarities are highlighted in colour.}
    \begin{tabular}{m{13em} m{13em}}
   
    \toprule
    DinoV2 \cite{oquab2023dinov2} & ConvNeXt V2 \cite{woo2023convnext}\\
    \midrule
    {\color{orange} Media and digital communication} & {\color{orange} Multi-panel news / documentary} \\
    \vspace{0.2em}
    {\color{orange} Information graphics and educational content} & {\color{orange} Entertainment media / pop culture}\\
    \vspace{0.2em}
    {\color{violet} Space / cosmic perspective} & Natural landscapes / phenomena\\
    \vspace{0.2em}
    {\color{teal} Wildlife and animals} & {\color{violet} Space / cosmic perspective}\\
    \vspace{0.2em}
    Disappearing arctic / sea level rise  & {\color{orange} Digital animation / graphics}\\
    \vspace{0.2em}
    Solutions and sustainability& {\color{purple}Human activity / urban life}\\ 
    \vspace{0.2em}
    {\color{blue}Marine ecosystem} & {\color{teal} Animal content}\\
    \vspace{0.2em}    
    {\color{purple} Urban / economic impact} & {\color{teal} Rural environment / wildlife} \\ 
    \vspace{0.2em}
    Climate emergency / disaster & {\color{orange} Scientific data visualization} \\ 
    \vspace{0.2em}    
    Artwork & {\color{blue}Marine ecosystem}\\
\bottomrule
    \end{tabular}
    \label{tab:placeholder}
\end{table}

\textbf{DINOv2} organizes videos into high-level thematic groups, guided by broad visual cues such as color palette, compositional structure, texture, and subject positioning.
This results in high intra-cluster visual coherence, often with clear climate-related context.

\textbf{ConvNeXt V2} tends to form clusters around specific objects, contexts, media formats, or production styles. 
The climate relevance of its largest clusters is less pronounced, and cluster content often mix professional and amateur-quality footage.
ConvNeXt V2 seems to encode format, object and contextual detail more distinctly, resulting in higher fragmentation. For example, videos with animals are grouped into different clusters. Cluster 19 captures \textit{domestic animals} like pets and companion animals, while cluster 21 tends to capture \textit{wildlife}, showing animals in natural habitats. 
Remarkably, upon inspecting smaller clusters (Figure \ref{fig:convnextv2animalandhuman}), ConvNeXt V2 also put \textit{cats} and \textit{monkeys} into two distinct clusters. 
Also, human-activities are segmented into different clusters. 
One cluster contains \textit{human in pop culture and entertainment media}, another captures \textit{human activities in nature}, besides \textit{human in urban space}, and \textit{captures human in rural areas}. 
Furthermore, smaller clusters captures \textit{human in hunger crisis and humanitarian context}, and \textit{human with vehicles}. 
Visualizations of aforementioned clusters can be found in \cref{app:clusterquali}.

\textbf{Embedding space comparison} shows that the choice of embedding model largely influenced clustering granularity. 
ConvNeXt V2 forms more clusters (71 vs. 50 in DINOv2). Table~\ref{tab:embedding_cluster_dist} presents the detailed cluster size distribution. Both models have highly imbalanced cluster distributions, though with notable differences. DINOv2's top 5 clusters account for 94.8\% of all videos, leaving only 5.2\% distributed among the remaining 45 clusters. ConvNeXt V2 shows more balanced distribution, with 13.1\% of videos remaining in clusters beyond the top 5, distributed across 66 additional clusters.

\begin{table}[htbp]
\centering
\caption{Cluster size distribution by embedding model}
\label{tab:embedding_cluster_dist}
\footnotesize
\begin{tabular}{lcc}
\hline
\textbf{Cluster Size Category} & \textbf{DINOv2} & \textbf{ConvNeXt V2} \\
\hline
Singletons (1 video) & 17 (34.0\%) & 10 (14.1\%) \\
Small (2-9 videos) & 24 (48.0\%) & 47 (66.2\%)  \\
Medium (10-49 videos) & 6 (12.0\%) & 10 (14.1\%) \\
Large (50-99 videos) & 0 & 1 (1.4\%) \\
Very Large (100+ videos) & 3 (6.0\%) & 3 (4.2\%) \\
\hline
Total Clusters & 50 & 71 \\
\hline
\end{tabular}
\end{table}

Notably, ConvNeXt V2 has less singleton clusters (14.1\% vs. 34.0\%). However, it showed increased fragmentation in the small cluster range (2-9 videos), which comprised 66.2\% of all ConvNeXt V2 clusters compared to 48.0\% for DINOv2. 

\begin{figure}[ht]
    \centering
    \includegraphics[trim={20cm 0 23cm 0},clip, width=8cm]{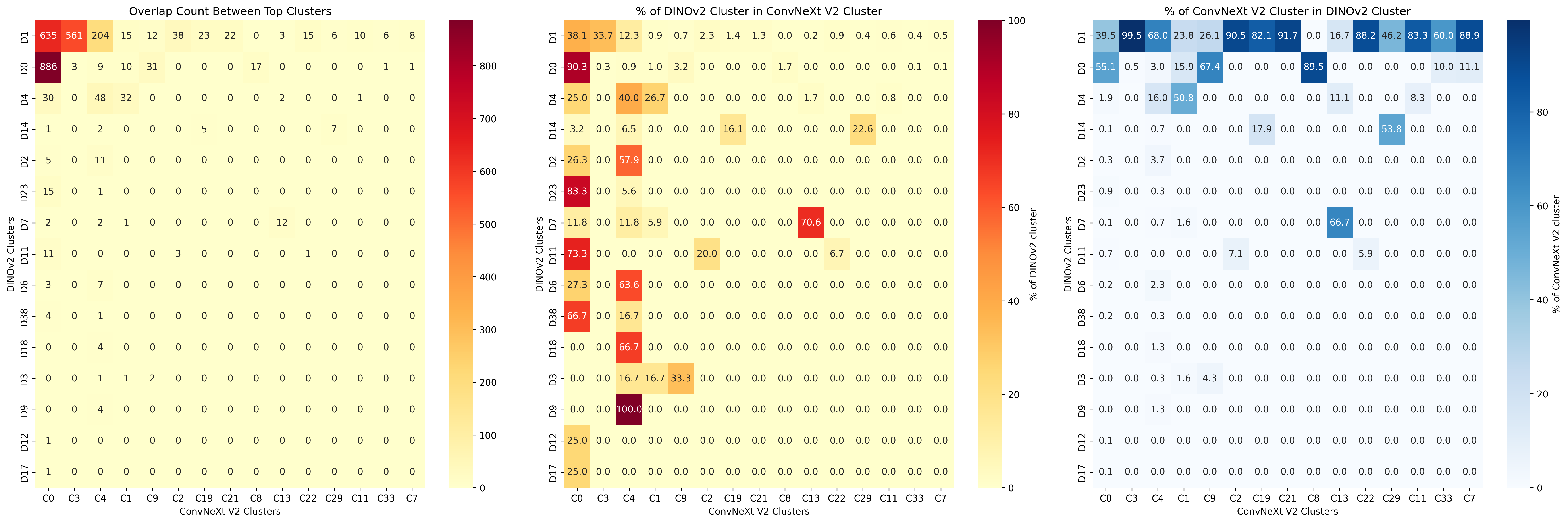}
    \caption{Top 15 Clusters overlap heatmap between DINOv2 and ConvNeXt~V2. 
    Darker cells indicate higher overlap percentages.}
    \label{fig:overlap_heatmap1}
\end{figure}

The overlap patterns in \cref{fig:overlap_heatmap1} show strong convergence on major visual categories. Most notably, 90.3\% of DINOv2's cluster~0 maps to ConvNeXt~V2's cluster~0, meaning that both models consistently identify this category (media and digital content) as a distinct visual theme. 
Furthermore, DINOv2's massive cluster~1 fragments across multiple ConvNeXt~V2 clusters, with 38.1\% mapping to cluster~0, 33.7\% to cluster~3, and 12.3\% to cluster~4, among others. 
More details can be found in \cref{app:clusteroverlap}

Overall, ConvNeXt~V2 seems to distinguish several subcategories within what DINOv2 perceives as a single unified group. Strong overlaps ($\geq 80\%$) for 17 ConvNeXt~V2 clusters mapping into DINOv2 clusters are observed. Particularly, multiple ConvNeXt~V2 clusters (3, 5, 15, 17) map entirely (100\%) into DINOv2's cluster~1. The Variation of Information (VI)\cite{meila_comparing_2003} score of 2.090 confirms the models have meaningful differences in clustering, where they largely agree on major video groupings but differ in how finely they subdivide the content.

\subsection{Metrics}
We evaluated the efficacy of clustering statistics, standard clustering metrics, and dataset statistics to find the most suitable visual themes clusterings.
Clustering statistics and dataset statistics provide insights whether a suitable number of clusters has been found: 
We aim to have sufficiently many clusters which are not too large (comprise $\geq 25\%$ of the dataset) and few single video clusters (singleton rate $\leq 15\%$).
Moreover, the dataset coverage of the ten largest clusters indicates how well the clustering is representing the dataset (Top10\% $\geq 80\%$).
Cluster size statistics (mean and median) as well as the GINI index can provide information on the distribution of cluster sizes.
We find however, that these are dataset dependent and are not helpful in determining the clustering quality.
Social media data visual themes are expected to have a high GINI index with few large clusters and a large number of small clusters.
Standard clustering metrics appear less suitable for social media data, given that their low scores stand in contrast with the visual alignment of the clusters.
DinoV2 cluster's silhouette score, Davies-Bouldin score, and Calinski-Harabasz score are consistently better which are not reflected in the manual inspection.
Given that clustering using the MP formulation theoretically yields the optimal solution, we conjecture that the score describe the embedding space rather than the returned clustering. 
See \cref{app:metrics} for more details.

\section{Disussion \& Conclusion}

For zero-shot video classification, frame-wise CLIP emerged as the most reliable model overall, but even it struggled with contextual reasoning, memes and out of distribution [ood] content e.g. Game of Thrones snippets. 
Meanwhile, the VLMs (Video-LLaVA, Video-ChatGPT, and PandaGPT) failed to deliver consistently accurate, interpretable and faithful results for social media climate change-related videos.
It appears that complex visual understanding remains an open research problem.
In terms of clustering, DINOv2 clusters tend to group together different instances of the same broad theme — for example, various scenes of arctic ice — while ConvNeXt V2 may split them into smaller groups depending on secondary cues such as style, presence of people or certain objects, or video format.
We recommend dynamic frame selection and averaging individual frame representations into a global one.

Our zero-shot analysis concluded that \textit{climate action} was the most frequent content in the videos ($65\%$), surpassing both \textit{consequences} ($56\%$) and \textit{animals} ($53\%$) by a large margin (see \cref{fig:clip_animals}, \cref{fig:clip_climateactions}, \cref{fig:clip_consequences}).
For \textit{animals}, \textit{polar bears} and \textit{insects} were more tweeted animals, while \textit{farm animals} exhibited the largest fluctuations.
While the \textit{polar bear} is certainly one of the most common climate visual \cite{o2023visual}, \textit{farm animals} allow for further investigating as a niche topic.
As they are certainly relevant to climate change \cite{bravo2025viral}, the presence and absence of their representations calls for further investigation \cite{blewett2025beyond}.
Given the time period, it was no surprise to see that \textit{Covid} and \textit{health} were highly prevalent in the data, with wildfires being the third most common theme.
Comparing the wildfire theme in videos to previous climate change image studies (e.g. \cite{mcgarry2025fire, ko2024experience}) can deepen our understanding of the differences in representations between the two modalities.
The predictions of \textit{setting} and \textit{type} appear less noisy compared to the other groups.
We assume this is due to higher consistency across frames and fewer options.
Moreover, we find that we can clean the results by removing \textit{memes} from other groups.
Given the distinctiveness of \textit{memes}, previous works investigate them separately to other visuals \cite{sharma2023you, bates2025template}.
We were surprised to see \textit{outer space} as one of the most prevalent settings and also among the largest clusters.
Sharing visualizations of the earth is an abstract format of discussing climate change, more personal communication visuals, showing residential or commercial areas are less frequently used.
Over time, the number of videos with a background increased, indicating a shift towards realistic content rather than visualizations.


Given that this analysis focused on open-source and open-weights methods, the comparison to closed-source models remains for future research.
Moreover, finding a quantitative clustering metric to evaluate visual themes clusterings remains an open research question.

\section*{Acknowledgments}
This project was funded by BMFTR project 16DKWN027b Climate Visions, which was co-funded by the European Union. Experiments were run on the hardware of the University of Mannheim.

{
    \small
    \bibliographystyle{ieeenat_fullname}
    \bibliography{main}

\begin{thebibliography}{79}
\providecommand{\natexlab}[1]{#1}
\providecommand{\url}[1]{\texttt{#1}}
\expandafter\ifx\csname urlstyle\endcsname\relax
  \providecommand{\doi}[1]{doi: #1}\else
  \providecommand{\doi}{doi: \begingroup \urlstyle{rm}\Url}\fi

\bibitem[Andres et~al.(2011{\natexlab{a}})Andres, Kappes, Beier, K{\"o}the, and Hamprecht]{andres2011probabilistic}
Bjoern Andres, J{\"o}rg~H Kappes, Thorsten Beier, Ullrich K{\"o}the, and Fred~A Hamprecht.
\newblock Probabilistic image segmentation with closedness constraints.
\newblock In \emph{International Conference on Computer Vision}. IEEE, 2011{\natexlab{a}}.

\bibitem[Andres et~al.(2011{\natexlab{b}})Andres, Kappes, Beier, Köthe, and Hamprecht]{andres_probabilistic_2011}
Bjoern Andres, Jörg~H. Kappes, Thorsten Beier, Ullrich Köthe, and Fred~A. Hamprecht.
\newblock Probabilistic image segmentation with closedness constraints.
\newblock In \emph{2011 {International} {Conference} on {Computer} {Vision}}, pages 2611--2618, 2011{\natexlab{b}}.
\newblock ISSN: 2380-7504.

\bibitem[Andres et~al.(2016)Andres, Ibeling, Kalofolias, Keuper, Lange, Levinkov, Matten, and Rempfler]{graph_mcmc}
Bjoern Andres, Duligur Ibeling, Giannis Kalofolias, Margret Keuper, Jan-Hendrik Lange, Evgeny Levinkov, Mark Matten, and Markus Rempfler.
\newblock Graphs and graph algorithms in c++.
\newblock \url{http://www.andres.sc/graph.html}, 2016.

\bibitem[Bain et~al.(2021)Bain, Nagrani, Varol, and Zisserman]{bain2021frozen}
Max Bain, Arsha Nagrani, G{\"u}l Varol, and Andrew Zisserman.
\newblock Frozen in time: A joint video and image encoder for end-to-end retrieval.
\newblock In \emph{Proceedings of the IEEE/CVF international conference on computer vision}, pages 1728--1738, 2021.

\bibitem[Bates et~al.(2025)Bates, Christensen, Nakov, and Gurevych]{bates2025template}
Luke Bates, Peter~Ebert Christensen, Preslav Nakov, and Iryna Gurevych.
\newblock A template is all you meme.
\newblock In \emph{Conference of the Nations of the Americas Chapter of the Association for Computational Linguistics: Human Language Technologies}, pages 10443--10475, 2025.

\bibitem[Blewett et~al.(2025)Blewett, Hayes, Westwood, White, and O'Neill]{blewett2025beyond}
Oliver Blewett, Sylvia Hayes, Ned Westwood, Veronica White, and Saffron O'Neill.
\newblock Beyond the “iconic” climate visual: Investigating absent representations of climate change.
\newblock In \emph{The Routledge Companion to Visual Journalism}, pages 214--224. Routledge, 2025.

\bibitem[Bravo et~al.(2025)Bravo, Silva~Luna, and Walter]{bravo2025viral}
Isaac Bravo, Daniel Silva~Luna, and Stefanie Walter.
\newblock Viral climate imagery: examining popular climate visuals on twitter.
\newblock \emph{Visual Communication}, page 14703572251320292, 2025.

\bibitem[Caliński and Harabasz(1974)]{calinski_dendrite_1974}
T. Caliński and J Harabasz.
\newblock A dendrite method for cluster analysis.
\newblock \emph{Communications in Statistics}, 3\penalty0 (1):\penalty0 1--27, 1974.

\bibitem[Chiang et~al.(2023)Chiang, Li, Lin, Sheng, Wu, Zhang, Zheng, Zhuang, Zhuang, Gonzalez, et~al.]{chiang2023vicuna}
Wei-Lin Chiang, Zhuohan Li, Ziqing Lin, Ying Sheng, Zhanghao Wu, Hao Zhang, Lianmin Zheng, Siyuan Zhuang, Yonghao Zhuang, Joseph~E Gonzalez, et~al.
\newblock Vicuna: An open-source chatbot impressing gpt-4 with 90\%* chatgpt quality.
\newblock \emph{See https://vicuna. lmsys. org (accessed 14 April 2023)}, 2\penalty0 (3):\penalty0 6, 2023.

\bibitem[Davies and Bouldin(1979)]{davies_cluster_1979}
David~L. Davies and Donald~W. Bouldin.
\newblock A {Cluster} {Separation} {Measure}.
\newblock \emph{IEEE Transactions on Pattern Analysis and Machine Intelligence}, PAMI-1\penalty0 (2):\penalty0 224--227, 1979.

\bibitem[de~Avila et~al.(2011)de~Avila, Lopes, da~Luz, and de~Albuquerque~Araújo]{de_avila_vsumm_2011}
Sandra Eliza~Fontes de Avila, Ana Paula~Brandão Lopes, Antonio da Luz, and Arnaldo de Albuquerque~Araújo.
\newblock {VSUMM}: {A} mechanism designed to produce static video summaries and a novel evaluation method.
\newblock \emph{Pattern Recognition Letters}, 32\penalty0 (1):\penalty0 56--68, 2011.

\bibitem[Ejaz et~al.(2013)Ejaz, Mehmood, and Wook~Baik]{ejaz_efficient_2013}
Naveed Ejaz, Irfan Mehmood, and Sung Wook~Baik.
\newblock Efficient visual attention based framework for extracting key frames from videos.
\newblock \emph{Signal Processing: Image Communication}, 28\penalty0 (1):\penalty0 34--44, 2013.

\bibitem[Galasso et~al.(2014)Galasso, Keuper, Brox, and Schiele]{Galasso_2014_CVPR}
Fabio Galasso, Margret Keuper, Thomas Brox, and Bernt Schiele.
\newblock Spectral graph reduction for efficient image and streaming video segmentation.
\newblock In \emph{Conference on Computer Vision and Pattern Recognition}. IEEE, 2014.

\bibitem[Gardam et~al.(2025)Gardam, Riedlinger, Angus, and {(Jane) Tan}]{gardam2025}
Caroline Gardam, Michelle Riedlinger, Daniel Angus, and Xue~Ying {(Jane) Tan}.
\newblock Multimodal narratives of climate denial: A novel, visual-first methodology for analysing conspiracy theory discourse on instagram.
\newblock \emph{Discourse, Context \& Media}, 68:\penalty0 100946, 2025.

\bibitem[Girdhar et~al.(2023)Girdhar, El-Nouby, Liu, Singh, Alwala, Joulin, and Misra]{girdhar2023imagebind}
Rohit Girdhar, Alaaeldin El-Nouby, Zhuang Liu, Mannat Singh, Kalyan~Vasudev Alwala, Armand Joulin, and Ishan Misra.
\newblock Imagebind: One embedding space to bind them all.
\newblock In \emph{Proceedings of the IEEE/CVF conference on computer vision and pattern recognition}, pages 15180--15190, 2023.

\bibitem[Gygli et~al.(2014)Gygli, Grabner, Riemenschneider, and Van~Gool]{gygli_creating_2014}
Michael Gygli, Helmut Grabner, Hayko Riemenschneider, and Luc Van~Gool.
\newblock Creating {Summaries} from {User} {Videos}.
\newblock In \emph{Computer {Vision} – {ECCV} 2014}, pages 505--520, Cham, 2014. Springer International Publishing.

\bibitem[Hayes and O’Neill(2024)]{hayes2024transformative}
Sylvia Hayes and Saffron O’Neill.
\newblock Transformative journalisms and the seductive power of imagery in digital climate niche journalism.
\newblock \emph{Journalism}, page 14648849251372742, 2024.

\bibitem[Hayes and O’Neill(2025)]{hayes2025visual}
Sylvia Hayes and Saffron O’Neill.
\newblock Visual politics, protest, and power: Who shaped the climate visual discourse at cop26?
\newblock \emph{Journalism Studies}, 26\penalty0 (4):\penalty0 441--463, 2025.

\bibitem[Ho et~al.(2020{\natexlab{a}})Ho, Kardoost, Pfreundt, Keuper, and Keuper]{ho2020two}
Kalun Ho, Amirhossein Kardoost, Franz-Josef Pfreundt, Janis Keuper, and Margret Keuper.
\newblock A two-stage minimum cost multicut approach to self-supervised multiple person tracking.
\newblock In \emph{Asian Conference on Computer Vision}. Springer, 2020{\natexlab{a}}.

\bibitem[Ho et~al.(2020{\natexlab{b}})Ho, Keuper, and Keuper]{ho2020unsupervised}
Kalun Ho, Janis Keuper, and Margret Keuper.
\newblock Unsupervised multiple person tracking using autoencoder-based lifted multicuts.
\newblock \emph{arXiv preprint arXiv:2002.01192}, 2020{\natexlab{b}}.

\bibitem[Ho et~al.(2020{\natexlab{c}})Ho, Keuper, Pfreundt, and Keuper]{ho2020learning}
Kalun Ho, Janis Keuper, Franz-Josef Pfreundt, and Margret Keuper.
\newblock Learning embeddings for image clustering: An empirical study of triplet loss approaches.
\newblock In \emph{International Conference of Pattern Recognition}. IEEE, 2020{\natexlab{c}}.

\bibitem[Ho et~al.(2021)Ho, Chatzimichailidis, Keuper, and Keuper]{ho2021msm}
Kalun Ho, Avraam Chatzimichailidis, Margret Keuper, and Janis Keuper.
\newblock Msm: Multi-stage multicuts for scalable image clustering.
\newblock In \emph{International Conference on High Performance Computing}. Springer, 2021.

\bibitem[Hu et~al.(2022)Hu, Shen, Wallis, Allen-Zhu, Li, Wang, Wang, Chen, et~al.]{hu2022lora}
Edward~J Hu, Yelong Shen, Phillip Wallis, Zeyuan Allen-Zhu, Yuanzhi Li, Shean Wang, Lu Wang, Weizhu Chen, et~al.
\newblock Lora: Low-rank adaptation of large language models.
\newblock \emph{ICLR}, 1\penalty0 (2):\penalty0 3, 2022.

\bibitem[Huynh-Lam et~al.(2024)Huynh-Lam, Ho-Thi, Tran, and Le]{huynh-lam_cluster-based_2024}
Hai-Dang Huynh-Lam, Ngoc-Phuong Ho-Thi, Minh-Triet Tran, and Trung-Nghia Le.
\newblock Cluster-{Based} {Video} {Summarization} with {Temporal} {Context} {Awareness}.
\newblock In \emph{Image and {Video} {Technology}}, pages 15--28, Singapore, 2024. Springer Nature.

\bibitem[Ilharco et~al.(2021)Ilharco, Wortsman, Wightman, Gordon, Carlini, Taori, Dave, Shankar, Namkoong, Miller, et~al.]{ilharco2021openclip}
Gabriel Ilharco, Mitchell Wortsman, Ross Wightman, Cade Gordon, Nicholas Carlini, Rohan Taori, Achal Dave, Vaishaal Shankar, Hongseok Namkoong, John Miller, et~al.
\newblock Openclip.
\newblock \emph{If you use this software, please cite it as below}, 7, 2021.

\bibitem[Janwe and Bhoyar(2016)]{janwe_video_2016}
Nitin~J Janwe and Kishor~K Bhoyar.
\newblock Video {Key}-{Frame} {Extraction} using {Unsupervised} {Clustering} and {Mutual} {Comparison}.
\newblock \emph{International Journal of Image Processing (IJIP)}, 10\penalty0 (2):\penalty0 73--84, 2016.

\bibitem[Jung and Keuper(2022)]{jung2022learning}
Steffen Jung and Margret Keuper.
\newblock Learning to solve minimum cost multicuts efficiently using edge-weighted graph convolutional neural networks.
\newblock In \emph{Joint European Conference on Machine Learning and Knowledge Discovery in Databases}. Springer, 2022.

\bibitem[Jung et~al.(2022)Jung, Ziegler, Kardoost, and Keuper]{jung2022optimizing}
Steffen Jung, Sebastian Ziegler, Amirhossein Kardoost, and Margret Keuper.
\newblock Optimizing edge detection for image segmentation with multicut penalties.
\newblock In \emph{DAGM German Conference on Pattern Recognition}. Springer, 2022.

\bibitem[Kahl and Atay(2025)]{kahl2025climate}
David~H Kahl and Ahmet Atay.
\newblock Climate change denial messages as post-truth.
\newblock \emph{Journal of Communication Pedagogy}, 9:\penalty0 77--83, 2025.

\bibitem[Kardoost and Keuper(2019)]{kardoost2019solving}
Amirhossein Kardoost and Margret Keuper.
\newblock Solving minimum cost lifted multicut problems by node agglomeration.
\newblock In \emph{Asian Conference on Computer Vision}. Springer, 2019.

\bibitem[Kardoost and Keuper(2021)]{kardoost21}
Amirhossein Kardoost and Margret Keuper.
\newblock Uncertainty in minimum cost multicuts for image and motion segmentation.
\newblock In \emph{Conference on Uncertainty in Artificial Intelligence}. PMLR, 2021.

\bibitem[Keuper(2017)]{K17}
Margret Keuper.
\newblock Higher-order minimum cost lifted multicuts for motion segmentation.
\newblock In \emph{International Conference on Computer Vision}. IEEE, 2017.

\bibitem[Keuper et~al.(2015{\natexlab{a}})Keuper, Andres, and Brox]{KB15b}
Margret Keuper, Bjoern Andres, and Thomas Brox.
\newblock Motion trajectory segmentation via minimum cost multicuts.
\newblock In \emph{International Conference on Computer Vision}. IEEE, 2015{\natexlab{a}}.

\bibitem[Keuper et~al.(2015{\natexlab{b}})Keuper, Levinkov, Bonneel, Lavou{\'e}, Brox, and Andres]{keuper2015efficient}
Margret Keuper, Evgeny Levinkov, Nicolas Bonneel, Guillaume Lavou{\'e}, Thomas Brox, and Bjorn Andres.
\newblock Efficient decomposition of image and mesh graphs by lifted multicuts.
\newblock In \emph{International Conference on Computer Vision}. IEEE, 2015{\natexlab{b}}.

\bibitem[Keuper et~al.(2018)Keuper, Tang, Andres, Brox, and Schiele]{KB19}
Margret Keuper, Siyu Tang, Bjoern Andres, Thomas Brox, and Bernt Schiele.
\newblock Motion segmentation and multiple object tracking by correlation co-clustering.
\newblock \emph{Transactions on Pattern Analysis and Machine Intelligence}, 42\penalty0 (1), 2018.

\bibitem[Ko et~al.(2024)Ko, Ni, Taylor, Chen, Huang, Kumar, Alsudais, Wang, Liu, Wang, et~al.]{ko2024experience}
Jessie~WY Ko, Shengquan Ni, Alexander Taylor, Xiusi Chen, Yicong Huang, Avinash Kumar, Sadeem Alsudais, Zuozhi Wang, Xiaozhen Liu, Wei Wang, et~al.
\newblock How the experience of california wildfires shape twitter climate change framings.
\newblock \emph{Climatic Change}, 177\penalty0 (1):\penalty0 17, 2024.

\bibitem[Lee et~al.(2020)Lee, Lee, Ng, and Natsev]{lee_large_2020}
Hyodong Lee, Joonseok Lee, Joe Yue-Hei Ng, and Paul Natsev.
\newblock Large {Scale} {Video} {Representation} {Learning} via {Relational} {Graph} {Clustering}.
\newblock In \emph{Conference on computer vision and pattern recognition}, pages 6807--6816, 2020.

\bibitem[Levinkov et~al.(2022)Levinkov, Kardoost, Andres, and Keuper]{kardoost22}
Evgeny Levinkov, Amirhossein Kardoost, Bjoern Andres, and Margret Keuper.
\newblock Higher-order multicuts for geometric mmdel fitting and motion segmentation.
\newblock \emph{Transactions on Pattern Analysis and Machine Intelligence}, 45\penalty0 (1), 2022.

\bibitem[Lin et~al.(2024)Lin, Ye, Zhu, Cui, Ning, Jin, and Yuan]{lin2024video}
Bin Lin, Yang Ye, Bin Zhu, Jiaxi Cui, Munan Ning, Peng Jin, and Li Yuan.
\newblock Video-llava: Learning united visual representation by alignment before projection.
\newblock In \emph{Conference on Empirical Methods in Natural Language Processing}, pages 5971--5984, 2024.

\bibitem[Liu et~al.(2023)Liu, Li, Wu, and Lee]{liu2023visual}
Haotian Liu, Chunyuan Li, Qingyang Wu, and Yong~Jae Lee.
\newblock Visual instruction tuning.
\newblock \emph{Advances in neural information processing systems}, 36:\penalty0 34892--34916, 2023.

\bibitem[Liu et~al.(2024)Liu, Li, Li, and Lee]{liu2024improved}
Haotian Liu, Chunyuan Li, Yuheng Li, and Yong~Jae Lee.
\newblock Improved baselines with visual instruction tuning.
\newblock In \emph{Proceedings of the IEEE/CVF Conference on Computer Vision and Pattern Recognition}, pages 26296--26306, 2024.

\bibitem[Liu and Fan(2005)]{liu_combined_2005}
Lijie Liu and Guoliang Fan.
\newblock Combined key-frame extraction and object-based video segmentation.
\newblock \emph{IEEE Transactions on Circuits and Systems for Video Technology}, 15\penalty0 (7):\penalty0 869--884, 2005.

\bibitem[Luke(2024)]{luke2024climate}
Timothy~W Luke.
\newblock Climate change deniers versus climate change decriers: the pragmatics of climate defense in the age of disinformation.
\newblock \emph{FAST CAPITALISM}, 21\penalty0 (1), 2024.

\bibitem[Luo et~al.(2023)Luo, Zhao, Yang, Dong, Li, Lu, Wang, Hu, Qiu, and Wei]{luo2023valley}
Ruipu Luo, Ziwang Zhao, Min Yang, Junwei Dong, Da Li, Pengcheng Lu, Tao Wang, Linmei Hu, Minghui Qiu, and Zhongyu Wei.
\newblock Valley: Video assistant with large language model enhanced ability.
\newblock \emph{arXiv preprint arXiv:2306.07207}, 2023.

\bibitem[Maaz et~al.(2023)Maaz, Rasheed, Khan, and Khan]{maaz2023video}
Muhammad Maaz, Hanoona Rasheed, Salman Khan, and Fahad~Shahbaz Khan.
\newblock Video-chatgpt: Towards detailed video understanding via large vision and language models.
\newblock \emph{arXiv preprint arXiv:2306.05424}, 2023.

\bibitem[McGarry and Trer{\'e}(2025)]{mcgarry2025fire}
Aidan McGarry and Emiliano Trer{\'e}.
\newblock Fire as an aesthetic resource in climate change communication: exploring the visual discourse of the california wildfires on twitter/x.
\newblock \emph{Visual Studies}, 40\penalty0 (2):\penalty0 337--351, 2025.

\bibitem[Meilă(2003)]{meila_comparing_2003}
Marina Meilă.
\newblock Comparing {Clusterings} by the {Variation} of {Information}.
\newblock In \emph{Learning {Theory} and {Kernel} {Machines}}, pages 173--187, Berlin, Heidelberg, 2003. Springer.

\bibitem[Mendy et~al.(2024)Mendy, Karlsson, and Lindvall]{mendy2024counteracting}
Laila Mendy, Mikael Karlsson, and Daniel Lindvall.
\newblock Counteracting climate denial: A systematic review.
\newblock \emph{Public Understanding of Science}, 33\penalty0 (4):\penalty0 504--520, 2024.

\bibitem[Mooseder et~al.(2023)Mooseder, Brantner, Zamith, and Pfeffer]{mooseder_social_2023}
Angelina Mooseder, Cornelia Brantner, Rodrigo Zamith, and Jürgen Pfeffer.
\newblock ({Social}) {Media} {Logics} and {Visualizing} {Climate} {Change}: 10 {Years} of \#climatechange {Images} on {Twitter}.
\newblock \emph{Social Media + Society}, 9\penalty0 (1):\penalty0 20563051231164310, 2023.
\newblock Publisher: SAGE Publications Ltd.

\bibitem[Mundur et~al.(2006)Mundur, Rao, and Yesha]{mundur_keyframe-based_2006}
Padmavathi Mundur, Yong Rao, and Yelena Yesha.
\newblock Keyframe-based video summarization using {Delaunay} clustering.
\newblock \emph{International Journal on Digital Libraries}, 6\penalty0 (2):\penalty0 219--232, 2006.

\bibitem[N and S(2016)]{n_image_2016}
Senthilkumaran N and Vaithegi S.
\newblock Image {Segmentation} {By} {Using} {Thresholding} {Techniques} {For} {Medical} {Images}.
\newblock \emph{Computer Science \& Engineering: An International Journal}, 6\penalty0 (1):\penalty0 1--13, 2016.

\bibitem[Nguyen et~al.(2022)Nguyen, Henschel, Rosenhahn, Sonntag, and Swoboda]{nguyen2022lmgp}
Duy~MH Nguyen, Roberto Henschel, Bodo Rosenhahn, Daniel Sonntag, and Paul Swoboda.
\newblock Lmgp: Lifted multicut meets geometry projections for multi-camera multi-object tracking.
\newblock In \emph{Conference on Computer Vision and Pattern Recognition}. IEEE, 2022.

\bibitem[Nwokolo(2025)]{nwokolo2025climate}
Samuel~Chukwujindu Nwokolo.
\newblock Climate hoax: The shift from scientific discourse to speculative rhetoric in climate change conversations.
\newblock \emph{Next Research}, page 100322, 2025.

\bibitem[O'Neill and Schäfer(2017)]{schafer_frame_2017}
Saffron O'Neill and Mike~S. Schäfer.
\newblock Frame {Analysis} in {Climate} {Change} {Communication}: {Approaches} for {Assessing} {Journalists}’ {Minds}, {Online} {Communication} and {Media} {Portrayals}.
\newblock In \emph{Schäfer, {Mike} {S}; {O}'{Neill}, {Saffron} (2017). {Frame} {Analysis} in {Climate} {Change} {Communication}: {Approaches} for {Assessing} {Journalists}’ {Minds}, {Online} {Communication} and {Media} {Portrayals}. {In}: {Nisbet}, {Matthew}; {Ho}, {Shirley}; {Markowitz}, {Ezra}; {O}'{Neill}, {Saffron}; {Schäfer}, {Mike} {S}; {Thaker}, {Jagadish}. {Oxford} {Encyclopedia} of {Climate} {Change} {Communication}. {New} {York}: {Oxford} {University} {Press}, n/a.}, page n/a. Oxford University Press, New York, 2017.

\bibitem[O'Neill et~al.(2023)O'Neill, Hayes, Strau\ss, Doutreix, Steentjes, Ettinger, Westwood, and Painter]{o2023visual}
Saffron O'Neill, Sylvia Hayes, Nadine Strau\ss, Marie-No{\"e}lle Doutreix, Katharine Steentjes, Joshua Ettinger, Ned Westwood, and James Painter.
\newblock Visual portrayals of fun in the sun in european news outlets misrepresent heatwave risks.
\newblock \emph{The Geographical Journal}, 189\penalty0 (1), 2023.

\bibitem[Oquab et~al.(2024)Oquab, Darcet, Moutakanni, Vo, Szafraniec, Khalidov, Fernandez, HAZIZA, Massa, El-Nouby, Assran, Ballas, Galuba, Howes, Huang, Li, Misra, Rabbat, Sharma, Synnaeve, Xu, Jegou, Mairal, Labatut, Joulin, and Bojanowski]{oquab2023dinov2}
Maxime Oquab, Timoth{\'e}e Darcet, Th{\'e}o Moutakanni, Huy~V. Vo, Marc Szafraniec, Vasil Khalidov, Pierre Fernandez, Daniel HAZIZA, Francisco Massa, Alaaeldin El-Nouby, Mido Assran, Nicolas Ballas, Wojciech Galuba, Russell Howes, Po-Yao Huang, Shang-Wen Li, Ishan Misra, Michael Rabbat, Vasu Sharma, Gabriel Synnaeve, Hu Xu, Herve Jegou, Julien Mairal, Patrick Labatut, Armand Joulin, and Piotr Bojanowski.
\newblock {DINO}v2: Learning robust visual features without supervision.
\newblock \emph{Transactions on Machine Learning Research}, 2024.

\bibitem[Prasse et~al.(2023)Prasse, Jung, Bravo, Walter, and Keuper]{prasse2023towards}
Katharina Prasse, Steffen Jung, Isaac~B Bravo, Stefanie Walter, and Margret Keuper.
\newblock Towards understanding climate change perceptions: A social media dataset.
\newblock In \emph{NeurIPS 2023 Workshop on Tackling Climate Change with Machine Learning}, 2023.

\bibitem[Prasse et~al.(2025)Prasse, Bravo, Walter, and Keuper]{prasse_i_2025}
Katharina Prasse, Isaac Bravo, Stefanie Walter, and Margret Keuper.
\newblock I {Spy} with {My} {Little} {Eye} a {Minimum} {Cost} {Multicut} {Investigation} of {Dataset} {Frames}.
\newblock In \emph{2025 {IEEE}/{CVF} {Winter} {Conference} on {Applications} of {Computer} {Vision} ({WACV})}, pages 2134--2143, 2025.
\newblock ISSN: 2642-9381.

\bibitem[Radford et~al.(2021)Radford, Kim, Hallacy, Ramesh, Goh, Agarwal, Sastry, Askell, Mishkin, Clark, et~al.]{radford2021learning}
Alec Radford, Jong~Wook Kim, Chris Hallacy, Aditya Ramesh, Gabriel Goh, Sandhini Agarwal, Girish Sastry, Amanda Askell, Pamela Mishkin, Jack Clark, et~al.
\newblock Learning transferable visual models from natural language supervision.
\newblock In \emph{International conference on machine learning}, pages 8748--8763. PmLR, 2021.

\bibitem[Rebich-Hespanha and Rice(2016)]{rebich-hespanha_dominant_2016}
Stacy Rebich-Hespanha and Ronald~E. Rice.
\newblock Dominant {Visual} {Frames} in {Climate} {Change} {News} {Stories}: {Implications} for {Formative} {Evaluation} in {Climate} {Change} {Campaigns}.
\newblock \emph{International Journal of Communication (19328036)}, 10, 2016.

\bibitem[Rousseeuw(1987)]{rousseeuw_silhouettes_1987}
Peter~J. Rousseeuw.
\newblock Silhouettes: {A} graphical aid to the interpretation and validation of cluster analysis.
\newblock \emph{Journal of Computational and Applied Mathematics}, 20:\penalty0 53--65, 1987.

\bibitem[Seelig et~al.(2022)Seelig, Deng, and Liang]{seelig_frame_2022}
Michelle~I. Seelig, Huixin Deng, and Songyi Liang.
\newblock A frame analysis of climate change solutions in legacy news and digital media.
\newblock \emph{Newspaper Research Journal}, 43\penalty0 (4):\penalty0 370--388, 2022.
\newblock Publisher: SAGE Publications Inc.

\bibitem[Sharma et~al.(2018)Sharma, Ding, Goodman, and Soricut]{sharma2018conceptual}
Piyush Sharma, Nan Ding, Sebastian Goodman, and Radu Soricut.
\newblock Conceptual captions: A cleaned, hypernymed, image alt-text dataset for automatic image captioning.
\newblock In \emph{Proceedings of the 56th Annual Meeting of the Association for Computational Linguistics (Volume 1: Long Papers)}, pages 2556--2565, 2018.

\bibitem[Sharma et~al.(2023)Sharma, Agarwal, Suresh, Nakov, Akhtar, and Chakraborty]{sharma2023you}
Shivam Sharma, Siddhant Agarwal, Tharun Suresh, Preslav Nakov, Md~Shad Akhtar, and Tanmoy Chakraborty.
\newblock What do you meme? generating explanations for visual semantic role labelling in memes.
\newblock In \emph{AAAI Conference on Artificial Intelligence}, pages 9763--9771, 2023.

\bibitem[Su et~al.(2023)Su, Lan, Li, Xu, Wang, and Cai]{su2023pandagpt}
Yixuan Su, Tian Lan, Huayang Li, Jialu Xu, Yan Wang, and Deng Cai.
\newblock Pandagpt: One model to instruction-follow them all.
\newblock \emph{arXiv preprint arXiv:2305.16355}, 2023.

\bibitem[Tang et~al.(2016)Tang, Andres, Andriluka, and Schiele]{tang2016multi}
Siyu Tang, Bjoern Andres, Mykhaylo Andriluka, and Bernt Schiele.
\newblock Multi-person tracking by multicut and deep matching.
\newblock In \emph{ECCV Workshop on Benchmarking Multi-Target Tracking: MOTChallenge}. Springer, 2016.

\bibitem[Tang et~al.(2017)Tang, Andriluka, Andres, and Schiele]{tang2017multiple}
Siyu Tang, Mykhaylo Andriluka, Bjoern Andres, and Bernt Schiele.
\newblock Multiple people tracking by lifted multicut and person re-identification.
\newblock In \emph{Conference on Computer Vision and Pattern Recognition}. IEEE, 2017.

\bibitem[Truong and Venkatesh(2007)]{truong_video_2007}
Ba~Tu Truong and Svetha Venkatesh.
\newblock Video abstraction: {A} systematic review and classification.
\newblock \emph{ACM Trans. Multimedia Comput. Commun. Appl.}, 3\penalty0 (1):\penalty0 3--es, 2007.

\bibitem[Wardekker and Lorenz(2019)]{wardekker_visual_2019}
Arjan Wardekker and Susanne Lorenz.
\newblock The visual framing of climate change impacts and adaptation in the {IPCC} assessment reports.
\newblock \emph{Climatic Change}, 156\penalty0 (1):\penalty0 273--292, 2019.

\bibitem[Woo et~al.(2023)Woo, Debnath, Hu, Chen, Liu, Kweon, and Xie]{woo2023convnext}
Sanghyun Woo, Shoubhik Debnath, Ronghang Hu, Xinlei Chen, Zhuang Liu, In~So Kweon, and Saining Xie.
\newblock Convnext v2: Co-designing and scaling convnets with masked autoencoders.
\newblock In \emph{Conference on Computer Vision and Pattern Recognition}. IEEE, 2023.

\bibitem[Wozniak et~al.(2017)Wozniak, Wessler, and Lück]{wozniak_who_2017}
Antal Wozniak, Hartmut Wessler, and Julia Lück.
\newblock Who {Prevails} in the {Visual} {Framing} {Contest} about the {United} {Nations} {Climate} {Change} {Conferences}?
\newblock \emph{Journalism Studies}, 18\penalty0 (11):\penalty0 1433--1452, 2017.
\newblock Publisher: Routledge \_eprint: https://doi.org/10.1080/1461670X.2015.1131129.

\bibitem[Yan and Sch{\"a}fer(2025)]{yan2025multimodal}
Xiaoyue Yan and Mike~S Sch{\"a}fer.
\newblock Multimodal climate change communication on wechat: analyzing visual/textual clusters on china’s largest social media platform.
\newblock \emph{Climatic Change}, 178\penalty0 (7):\penalty0 133, 2025.

\bibitem[Yang and Lin(2005)]{yang_key_2005}
Shuping Yang and Xinggang Lin.
\newblock Key {Frame} {Extraction} {Using} {Unsupervised} {Clustering} {Based} on a {Statistical} {Model}.
\newblock \emph{Tsinghua Science \& Technology}, 10\penalty0 (2):\penalty0 169--173, 2005.

\bibitem[Yeung and Liu(1995)]{yeung_efficient_1995}
M.M. Yeung and Bede Liu.
\newblock Efficient matching and clustering of video shots.
\newblock In \emph{Proceedings., {International} {Conference} on {Image} {Processing}}, pages 338--341 vol.1, 1995.

\bibitem[Zeng and Yan(2024)]{zeng2024understanding}
Jing Zeng and Xiaoyue Yan.
\newblock Understanding climate-related visual storytelling on tiktok: A cross-national multimodal analysis.
\newblock \emph{Journal of digital social research}, 6\penalty0 (2):\penalty0 66--84, 2024.

\bibitem[Zhang et~al.(2016)Zhang, Chao, Sha, and Grauman]{zhang_video_2016}
Ke Zhang, Wei-Lun Chao, Fei Sha, and Kristen Grauman.
\newblock Video {Summarization} with {Long} {Short}-{Term} {Memory}.
\newblock In \emph{Computer {Vision} – {ECCV} 2016}, pages 766--782, Cham, 2016. Springer International Publishing.

\bibitem[Zhu et~al.(2023{\natexlab{a}})Zhu, Lin, Ning, Yan, Cui, Wang, Pang, Jiang, Zhang, Li, et~al.]{zhu2023languagebind}
Bin Zhu, Bin Lin, Munan Ning, Yang Yan, Jiaxi Cui, HongFa Wang, Yatian Pang, Wenhao Jiang, Junwu Zhang, Zongwei Li, et~al.
\newblock Languagebind: Extending video-language pretraining to n-modality by language-based semantic alignment.
\newblock \emph{arXiv preprint arXiv:2310.01852}, 2023{\natexlab{a}}.

\bibitem[Zhu et~al.(2023{\natexlab{b}})Zhu, Chen, Shen, Li, and Elhoseiny]{zhu2023minigpt}
Deyao Zhu, Jun Chen, Xiaoqian Shen, Xiang Li, and Mohamed Elhoseiny.
\newblock Minigpt-4: Enhancing vision-language understanding with advanced large language models.
\newblock \emph{arXiv preprint arXiv:2304.10592}, 2023{\natexlab{b}}.

\bibitem[Zhuang et~al.(1998)Zhuang, Rui, Huang, and Mehrotra]{zhuang_adaptive_1998}
Yueting Zhuang, Yong Rui, T.S. Huang, and S. Mehrotra.
\newblock Adaptive key frame extraction using unsupervised clustering.
\newblock In \emph{Proceedings 1998 {International} {Conference} on {Image} {Processing}. {ICIP98} ({Cat}. {No}.{98CB36269})}, pages 866--870 vol.1, 1998.

\end{thebibliography}
}

\newpage
\appendix
\onecolumn
\begin{center}
\appendixhead
{\Large{--- Supplemental Material ---}}
\end{center}

\section{Prompts}
\label{app:promts}

\subsection{Animals}
The initial prompt used for the \textit{animals} category is as follows:

\begin{quote}
Analyze the video. 
If the video contains animals like for example
pets, farm animals, polar bears, land mammals, sea mammals, fish, amphibians, reptiles, invertebrates or birds
answer with "Yes, ...". If the video is not about animals answer with "No, ...".
\end{quote}

An detailed prompt used for the \textit{animals} group is as follows:
\begin{quote}
Analyze the video. What kind of animals are featured? Select from the classes 1 to 12: \newline
1: Pets (e.g. dogs, cats, guinea pigs, horses, bunnies, canaries); \newline
2: Farm animals (e.g. cattle, pigs, chickens, sheep, goats);\newline
3: Polar bears;  \newline
4: Land mammals (e.g. lions, tigers, elephants, giraffes, deer, wolves, bears);  \newline
5: Sea mammals (e.g. whales, dolphins, seals, walruses - excludes polar bears and penguins); \newline
6: Fish/Non-mammal sea life (e.g. fish, jellyfish, crabs, lobsters);  \newline
7: Amphibians (e.g. frogs, toads, salamanders);  \newline
8: Reptiles (e.g. snakes, lizards, turtles, crocodiles); \newline
9: Invertebrates (e.g. spiders, worms, snails);  \newline
10: Birds (e.g. eagles, owls, parrots, penguins);  \newline
11: Insects (e.g. ants, bees, butterflies);  \newline
12: Other animals (no pets, farm animals, polar bears, land or sea mammals, fish, amphibians, reptiles, invertebrates, birds or insects);\newline
Answer only with the class-number of the relevant categories. No extra words or explanations. 
Before answering check if your answer contains the class-number of the category. 
\end{quote}

\subsection{Climate action}
The initial prompt used for the \textit{climate action} category is as follows:
\begin{quote}
    Analyze the video. 
If the video is about climateactions like for example
protests, politics, sustainable energy (wind, solar, hydropower, biogas) or fossil energy (carbon, natural gas, oil, fossil fuel)
answer with "Yes, ...". If the video is not about climateactions answer with "No, ...".
\end{quote}

The detailed prompt used for the \textit{climate action} category is as follows:

\begin{quote}
    Analyze the video. What kind of climate actions is the video featuring? Select from the classes 1 to 10:
1: Politics (e.g. COP conferences, climate summits, policy meetings);
2: Protests (e.g. climate marches, Fridays for Future, Extinction Rebellion);
3: Solar energy (e.g. solar panels, solar farms);
4: Wind energy (e.g. wind turbines, wind farms);
5: Hydropower (e.g. hydroelectric dams, tidal energy);
6: Bioenergy (e.g. biogas, biomass plants);
7: Coal (e.g. coal mines, coal power plants);
8: Oil (e.g. oil rigs, oil pipelines);
9: Natural gas (e.g. fracking, gas power plants);
10: Other climate action (no politics, protest, solar energy, wind energy, hydropower, bioenergy, coal or natural gas);
Answer only with the class-number of the relevant categories. No extra words or explanations. 
Before answering check if your answer contains the class-number of the category. 
\end{quote}

\subsection{Consequences}
The initial prompt used for the \textit{consequences} category is as follows:
\begin{quote}
    Analyze the video. 
If the video is about climate consequences like for example
biodeversity loss, covid, health, extrem weather (drough, flood, wildfire), melting ice, sea-level rise, rising temperatures, human rights or economic consequences 
answer with "Yes, ...". If the video is not about climate consequences answer with "No, ...".
\end{quote}
The detailed prompt used for the \textit{consequences} category is as follows:
\begin{quote}
    Analyze the video. What kind of climate consequences is the video featuring? Select from the classes 1 to 13:
1: Floods (e.g. inundation from heavy rain, river overflow);
2: Drought (e.g. dry land, parched conditions, water scarcity);
3: Wildfires (e.g. forest fires, grassland fires);
4: Rising temperature (e.g. heatwaves, heat stress indicators);
5: Other extreme weather events (no floods, droughts, wildfires or rising temperatures);
6: Melting Ice (e.g. glaciers, icebergs, polar ice loss);
7: Sea level rise (e.g. coastal erosion, submerged areas);
8: Human rights (e.g. climate migration, conflict over resources);
9: Economic consequences (e.g. market impacts, financial losses);
10: Biodiversity loss (e.g. species extinction, deforestation);
11: Covid (e.g. corona crisis related);
12: Health (e.g. pandemics, health crises);
13: Other consequence (no floods, droughts, wildfires, rising temperature, other extreme weather events, melting ice, sea level rise, human rights, economic consequences, bioddiversity loss, covid or health);
Answer only with the class-number of the relevant categories. No extra words or explanations. 
Before answering check if your answer contains the class-number of the category. 

\end{quote}
\subsection{Setting}
The prompt used for the \textit{setting} category is as follows:
\begin{quote}
    Analyze the video. What kind of setting is the video in? Select from the classes 1 to 15:
1: No setting (e.g. single-color background, abstract patterns); 
2: Residential area(e.g. apartment buildings, lawns);
3: Industrial area (e.g. factories, warehouses, pipelines);
4: Commercial area (e.g. shops, shopping centre, malls);
5: Agricultural (e.g. farms, barns, fields, feedlots); 
6: Rural (e.g. rural landscape, smalll town);
7: Indoor space (e.g. rooms, offices, indoor facilities); 
8: Arctic, Antarctica (e.g. polar landscapes, ice sheets); 
9: Ocean (e.g. mostly sea views, underwater);
10: Coastal (e.g. beaches, sea views with land); 
11: Desert (e.g. sand dunes, arid landscapes); 
12: Forest, jungle (e.g. woodlands, rainforests); 
13: Other Nature (no artic or antarctica, ocean, coastal, desert, forest or jungle); 
14: Outer space (e.g. planets, stars, space missions); 
15: Other setting (no residential, industrial, commercial, aggricultural, rural area, no indoor space, artic or antarctica, ocean, coastal, desert, forest, jungle or outer space);
Answer only with the class-number of the relevant categories. No extra words or explanations. 
Before answering check if your answer contains the class-number of the category. 
\end{quote}
\subsection{Type}
The prompt used for the \textit{type} category is as follows:
\begin{quote}
    Analyze the video. What type of video is it? Select from the classes 1 to 9:
1: Event invitations (e.g. event ads, promotional materials);
2: Meme (e.g. humorous/satirical image-text combinations);
3: Infographic (e.g. visual overviews with minimal text);
4: Data visualization (e.g. graphs, charts, detailed data maps);
5: Illustration (e.g. drawings, paintings, cartoons);
6: Screenshot (e.g. captured text, news articles, social media posts);
7: Single photo (e.g. individual photographs, may include text overlay);
8: Photo collage (e.g. multiple combined images);
9: Other type (no event inivitation, meme, infographic, data visualization, illustration, screenshot, single photo or photo collage);
Answer only with the class-number of the relevant categories. No extra words or explanations. 
Before answering check if your answer contains the class-number of the category. 
\end{quote}
\newpage
\section{Frame Selection Methods}
\label{app:frameselection}

We introduce the following notation: let a video $V$ consist of $N$ frames with corresponding embeddings $\{\mathbf{f}_1, \mathbf{f}_2, ..., \mathbf{f}_N\}$ where $\mathbf{f}_i \in \mathbb{R}^d$. Each combination method produces a video-level representation $\mathbf{v} \in \mathbb{R}^d$ through different aggregation strategies.

\subsection{Baseline: Simple Averaging}
\label{sec:avg_pooling}
The baseline approach assigns equal importance to all frames:
\[
\mathbf{v}_{\text{avg}} = \frac{1}{N} \sum_{i=1}^N \mathbf{f}_i
\]
This method makes no assumptions about frame relevance, treating frames in transitional footage or key moments identically. While computationally efficient with $O(N)$ complexity, it might dilute distinctive visual elements within averaged representations.

\subsection{Single Frame Method: Max Confidence (ConvNeXt V2) }
\label{sec:Maxconfidence}

This approach identifies the most semantically clear frame according to the pre-trained classifier:
\[
\mathbf{v}_{\text{max}} = \mathbf{f}_{k^*} \quad \text{where} \quad k^* = \arg\max_{i \in [1,N]} \max_{j} p_{ij}
\]
with $p_{ij}$ representing the softmax probability for class $j$ in frame $i$. Although computationally efficient after initial classification, this method reduces entire videos to single frames, potentially missing temporal dynamics and multiple visual themes. Also, another potential drawback is that the classifier confidence is based on ImageNet categories, which may not align with climate-relevant content.

\subsection{Weighted Combination Methods}
\label{sec:weighted_combination}

These methods calculate frame-specific weights $w_i$ where $\sum_{i=1}^N w_i = 1$:
\[
\mathbf{v}_{\text{weighted}} = \sum_{i=1}^N w_i \mathbf{f}_i
\]

\subsubsection{Weighted Diversity}
This approach emphasizes frames that differ from others in the video. For each frame, this work measures its distinctiveness through similarity variance:
\[
d_i = \text{var}(\{s_{ij} : j \in [1,N], j \neq i\})
\]
where $s_{ij} = \frac{\mathbf{f}_i^T \mathbf{f}_j}{||\mathbf{f}_i|| \cdot ||\mathbf{f}_j||}$ represents cosine similarity. Weights follow a temperature-controlled distribution:
\[
w_i = \frac{\exp(d_i / \tau)}{\sum_{k=1}^N \exp(d_k / \tau)}
\]
with temperature $\tau$ controlling weight concentration. Intuitively, frames with high variance in their similarities to other frames are considered more distinctive and receive higher weights, which captures moments that stand out visually from the rest of the video.

\subsubsection{Weighted Confidence (ConvNeXt V2)}
Building on the max confindence approach, frames receive weights proportional to their max classification confidence:
\[
c_i = \max_{j} p_{ij} \quad \text{and} \quad w_i = \frac{\exp(c_i / \tau)}{\sum_{k=1}^N \exp(c_k / \tau)}
\]
This approach uses all frames and reduces single-frame bias. However, the temperature parameter partially mitigates but doesn't eliminate the ImageNet bias.

\subsubsection{Temporal Coherence Weighting (DINOv2)}
This method favors frames that maintain visual coherence with temporal neighbors. For frame $i$ with temporal window $\mathcal{W}_i$ of size $2r+1$ centered at position $i$:
\[
t_i = \frac{1}{|\mathcal{W}_i|-1} \sum_{j \in \mathcal{W}_i \setminus \{i\}} s_{ij}
\]
Weights again follow the softmax formulation with consistency scores $t_i$. This approach assumes narrative continuity, which potentially suits documentary-style content, but it may underweight abrupt but important changes in videos.

\subsection{Implementation Considerations}
\label{sec:implementation}

\cref{tab:method_properties} summarizes key properties of each method. 
Temperature parameters were set to $\tau = 2.0$ for distinctiveness and confidence weighting, aiming for moderate weight distribution, and $\tau = 1.0$ for temporal consistency. 
Window radius is set to $r = 1$ (3-frame windows) for balancing temporal context with computational efficiency.

\begin{table}[ht]
\centering
\footnotesize
\caption{Comparison of frame combination methods}
\begin{tabular}{lccccc}
\hline
\textbf{Method} & \textbf{Models} & \textbf{Complexity} & \textbf{Temporal} & \textbf{External Bias} \\
\hline
Simple Average & Both & $O(N)$ & No & None \\
Max Confidence & ConvNeXt V2 & $O(N)$ & No & ImageNet \\
Weighted Diversity & Both & $O(N^2)$ & No & None \\
Weighted Confidence & ConvNeXt V2 & $O(N)$ & No & ImageNet \\
Temporal Coherence & DINOv2 & $O(Nr)$ & Yes & None \\
\hline
\end{tabular}
\label{tab:method_properties}
\end{table}

Each method represents specific assumptions about video content. Simple averaging assumes consistency across frames, potentially diluting brief but critical moments. Confidence-based approaches suppose ImageNet categories (e.g., "person," "building") correlate with climate communication importance, which may not hold for climate-related social media video content. Weighted diversity prioritizes visual uniqueness, which may overemphasize transition frames rather than semantically important content. Temporal coherence assumes smooth narrative flow, potentially missing important discontinuities in videos.

\section{Clustering Evaluation}
\label{app:clusteval}
To evaluate the clustering results quantitatively, this work proposes a multi-metric scoring framework, combining six metrics normalized to the [0, 1] range and weighted by their relevance. With the goal of identifying meaningful and interpretable visual frames, the scoring system mainly takes into account clustering quality and coverage balance. The three core quality metrics—silhouette score, Davies-Bouldin, and Calinski-Harabasz—receive 70\% of the total weight, prioritizing clustering cohesion, separation, and density. The remaining 30\% account for pragmatic factors such as cluster balance (via the Gini coefficient), coverage efficiency(top-10 coverage ratio), and over-fragmentation (singleton ratio), which are important for qualitative inspection.

\textbf{Cluster Quality Assessment.} The silhouette coefficient~\cite{rousseeuw_silhouettes_1987} (30\% weight) measures how similar a video is to its own cluster compared to other clusters. For each video $i$, the silhouette score is defined as:
\begin{equation}
s(i) = \frac{b(i) - a(i)}{\max\{a(i), b(i)\}}
\end{equation}
where $a(i)$ is the average distance from $i$ to all other videos in the same cluster, and $b(i)$ is the minimum average distance from $i$ to videos in any other cluster. The coefficient ranges from -1 to 1, where values near 1 indicate that videos are well-matched to their clusters. This metric is important for visual frame detection as it quantifies whether  videos in one cluster are internally coherent.

The Calinski-Harabasz index~\cite{calinski_dendrite_1974} (20\% weight) evaluates cluster quality through the ratio of between-cluster to within-cluster variance:
\begin{equation}
CH = \frac{\text{trace}(B_k)/(k-1)}{\text{trace}(W_k)/(n-k)}
\end{equation}
where $B_k$ is the between-group sum of squares matrix, $W_k$ is the within-group sum of squares matrix, $k$ is the number of clusters, and $n$ is the number of videos. Higher values mean better-defined clusters with tight intra-cluster cohesion and good inter-cluster separation.

The Davies-Bouldin index~\cite{davies_cluster_1979} (20\% weight) quantifies the average similarity ratio between each cluster and its most similar neighbor:
\begin{equation}
DB = \frac{1}{k} \sum_{i=1}^{k} \max_{j \neq i} \left( \frac{s_i + s_j}{d_{ij}} \right)
\end{equation}
where $s_i$ is the average distance between videos in cluster $i$ and its centroid, and $d_{ij}$ is the distance between centroids of clusters $i$ and $j$. Lower values indicate better separation, making this metric valuable for ensuring discovered video clusters are distinct from each other.

\textbf{Practical Considerations.} Three additional metrics address practical concerns. The Gini coefficient (10\% weight) measures cluster size inequality:
\begin{equation}
G = \frac{\sum_{i=1}^{k} \sum_{j=1}^{k} |n_i - n_j|}{2k \sum_{i=1}^{k} n_i}
\end{equation}
where $n_i$ is the size of cluster $i$. Values near 0 indicate equal-sized clusters, while values near 1 indicate severe imbalance.

Coverage efficiency (10\% weight) measures the proportion of videos contained in the 10 largest clusters:
\begin{equation}
C_{10} = \frac{\sum_{i=1}^{10} n_i}{\sum_{i=1}^{k} n_i}
\end{equation}
This metric rewards configurations where a manageable number of clusters represent the majority of the dataset, which helps manual inspection.

The singleton ratio (10\% weight) quantifies over-fragmentation:
\begin{equation}
R_s = \frac{|\{i : n_i = 1\}|}{k}
\end{equation}
High singleton ratios suggest excessive segmentation, which could complicate frame interpretation.

Each metric is normalized to [0, 1]: silhouette scores are linearly scaled from [-1, 1]; Davies-Bouldin scores are inverted using $1 - \min(\text{value}/5, 1)$ to align with the scoring direction; and Calinski-Harabasz scores are logarithmically scaled via $\min(\log(1 + \text{value})/10, 1)$ to tackle their unbounded nature. The final composite score is the weighted sum of all normalized metrics.

Additionaly, for the comparative analysis, we also applies a Chi-square test to assess temporal independence of clusters. For a contingency table of cluster assignments versus years, the test statistic is:
\begin{equation}
\chi^2 = \sum_{i,j} \frac{(O_{ij} - E_{ij})^2}{E_{ij}}
\end{equation}
where $O_{ij}$ is the observed count and $E_{ij}$ is the expected count under independence. Per-cluster temporal concentration is measured as the maximum proportion of videos from any single year. Clusters with over 50\% concentration are flagged as potentially event-specific, which is important for understanding whether discovered visual frames represent persistent themes or temporal phenomena.

Pairwise cosine similarities between cluster centroids are also reported in the comparative analysis. For clusters $i$ and $j$ with centroids $\mathbf{c}_i$ and $\mathbf{c}_j$, the similarity is:
\begin{equation}
\text{sim}(i,j) = \frac{\mathbf{c}_i \cdot \mathbf{c}_j}{||\mathbf{c}_i|| \cdot ||\mathbf{c}_j||}
\end{equation}

Cluster pairs with similarity above 0.9 are flagged as potential merge candidates, indicating possible over-segmentation. The analysis also reports mean inter-cluster similarity, which indicates overall cluster separation, and maximum off-diagonal similarity, which identifies the most similar distinct clusters.

\section{Detailed Zero-shot Classification Results}
\label{app:ZS}

\subsection{Video-LLaVa}
\label{app:videollava}

\begin{figure}[ht]
    \centering
    \includegraphics[width=0.7\textwidth]{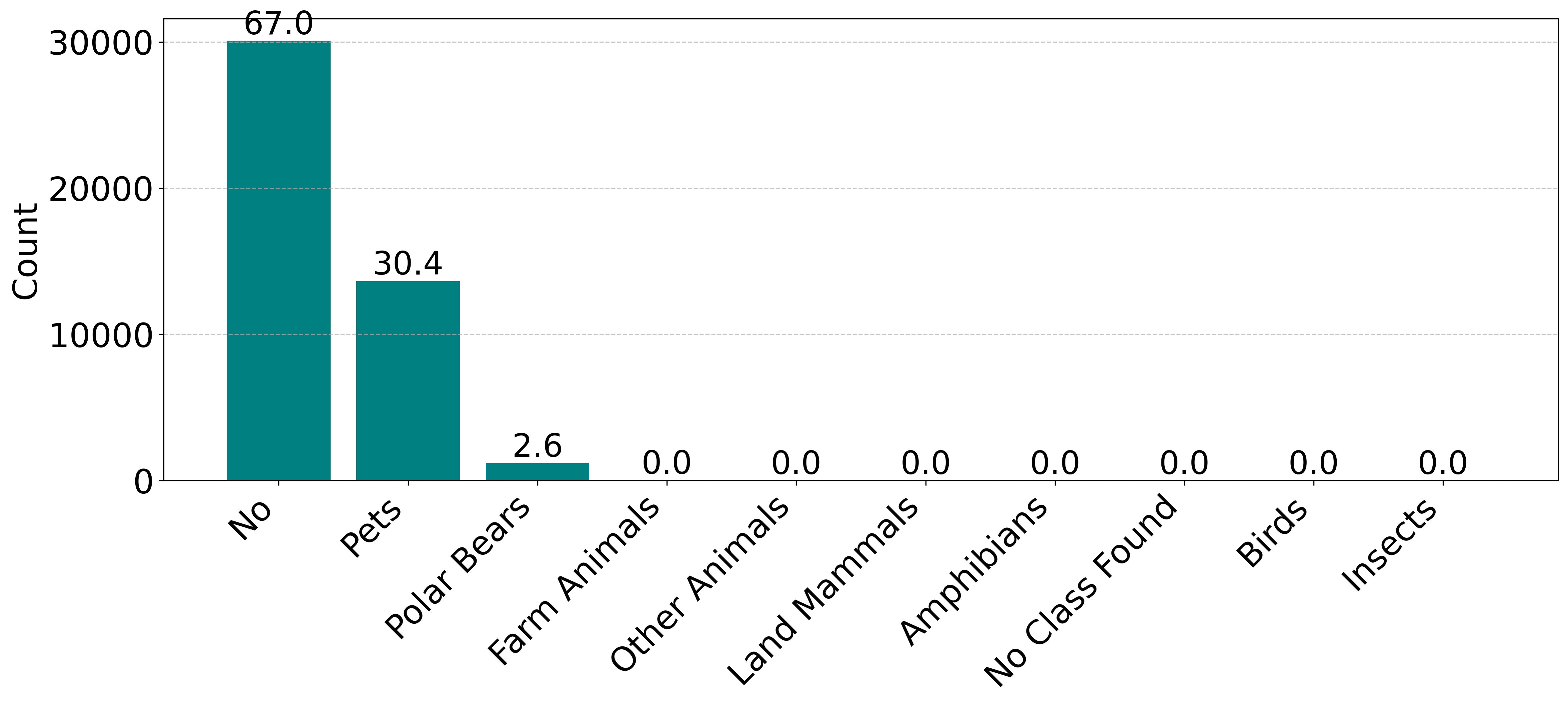}
    \caption{Video-LLaVA - Classification Results for the \textit{animals} group. 
Bar values indicate the relative frequency of each assigned category, based on a total of 44,927 videos.}       
    \label{fig:videollava_animals}
\end{figure}

\begin{figure}[ht]
    \centering
    \includegraphics[width=0.7\textwidth]{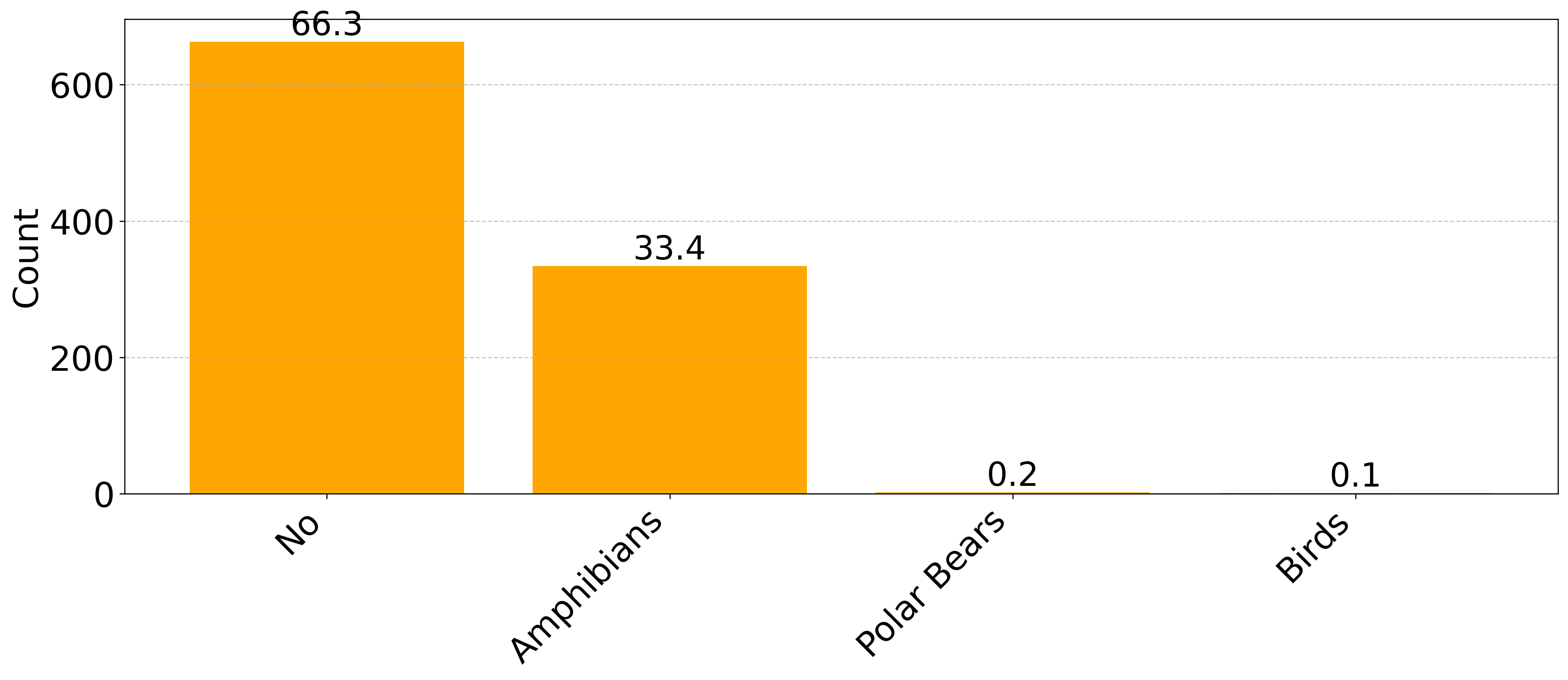}
    \caption{Video-LLaVA - Classification Results of the subset for the \textit{animals} group. 
Bar values indicate the relative frequency of each assigned category, based on a total of 1,000 videos.}
    \label{fig:videollava_animals_subset}
\end{figure}

As shown in \cref{fig:videollava_animals}, Video-LLaVA \cite{lin2024video} predicts the first category \textit{Pets} with a much higher frequency than the subsequent categories.
To investigate this potential bias, a secondary experiment was conducted using a subset
of 1,000 random videos with a shuffled version of the prompt, in which the category order was
randomized. 
The results of this test are displayed in \cref{fig:videollava_animals_subset} and confirms the presence of a positional bias, as the model
tends to favor whichever category is mentioned first.
This effect was not limited to the \textit{animals} group but is also observed in all other groups.


Both the original and shuffled prompts yielded a similar relative number of videos labeled as ``No” across the \textit{animals}, \textit{consequences} and \textit{climate action} group.
Given that ``No” was the second response option in the initial binary prompt, the absence of the bias towards the first option can be assumed for simpler or binary prompts.
To assess the consistency and faithfulness of the model’s classification for the ``No” category the overlap between the two results was calculated using the Jaccard Similarity.
This is defined as the ratio of videos labeled ``No” in both conditions to the total number of videos labeled ``No” in at least one condition. For the calculation only videos of the subset were used.
The resulting overlap rates are as follows:
\begin{itemize}
    \item Animals: 75.81\%
    \item Consequences: 49.89\%
    \item climate actions: 63.78\%
\end{itemize}

\newpage
\subsection{Video-ChatGPT}
\label{app:videochatgpt}
The results of Video-ChatGPT \cite{maaz2023video} contain unusual combinations of labels.
\subsubsection{Category Group: Animals}
\begin{figure}[ht]
    \centering
    \includegraphics[width=0.7\textwidth]{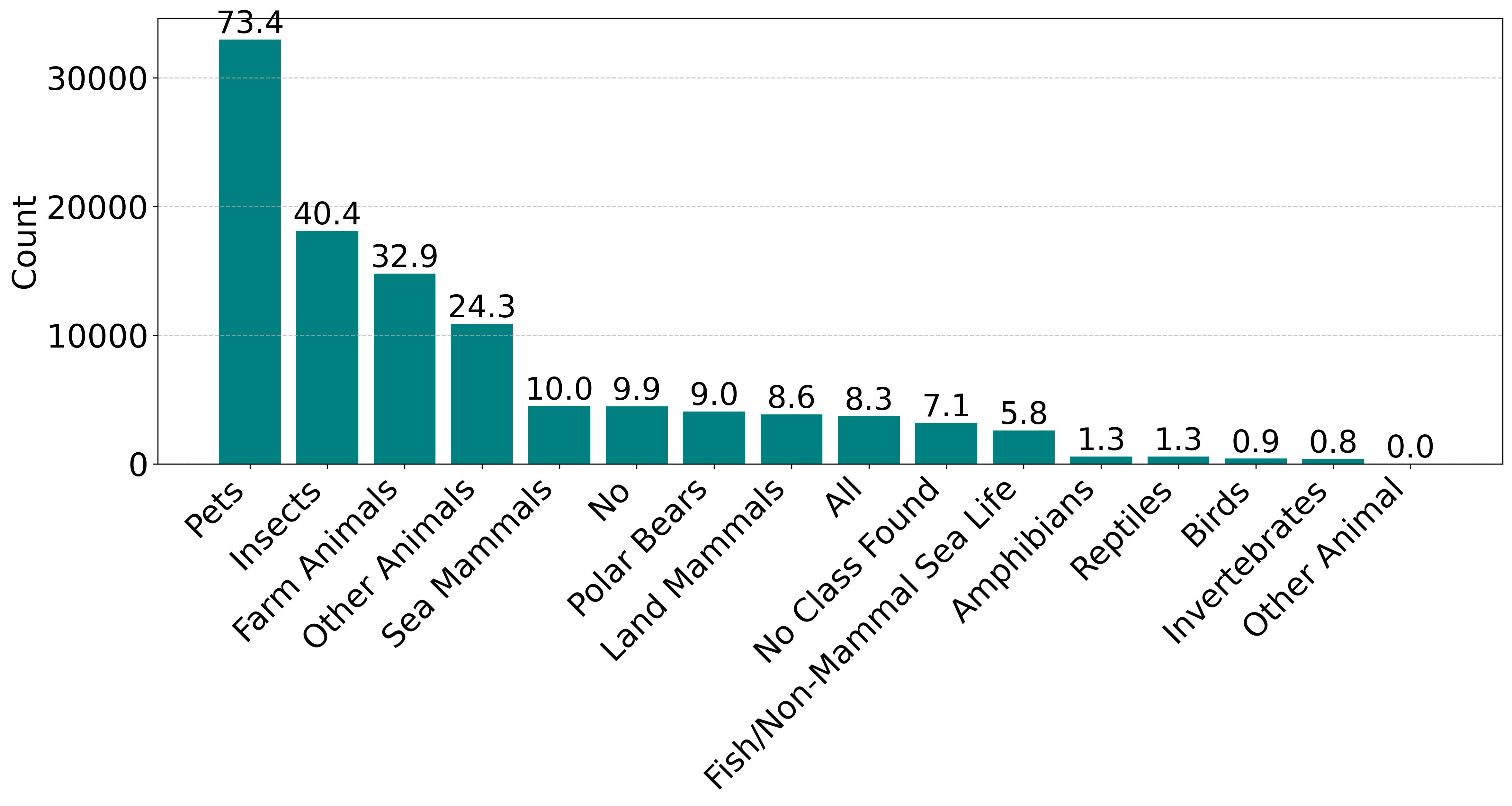}
    \caption{Video-ChatGPT - Classification Results for the \textit{animals} group.
Bar values indicate the relative frequency of each assigned category, based on a total of 44,927 videos.}       
    \label{fig:videochatgpt_animals}
\end{figure}

\begin{table}[ht]
    \centering
    \footnotesize
    \caption{Distribution of category combinations in the Animal group as classified by Video-ChatGPT. Only combinations occurring in more than 1\% of all cases are included.}
        \begin{tabularx}{\textwidth}{l|X}
        \toprule
        \textbf{Percentage} & \textbf{Category} \\ \hline
        37.9\% & Insects \& Pets \\ \hline
        22.3\% & Farm Animals \& Other Animals \& Pets \\ \hline
        9.9\%  & No \\ \hline
        8.2\%  & All \\ \hline
        7.1\%  & No Class Found \\ \hline
        3.6\%  & Farm Animals \& Fish/Non-Mammal Sea Life \& Land Mammals \& Pets \& Polar Bears \& Sea Mammals \\ \hline
        3.4\%  & Farm Animals \& Land Mammals \& Pets \& Polar Bears \& Sea Mammals \\ \hline
        2.6\%  & Pets \\
        \bottomrule
    \end{tabularx}
\label{tab:category_distribution_videochatgpt_animals}
\end{table}

\cref{fig:videochatgpt_animals} shows the results of the classification of the \textit{animal} group. It is immediately noticeable that more labels were assigned than there are videos, indicating that many videos received multiple categories. Along with the very similar assignment rates among categories such as \textit{polar bears}, \textit{sea mammals}, \textit{land mammals} and \textit{fish/non-mammal sea life} the implication of recurring classification patterns can be made. In addition, categories like \textit{pets}, \textit{insects}, and \textit{farm animals} were assigned at unusually high rates.

With the patterns shown in \cref{tab:category_distribution_videochatgpt_animals} it can be calculated that 75.4\% (37.9\% + 22.3\% + 8.2\% + 3.6\% + 3.4\%) of all videos fall into overly broad or implausible label combinations. Excluding the \textit{No Class Found} cases, this share increases to 81.16\% (75.4\% / [100\% - 7.1\%]).

This strong concentration of repetitive label combinations indicates low classification faithfulness. As a result, a trend analysis is not feasible.
\newpage
\subsubsection{Category Group: Consequences}
\begin{figure}[ht]
    \centering
    \includegraphics[width=0.7\textwidth]{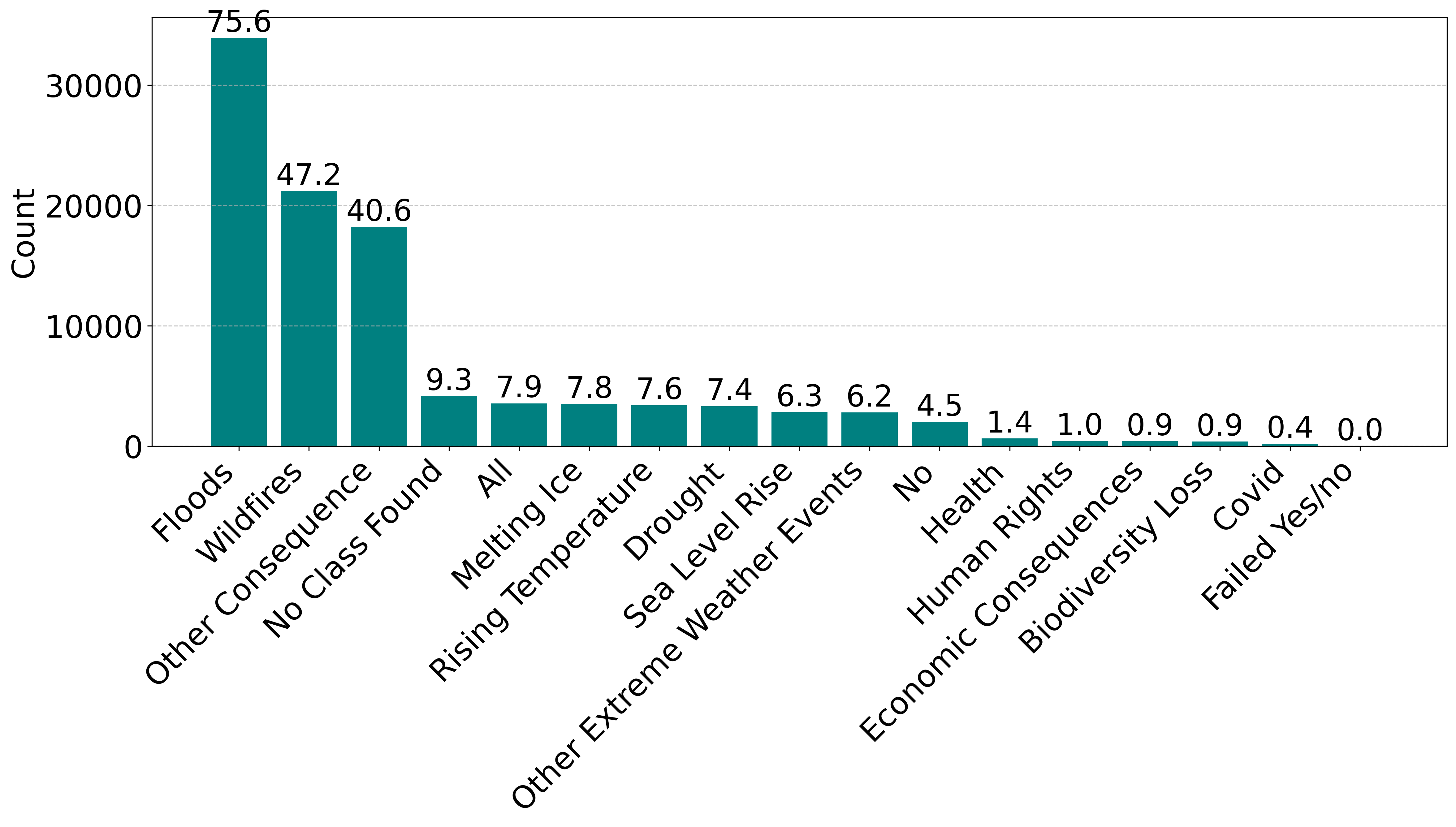}
    \caption{Video-ChatGPT - Classification Results for the \textit{consequence} group.
Bar values indicate the relative frequency of each assigned category, based on a total of 44,927 videos.}       
    \label{fig:videochatgpt_consequences}
\end{figure}

\begin{table}[ht]
    \footnotesize
    \centering
    \caption{Distribution of category combinations in the \textit{consequences} group as classified by Video-ChatGPT. Only combinations occurring in more than 1\% of all cases are included.}
    \begin{tabularx}{\textwidth}{l|X}
        \toprule
        \textbf{Percentage} & \textbf{Category} \\ \hline
        40.5\% & Floods \& Other Consequence \& Wildfires \\ \hline
        27.0\% & Floods \\ \hline
        9.3\%  & No Class Found \\ \hline
        7.9\%  & All \\ \hline
        5.3\%  & Drought \& Floods \& Melting Ice \& Other Extreme Weather Events \& Rising Temperature \& Sea Level Rise \& Wildfires \\ \hline
        4.5\%  & No \\ \hline
        1.2\%  & Rising Temperature \\ \hline
        1.0\%  & Melting Ice \\
        \bottomrule
    \end{tabularx}
    \label{tab:category_distribution_videochatgpt_consequences}
\end{table}

A strong bias toward the categories \textit{floods}, \textit{wildfires}, and \textit{other consequences} is evident in \cref{fig:videochatgpt_consequences} and further confirmed by the patterns in \cref{tab:category_distribution_videochatgpt_consequences}. This data also explains the similar percentages for categories like \textit{melting ice}, \textit{rising temperature}, \textit{drought}, \textit{sea level rise}, and \textit{other extreme weather events}. As a result of this lean towards a few dominant categories and the frequent appearance of unlikely category combinations, a trend analysis can be considered non-representative.
\newpage
\subsubsection{Category Group: Climate Action}
\begin{figure}[ht]
    \centering
    \includegraphics[width=0.7\textwidth]{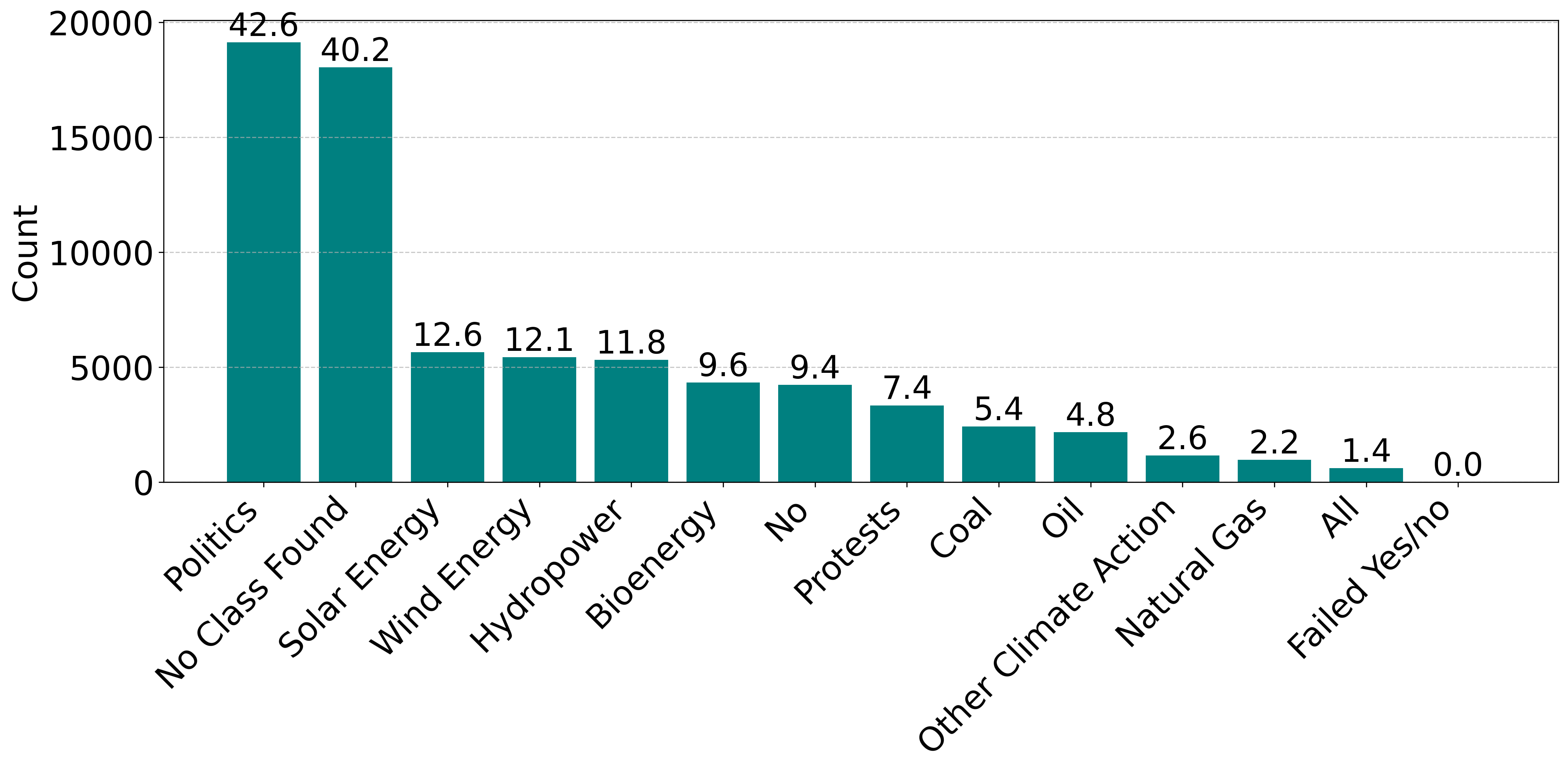}
    \caption{Video-ChatGPT - Classification Results for the \textit{climate action} group. 
Bar values indicate the relative frequency of each assigned category, based on a total of 44,927 videos.}       
    \label{fig:videochatgpt_climateaction}
\end{figure}

\begin{table}[ht]
    \centering
    \footnotesize
    \caption{Distribution of category combinations in the \textit{climate action} group as classified by Video-ChatGPT. Only combinations occurring in more than 1\% of all cases are included.}
    \begin{tabularx}{\textwidth}{l|X}
        \toprule
        \textbf{Percentage} & \textbf{Category} \\ \hline
        40.2\% & No Class Found \\ \hline
        34.0\% & Politics \\ \hline
        9.4\%  & No \\ \hline
        3.2\%  & Bioenergy \& Coal \& Hydropower \& Oil \& Politics \& Protests \& Solar Energy \& Wind Energy \\ \hline
        2.2\%  & Bioenergy \& Hydropower \& Politics \& Protests \& Solar Energy \& Wind Energy \\ \hline
        1.9\%  & Hydropower \& Solar Energy \& Wind Energy \\ \hline
        1.5\%  & Other Climate Action \& Politics \\ \hline
        1.5\%  & Bioenergy \& Hydropower \& Solar Energy \& Wind Energy \\ \hline
        1.3\%  & All \\
        \bottomrule
    \end{tabularx}
    \label{tab:category_distribution_videochatgpt_climateaction}
\end{table}

\begin{figure}[ht]
    \centering
    \includegraphics[width=0.7\textwidth]{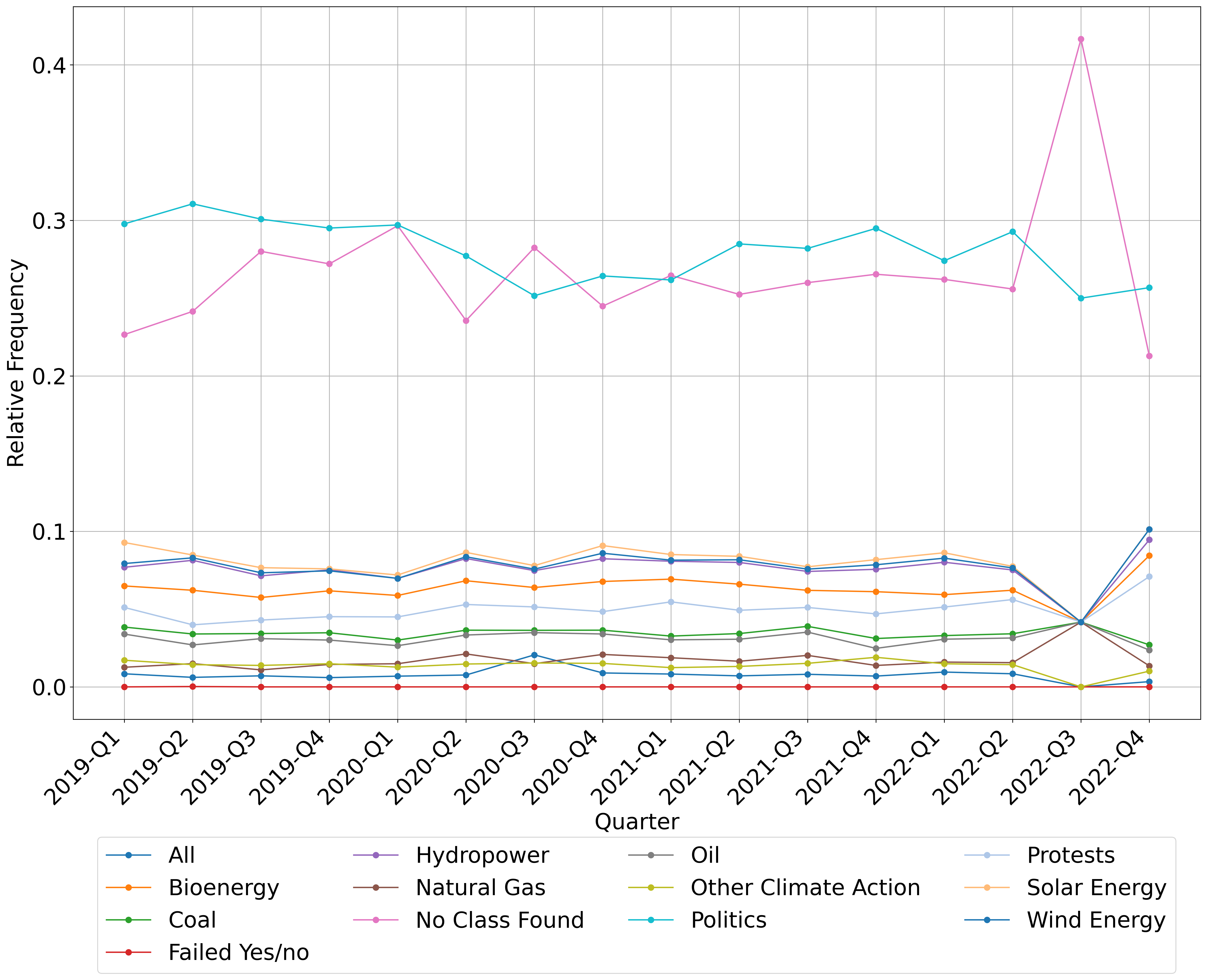}
    \caption{Video-ChatGPT: Trend of \textit{climate action}}       
    \label{fig:videochatgpt_climateaction_trend}
\end{figure}

For the group \textit{climate action}, Video-ChatGPT reveals several notable patterns. As shown in \cref{fig:videochatgpt_climateaction}, 40.2\% of all videos were labeled as \textit{No Class Found}, despite the model initially identifying them as related to \textit{climate action}. 
Additionally, 42.6\% of the videos appear to be \textit{Politics}. This means that of the correctly labeled videos 71.24\% (42.6\%/[100\% - 40.2\%]) got classified as \textit{Politics}.

Each renewable energy source such as \textit{Bioenergy}, \textit{Hydropower}, \textit{Solar Energy} and \textit{Wind Energy} shows a relatively uniform distribution across the dataset. These observations are supported by the data in \cref{tab:category_distribution_videochatgpt_climateaction}.
\newpage
In the trend graph in \cref{fig:videochatgpt_climateaction_trend}, \textit{No Class Found} and \textit{Politics} have by far the highest relative frequency in contrast to all other categories.
Categories that have similar trends over time are renewable energy categories except for \textit{Bioenergy}. Other categories that move closely together are \textit{Coal} and \textit{Oil}. It is important to mention that in 2022-Q3 a lot of categories merge to the same relative frequency, while \textit{No Class Found} has a clear spike. 
\newpage
\subsubsection{Category Group: Setting}
\begin{figure}[ht]
    \centering
    \includegraphics[width=0.7\textwidth]{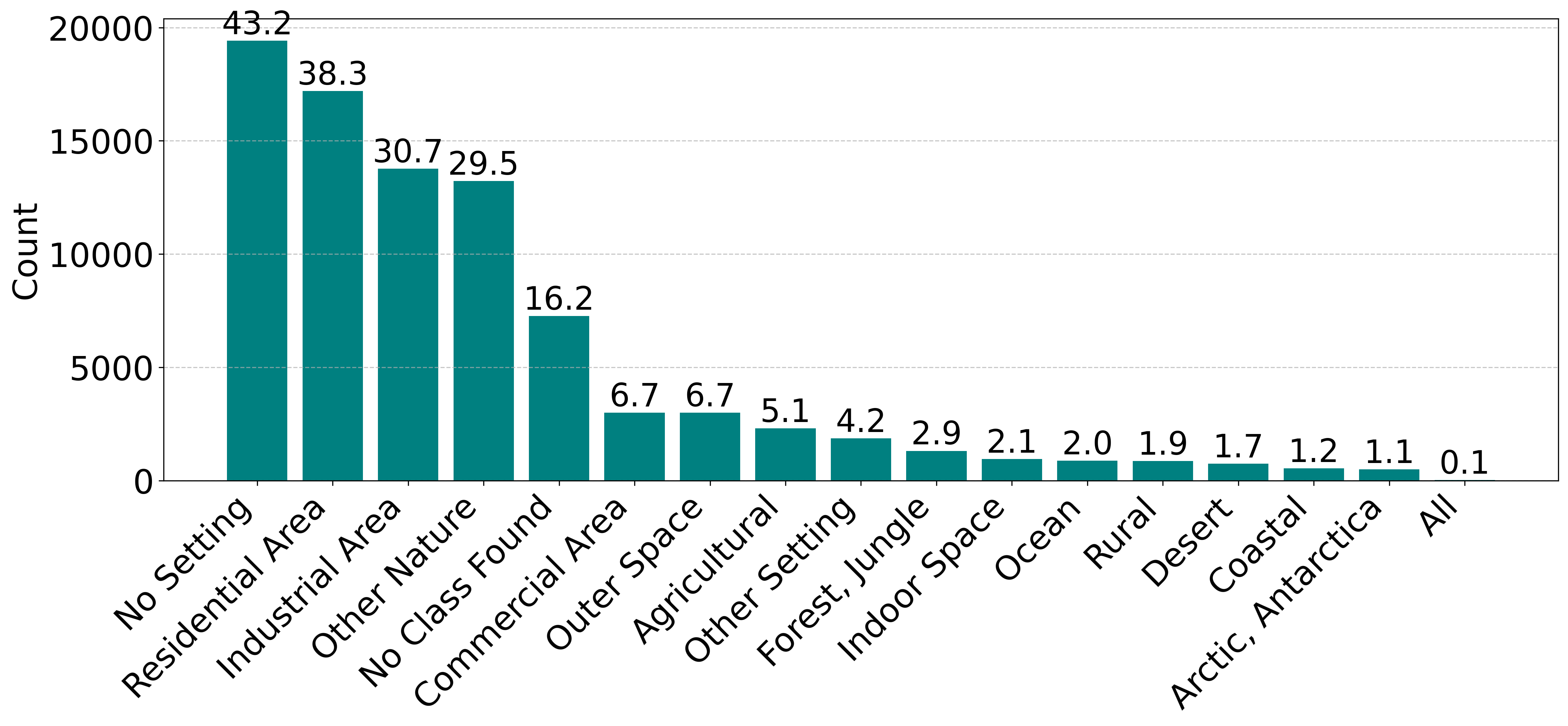}
    \caption{Video-ChatGPT - Classification Results for the \textit{setting} group. 
Bar values indicate the relative frequency of each assigned category, based on a total of 44,927 videos.}       
    \label{fig:videochatgpt_setting}
\end{figure}

\begin{table}[ht]
   \centering
   \footnotesize
    \caption{Distribution of category combinations in the \textit{setting} group as classified by Video-ChatGPT. Only combinations occurring in more than 1\% of all cases are included.}
    \begin{tabularx}{\textwidth}{l|X}
        \toprule
        \textbf{Percentage} & \textbf{Category} \\ \hline
        35.9\% & Residential Area \\ \hline
        28.4\% & Industrial Area \& No Setting \& Other Nature \\ \hline
        16.2\% & No Class Found \\ \hline
        5.4\%  & Commercial Area \& No Setting \& Outer Space \\ \hline
        3.8\%  & Agricultural \& No Setting \& Other Setting \\ \hline
        2.3\%  & No Setting \\ \hline
        1.1\%  & Indoor Space \\
        \bottomrule
    \end{tabularx}
    \label{tab:category_distribution_videochatgpt_setting}
\end{table}

The results of classifying the \textit{setting} group (\cref{fig:videochatgpt_setting}) reveal that several categories appear with substantially higher relative frequencies than others. 
Given that all videos were selected based on the topic of climate-change and sourced from social media, the prominence of certain categories is to some extent expected. That said, in this case the imbalance of e.g. \textit{Residential Area} is excessive.
Furthermore, examining the distribution of categories assigned to individual videos (\cref{tab:category_distribution_videochatgpt_setting}) reveals questionable pairings of categories. For instance, it is highly unlikely that 28.4\% of all videos simultaneously feature an \textit{industrial area}, \textit{no setting} and \textit{other nature}.  These and similar contradictions within the data, as well as a possible bias towards \textit{residential area} indicates that a reliable trend analysis is not possible.
\newpage
\subsubsection{Category Group: Type}
\begin{figure}[ht]
    \centering
    \includegraphics[width=0.7\textwidth]{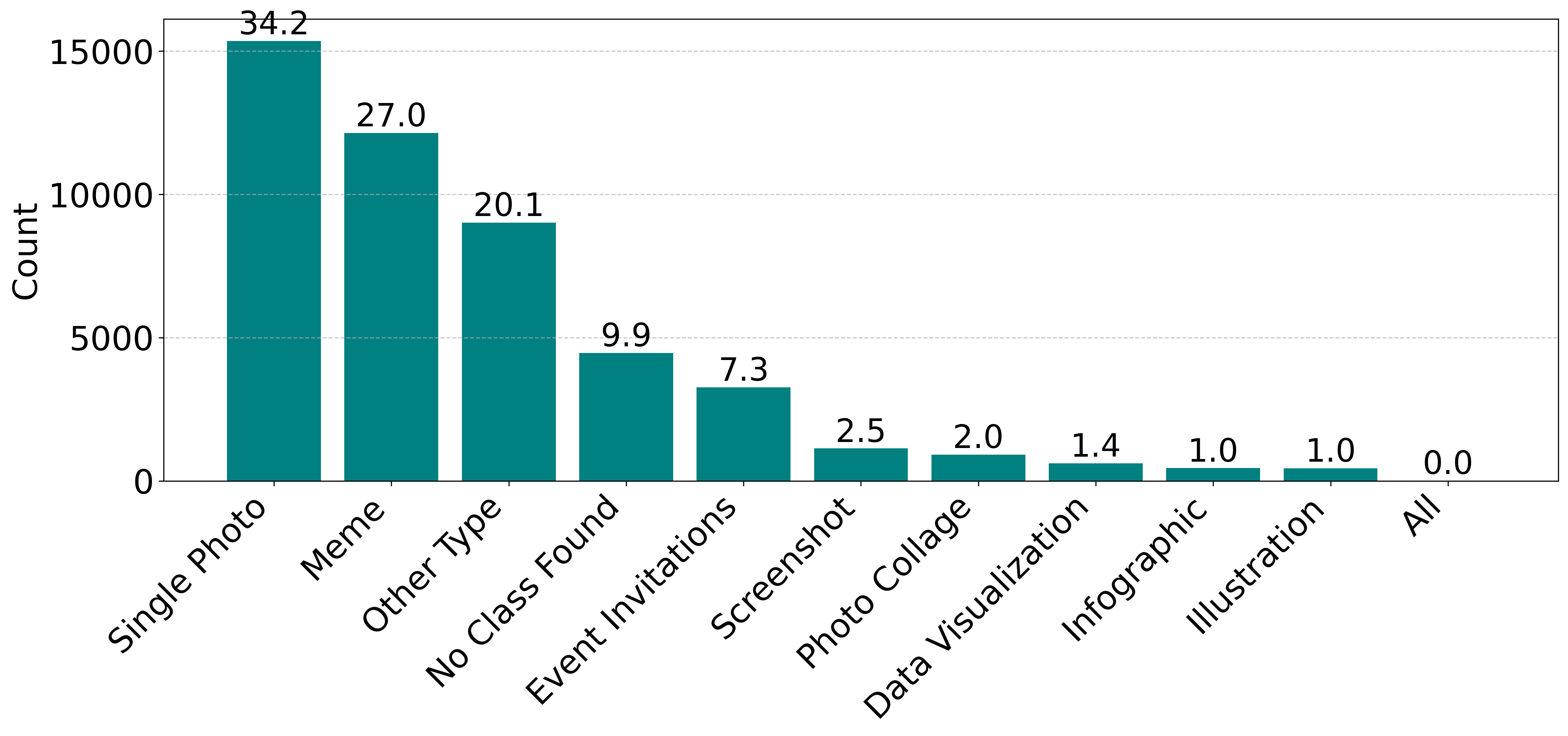}
    \caption{Video-ChatGPT - Classification Results for the \textit{type} group.
Bar values indicate the relative frequency of each assigned category, based on a total of 44,927 videos.}       
    \label{fig:videochatgpt_type}
\end{figure}

\begin{table}[ht]
    \footnotesize
    \centering
    \caption{Distribution of category combinations in the \textit{type} group as classified by Video-ChatGPT. Only combinations occurring in more than 1\% of all cases are included.}
    \begin{tabularx}{\textwidth}{l|X}
       \toprule
        \textbf{Percentage} & \textbf{Category} \\ \hline
        33.1\% & Single Photo \\ \hline
        26.0\% & Meme \\ \hline
        19.2\% & Other Type \\ \hline
        9.9\%  & No Class Found \\ \hline
        6.6\%  & Event Invitations \\ \hline
        1.7\%  & Screenshot \\ \hline
        1.3\%  & Photo Collage \\
        \bottomrule
    \end{tabularx}
    \label{tab:category_distribution_videochatgpt_type}
\end{table}

The type classification does not show signs of substantial randomness or clear indicators of errors (\cref{fig:videochatgpt_type}). However, it is notable that 34.2\% of videos were assigned the category \textit{Single Photo}, which is unusual given that the dataset consists of videos. A review of the data in \cref{tab:category_distribution_videochatgpt_type} does not reveal any obvious mistakes.
\newpage
\begin{figure}[ht]
    \centering
    \includegraphics[width=0.7\textwidth]{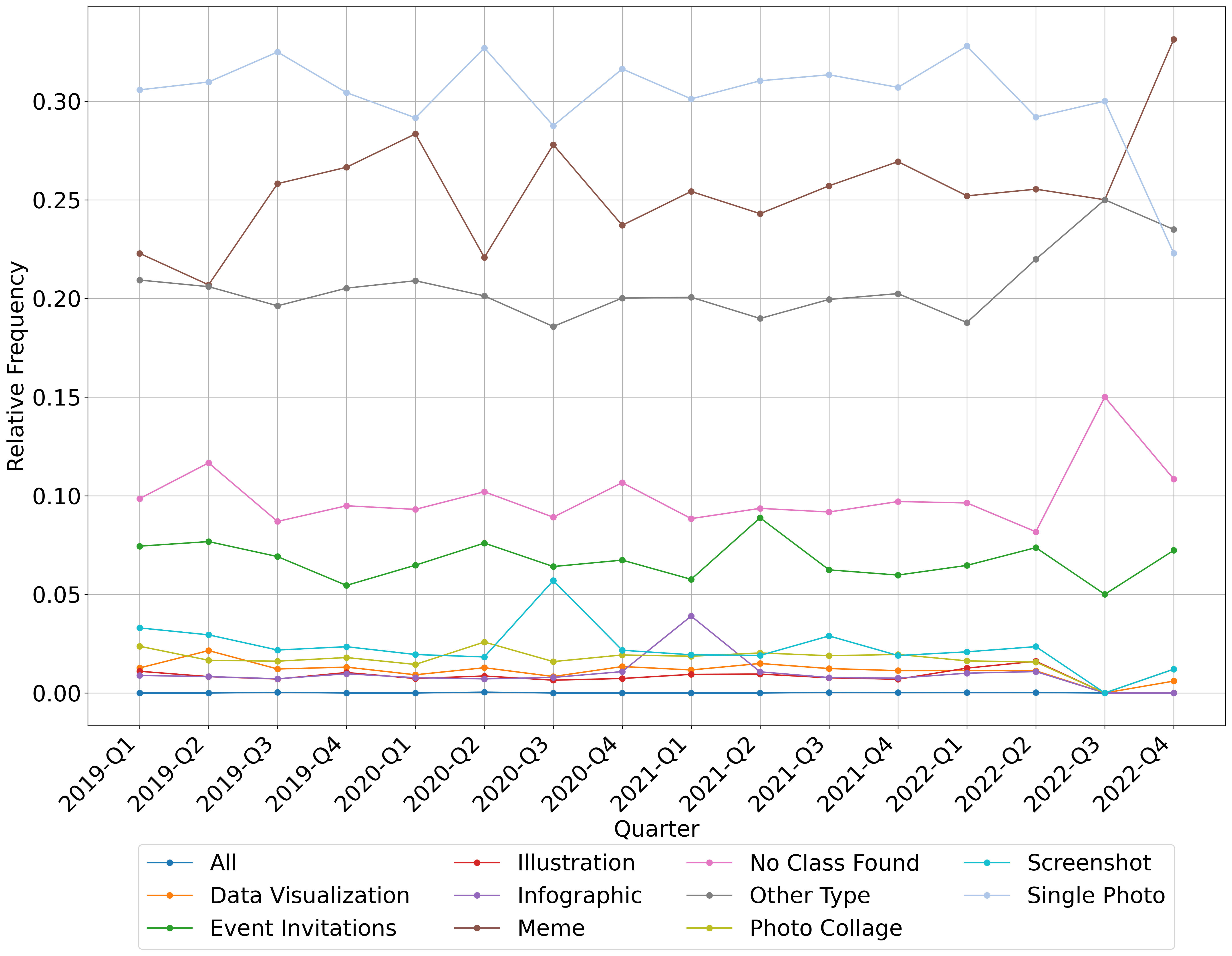}
    \caption{Video-ChatGPT: Trend of \textit{type}}       
    \label{fig:videochatgpt_type_trend}
\end{figure}

The Trend Graph \ref{fig:videochatgpt_type_trend} reveals three distinct groups of categories based on the relative frequency. The most dynamic categories, also having the highest relative frequency, are \textit{single photo}, \textit{meme}, and \textit{other type}. 
In contrast, \textit{event invitations} and \textit{No Class Found} demonstrate less volatility, but maintaining a relative frequency between 5\% and 15\%.
Of particular interest is the period of 2022-Q3, where a significant increase in volatility across many video categories is observed. Especially the categories with the fewest assigned videos and volatility (\textit{All}, \textit{data visualization}, \textit{Illustration}, \textit{infographic}, \textit{photo collage}, and \textit{screenshot}) all register a drop to 0\% in this specific quarter.

\newpage
\subsection{PandaGPT}
\label{app:pandagpt}

PandaGPT \cite{su2023pandagpt} contains implausible combinations of labels for all categories except for \textit{type}.

\subsubsection{Category Group: Animals \& Consequences}
\begin{figure}[ht]
    \centering
    \includegraphics[width=0.7\textwidth]{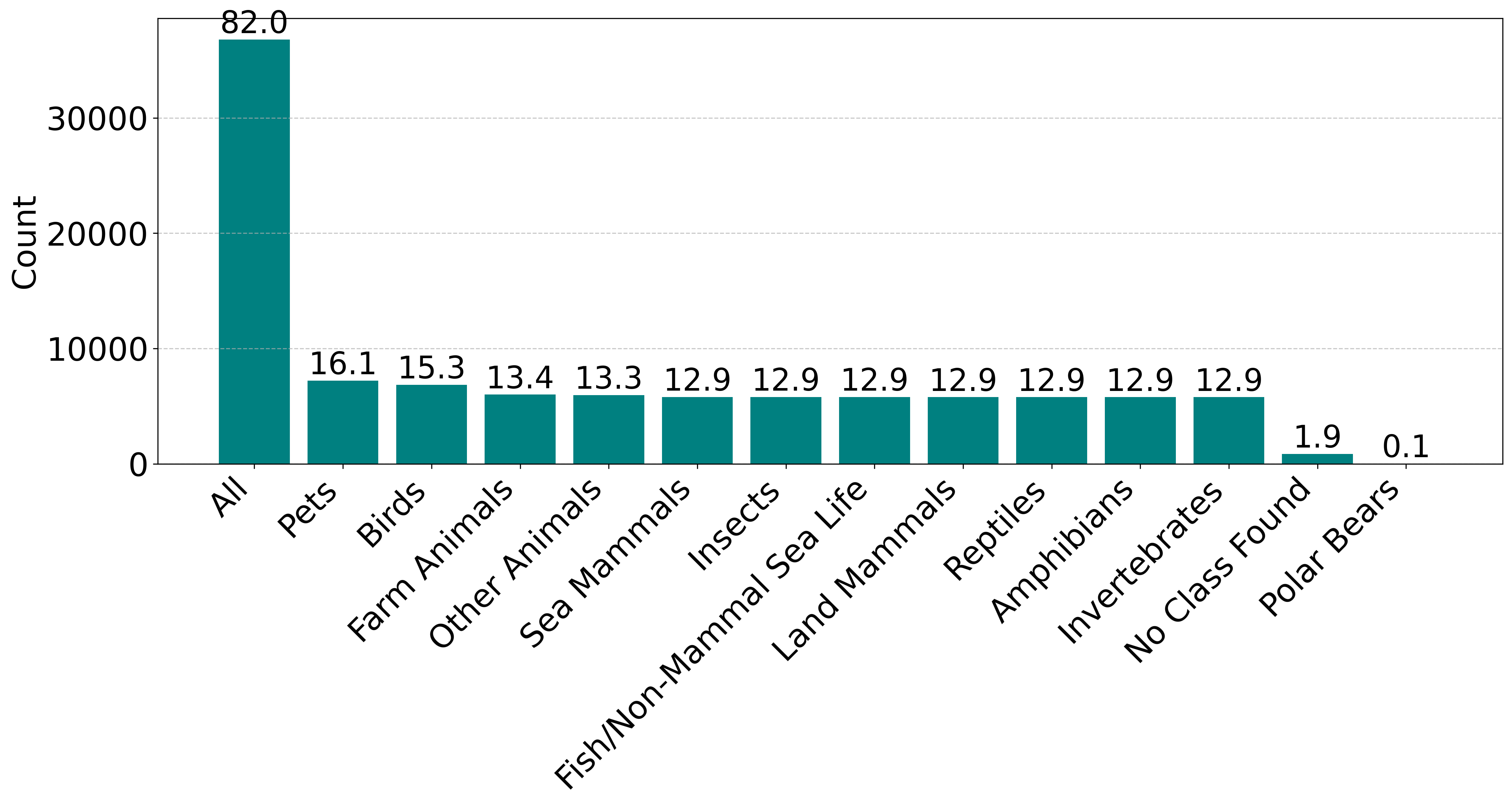}
    \caption{PandaGPT - Classification Results for the \textit{animals} group. 
Bar values indicate the relative frequency of each assigned category, based on a total of 44,927 videos.}       
    \label{fig:pandagpt_animals}
\end{figure}

\begin{table}[ht]
    \centering
    \footnotesize
    \caption{Distribution of category combinations in the Animal group as classified by PandaGPT. Only combinations occurring in more than 1\% of all cases are included.}
    \begin{tabularx}{\textwidth}{l|X}
      \toprule
      \textbf{Percentage} & \textbf{Category} \\ \hline
      82.0\% & All \\ \hline
      12.8\% & Pets \& Farm Animals \& Land Mammals \& Sea Mammals \& Fish/Non-Mammal Sea Life \&Amphibians \& Reptiles \& Invertebrates \& Birds \& Insects \& Other Animals \\ \hline
      2.4\% & Pets \& Birds \\ \hline
      1.9\% & No Class Found \\
      \bottomrule
  \end{tabularx}

\label{tab:category_distribution_pandagpt_animal}
\end{table}

\cref{fig:pandagpt_animals} illustrates a relatively uniform distribution across most categories, with the notable exceptions of \textit{All}, \textit{No Class Found}, and \textit{Polar Bears}. Furthermore, the category \textit{All} is present in 82.0\% of all classified videos, a finding matched by the data presented in \cref{tab:category_distribution_pandagpt_animal}.

\begin{figure}[ht]
    \centering
    \includegraphics[width=0.7\textwidth]{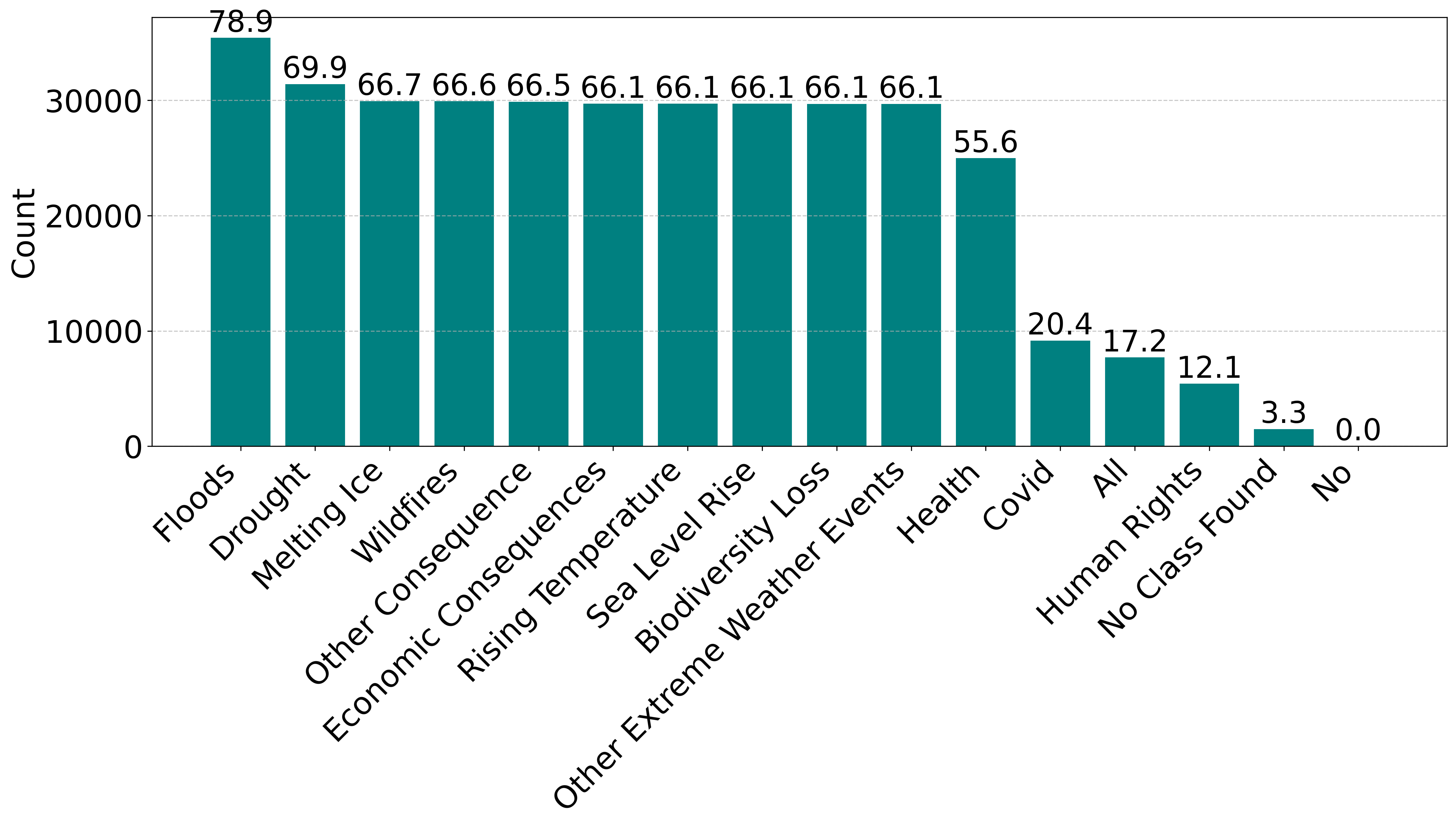}
    \caption{PandaGPT - Classification Results of the subset for the \textit{consequences} group. Bar values indicate the relative frequency of each assigned category, based on a total of 44,927 videos.}       
    \label{fig:pandagpt_consequences}
\end{figure}

Given the common use of the unrealistic classification \textit{All} and the fact that 12.8\% of all videos in the Animal group were assigned to all categories (excluding \textit{Polar Bears}), a meaningful trend analysis for this group becomes unfeasible.
Furthermore, it's crucial to note that the \textit{All} category inherently lacks viability for multicategory classification. When excluding \textit{Polar Bears}, 94.8\% of videos were assigned every other category, rendering these results invalid for trend analysis.
\newpage
PandaGPT exhibits similar issues of uniform distribution within the \textit{consequences} group as shown in \cref{fig:pandagpt_consequences} and \cref{tab:category_distribution_pandagptgpt_consequences}.

\begin{table}[ht]
  \centering
  \footnotesize
  \caption{Distribution of category combinations in the \textit{consequences} group as classified by PandaGPT. Only combinations occurring in more than 1\% of all cases are included.}
    \begin{tabularx}{\textwidth}{l|X}
    \toprule
    \textbf{Percentage} & \textbf{Category} \\ \hline
    51.8\% & Floods \& Drought \& Wildfires \& Rising Temperature \& Other Extreme Weather Events \& Melting Ice \& Sea Level Rise \& Economic Consequences \& Biodiversity Loss \& Health \& Other Consequence \\ \hline
    17.2\% & All \\ \hline
    12.1\% & Floods \& Drought \& Wildfires \& Rising Temperature \& Other Extreme Weather Events \& Melting Ice \& Sea Level Rise \& Human Rights \& Economic Consequences \& Biodiversity Loss \& Covid \& Other Consequence \\ \hline
    8.3\% & Floods \& Covid \\ \hline
    3.8\% & Floods \& Drought \& Health \\ \hline
    3.3\% & No Class Found \\ \hline
    2.2\% & Floods \& Drought \& Wildfires \& Rising Temperature \& Other Extreme Weather Events \& Melting Ice \& Sea Level Rise \& Economic Consequences \& Biodiversity Loss \& Other Consequence \\
    \bottomrule
  \end{tabularx}
\label{tab:category_distribution_pandagptgpt_consequences}
\end{table}
\newpage

\subsubsection{Category Group: Climate Action}
\begin{figure}[ht]
    \centering
    \includegraphics[width=0.7\textwidth]{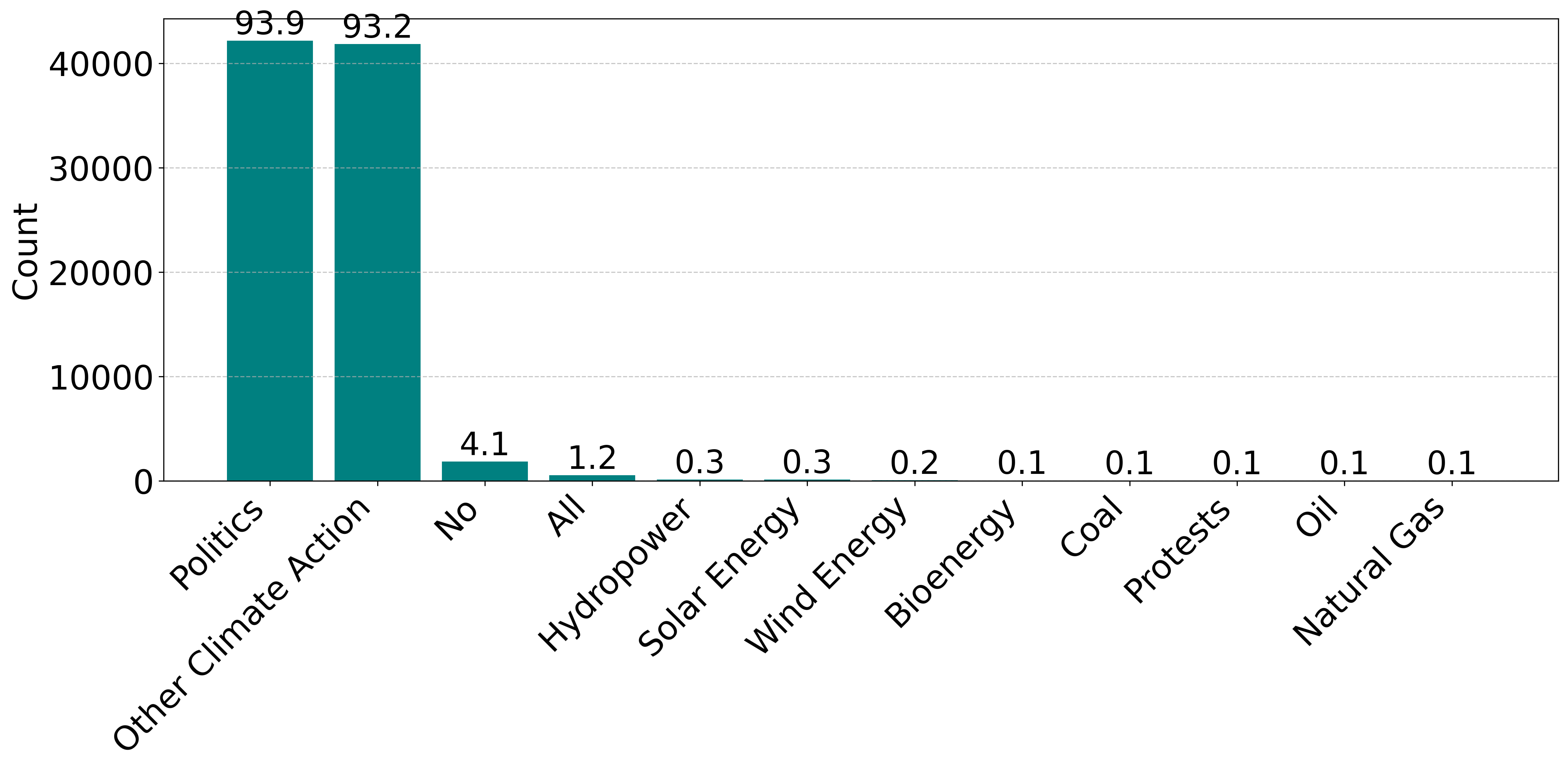}
    \caption{PandaGPT - Classification Results for the \textit{climate action} group. 
Bar values indicate the relative frequency of each assigned category, based on a total of 44,927 videos.}       
    \label{fig:pandagpt_climateaction}
\end{figure}

\begin{figure}[ht]
    \centering
    \includegraphics[width=0.7\textwidth]{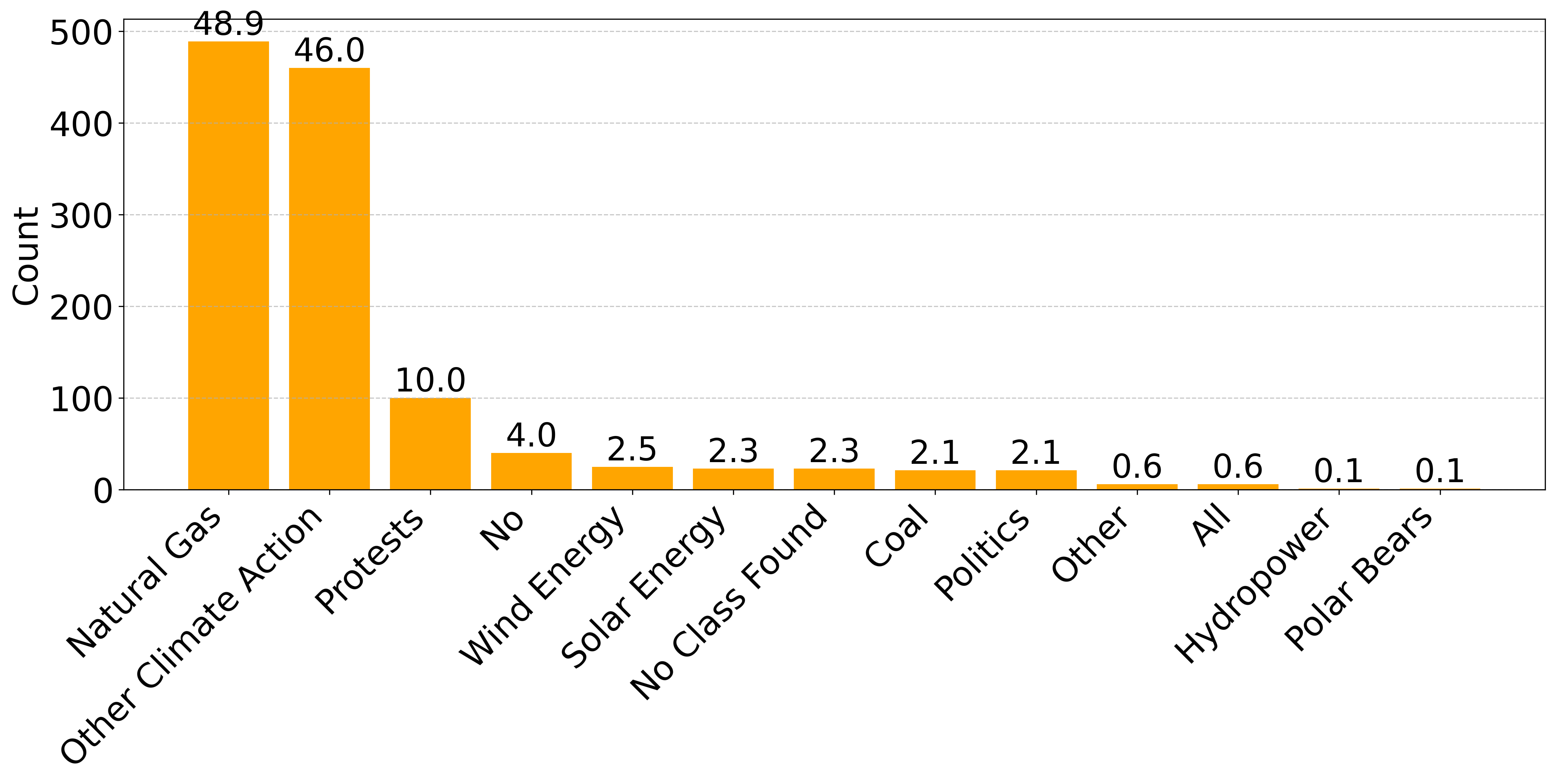}
    \caption{PandaGPT - Classification Results of the subset for the \textit{climate action} group. 
Bar values indicate the relative frequency of each assigned category, based on a total of 1,000 videos.}       
    \label{fig:pandagpt_climateaction_subset}
\end{figure}

The way videos are categorized in the \textit{climate action} group strongly favors \textit{Politics} and \textit{Other Climate Action} (\cref{fig:pandagpt_climateaction}). This likely happened because these were the first and last categories listed in the prompt. For confirmation the order of the categories in the prompt were shuffled. The results are displayed in \cref{fig:pandagpt_climateaction_subset}.
It seems like that always the first category in the prompt as well as \textit{Other Climate Action}, no matter the position, is most frequently chosen. 
The observed bias makes a valid trend analysis impossible.
\newpage
\subsubsection{Category Group: Setting}
\begin{figure}[ht]
    \centering
    \includegraphics[width=0.7\textwidth]{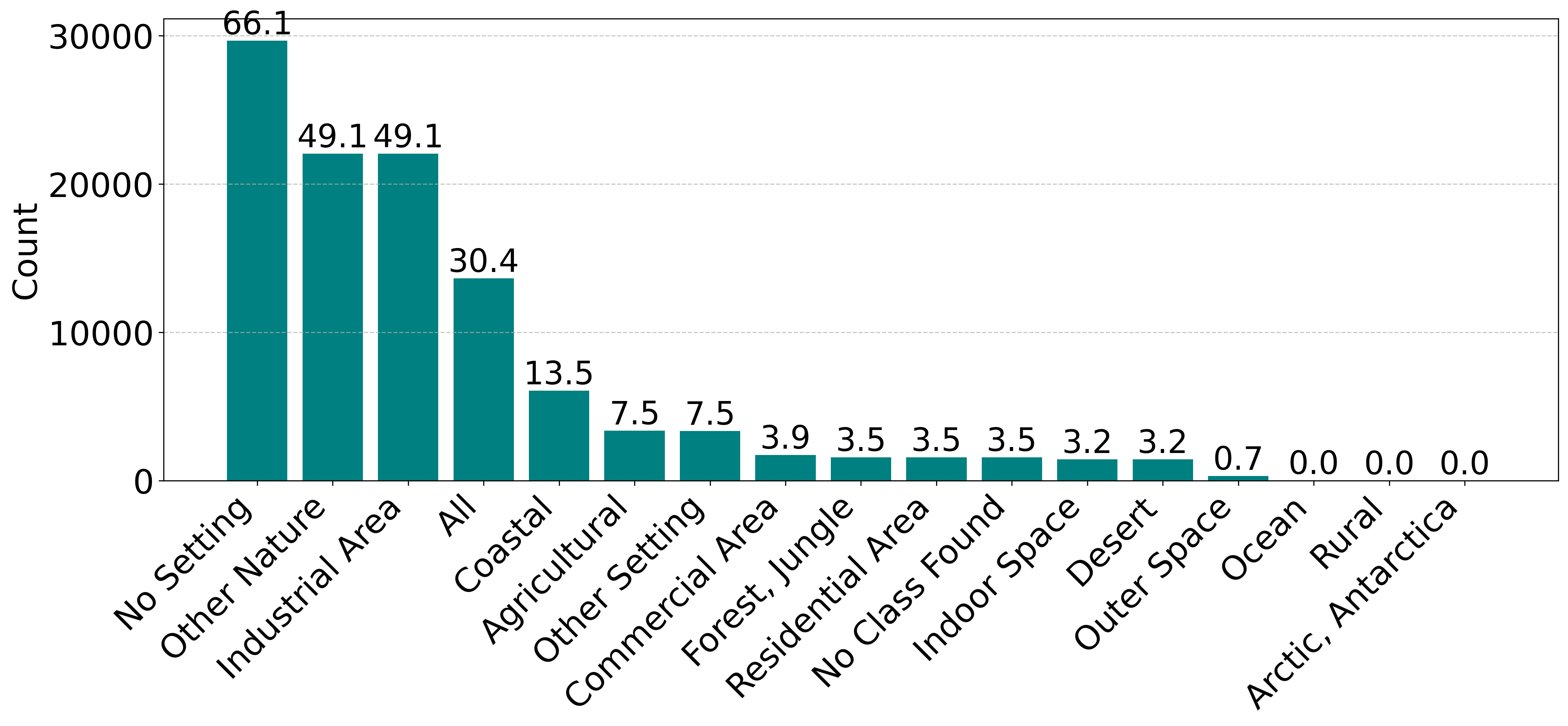}
    \caption{PandaGPT - Classification Results for the \textit{setting} group. 
Bar values indicate the relative frequency of each assigned category, based on a total of 44,927 videos.}       
    \label{fig:pandagpt_setting}
\end{figure}

\begin{table}[ht]
  \centering
  \footnotesize
  \caption{Distribution of category combinations in the \textit{setting} group as classified by PandaGPT. Only combinations occurring in more than 1\% of all cases are included.}
    \begin{tabularx}{\textwidth}{l|X}
    \toprule
    \textbf{Percentage} & \textbf{Category} \\ \hline
    45.9\% & No Setting \& Industrial Area \& Other Nature \\ \hline
    30.4\% & All \\ \hline
    10.3\% & No Setting \& Coastal \\ \hline
    4.3\% & No Setting \& Agricultural \& Other Setting \\ \hline
    3.5\% & No Class Found \\ \hline
    3.2\% & No Setting \& Residential Area \& Industrial Area \& Commercial Area \& Agricultural \& Indoor Space \& Coastal \& Desert \& Forest, Jungle \& Other Nature \& Other Setting \\ \hline
    1.5\% & No Setting \\
    \bottomrule
  \end{tabularx}
\label{tab:category_distribution_pandagpt_setting}
\end{table}

Looking at the results of classifying the \textit{setting} group, several observations can be made (\cref{fig:pandagpt_setting}).
First, the category \textit{no setting} has a high count with 66.1\%. Second, \textit{other nature} and \textit{industrial area} appear with the same frequency. 
Furthermore, there's an unusual high number of videos classified as \textit{all} within the \textit{setting} group.
Additionally, \textit{Agricultural} and \textit{other setting} show a similar number of assignments, as do \textit{commercial area}, \textit{forest or jungle}, \textit{residential area}, \textit{no class found}, \textit{indoor space}, \textit{desert}, and \textit{outer space}.
Given these odd groupings within the \textit{setting} group and considering the data from \cref{tab:category_distribution_pandagpt_setting}, we conclude that the model favors certain categories and category combinations.
A meaningful trend analysis is thus not useful.
\newpage
\subsubsection{Category Group: Type}
\begin{figure}[ht]
    \centering
    \includegraphics[width=0.7\textwidth]{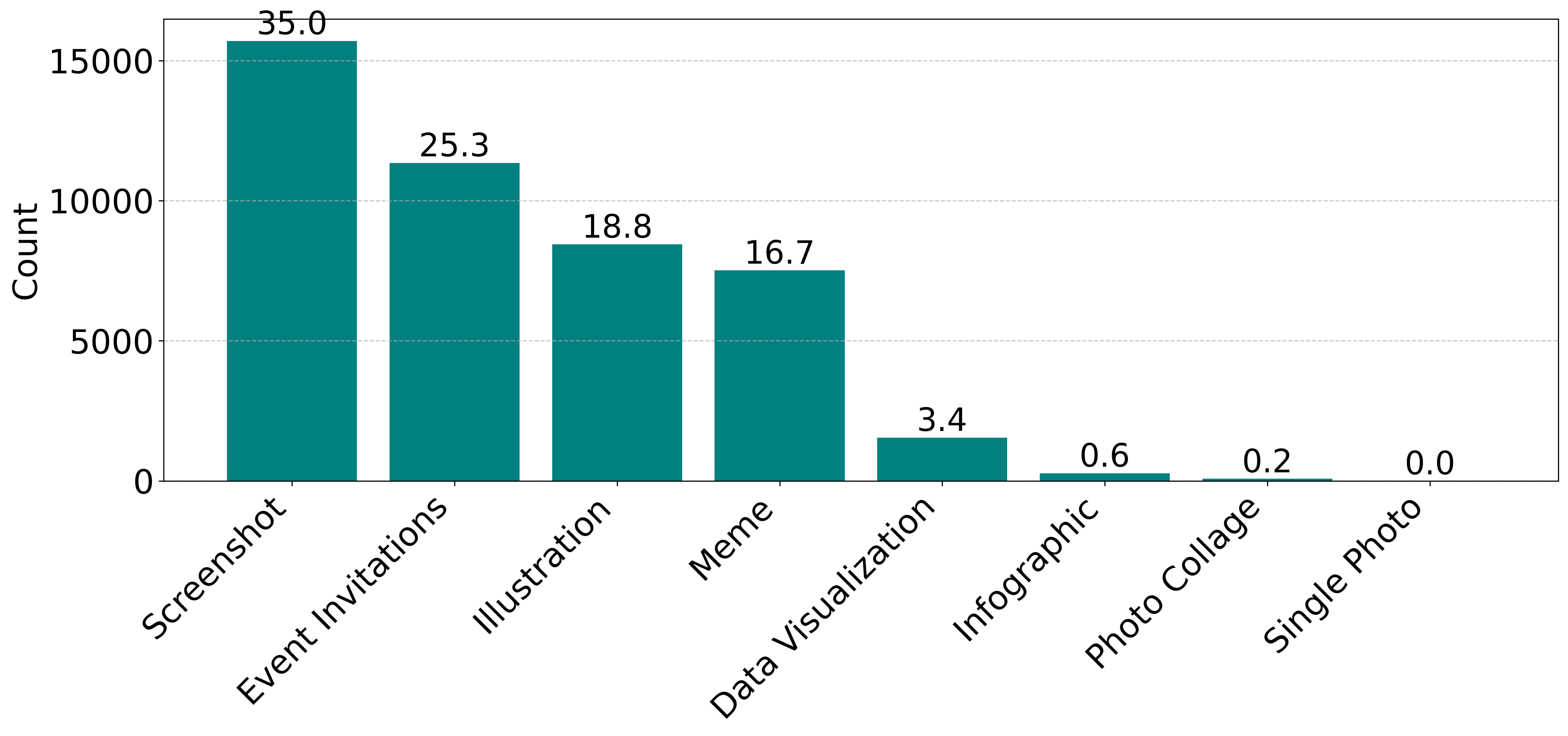}
    \caption{PandaGPT - Classification Results for the \textit{type} group. 
Bar values indicate the relative frequency of each assigned category, based on a total of 44,927 videos.}       
    \label{fig:pandagpt_type}
\end{figure}

\cref{fig:pandagpt_type} does not reveal any immediately obvious abnormalities. 
While the high number of videos classified as \textit{Screenshot} and \textit{Illustration} might seem unusual, it does not necessarily indicate a significant issue.

\begin{figure}[ht]
    \centering
    \includegraphics[width=0.7\textwidth]{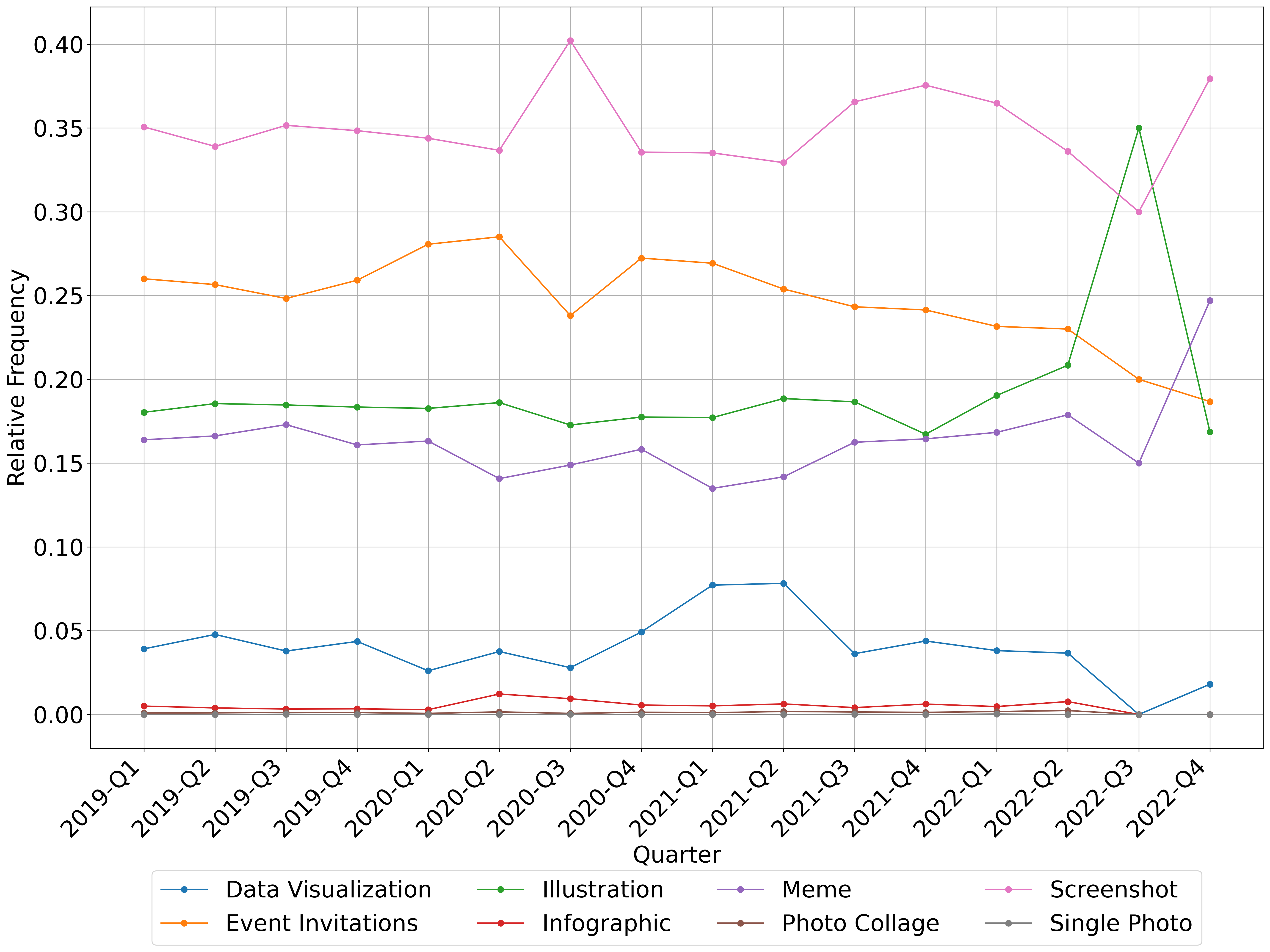}
    \caption{PandaGPT: Trend of Type}       
    \label{fig:pandagpt_type_trend}
\end{figure}

The Trend Graph (\cref{fig:pandagpt_type_trend}) shows some degree of volatility across all categories with a relative frequency exceeding 2.5\%, with the most pronounced fluctuations occurring in 2022-Q3. 
Notably, no clear, consistent trends are evident across different categories.

\newpage
\subsection{CLIP}
\label{app:clip}
Since the results from CLIP \cite{radford2021learning} appear to be the most reliable, each group includes a bar chart, as well as a line graph illustrating the trend of each category over time. The bar chart is based solely on the unique dataset and therefore reflects only unique videos. In contrast, the Trend Graph includes both unique and duplicate videos.

\subsubsection{Category Group: Animals}
\begin{figure}[ht]
    \centering
    \includegraphics[width=0.7\textwidth]{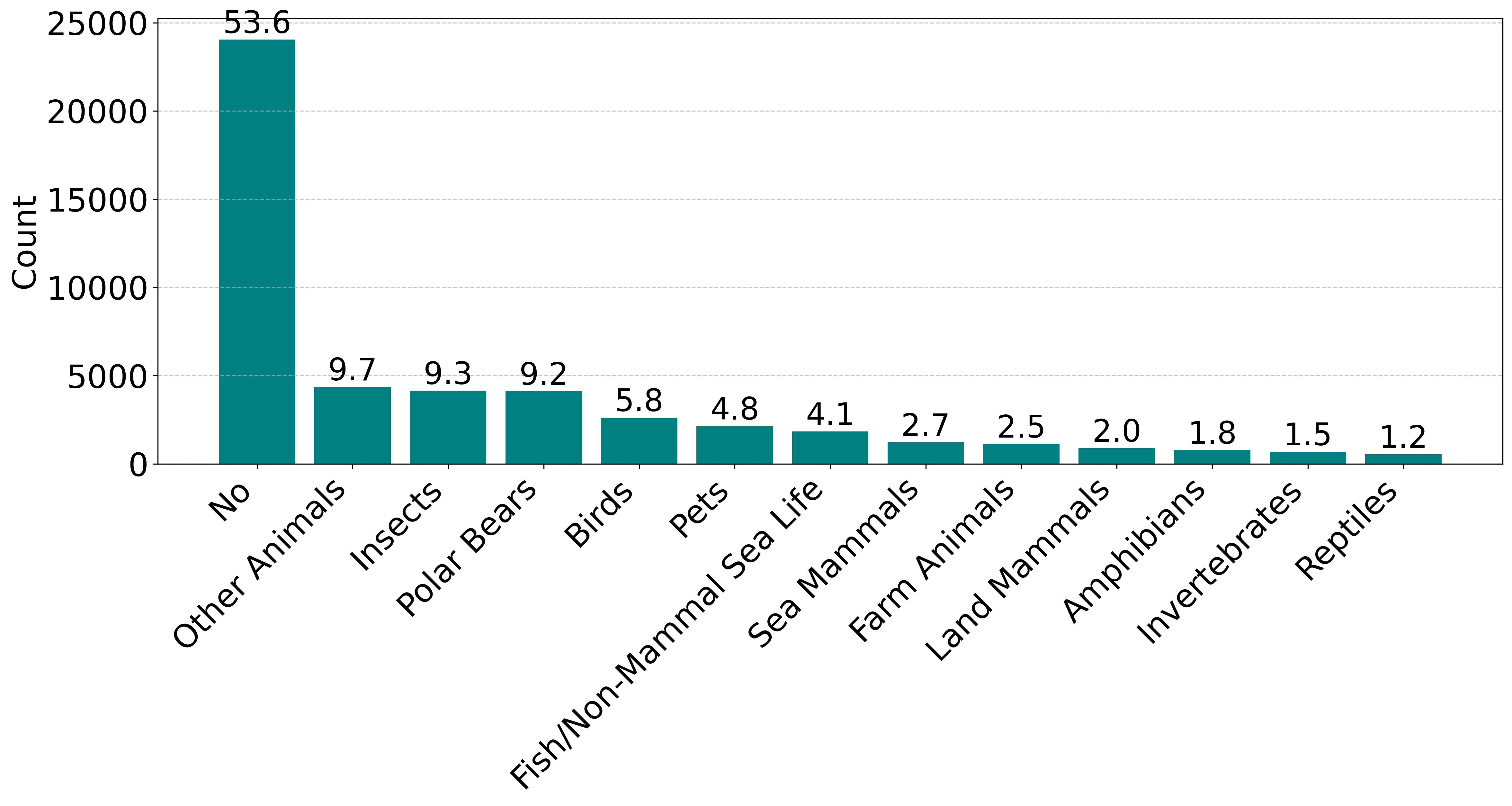}
    \caption{CLIP - Classification Results for the \textit{animals} group. 
Bar values indicate the relative frequency of each assigned category, based on a total of 44,927 videos.}       
    \label{fig:clip_animals}
\end{figure}

\cref{fig:clip_animals} shows the classification results for the \textit{animals} group. It is worth noting that CLIP assigned at least one \textit{animal} category to only 46.4\% of all videos.
Furthermore, no single category exceeds 10\% of the total classified videos. \textit{other animals}, \textit{insects}, and \textit{polar bears} each account for over 9\%, while the remaining categories fall between 1\% and 6\%.

\begin{figure}[ht]
    \centering
    \includegraphics[width=0.7\textwidth]{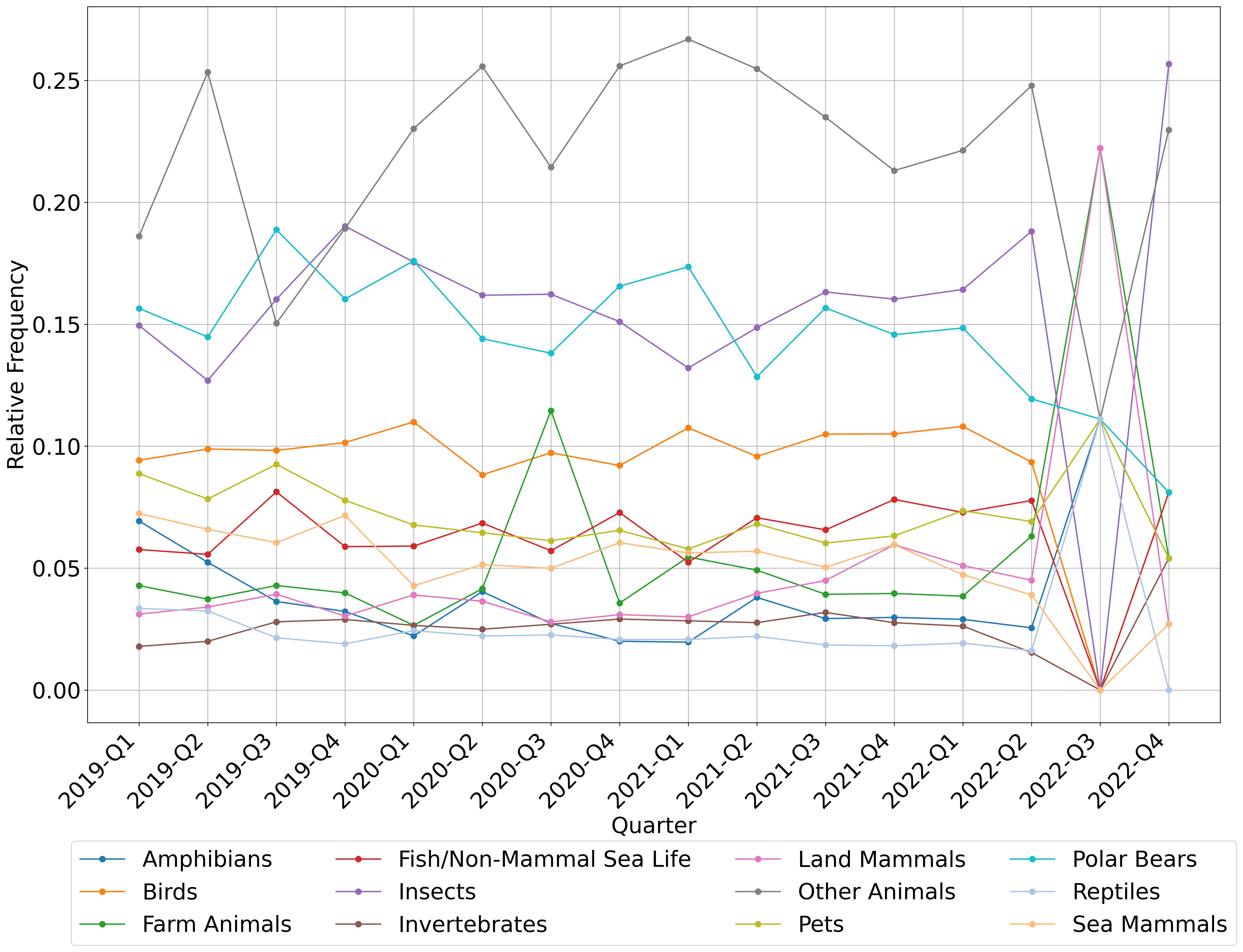}
    \caption{CLIP: Trend of \textit{animals}}       
    \label{fig:clip_animals_trend}
\end{figure}

\newpage
The trends observed for the \textit{animals} categories in \cref{fig:clip_animals_trend} indicate a correlation between high relative frequency and high volatility. Specifically, the categories with the highest count (\textit{other animals}, \textit{insects} and \textit{polar bears}) also exhibit the greatest volatility. In contrast, the remaining categories show minimal volatility, with the exception of \textit{farm animals} in 2020-Q3.
Furthermore, a period of extreme volatility is noticeable across nearly all categories in 2022-Q3.

\newpage
\subsubsection{Category Group: Consequences}
\begin{figure}[ht]
    \centering
    \includegraphics[width=0.7\textwidth]{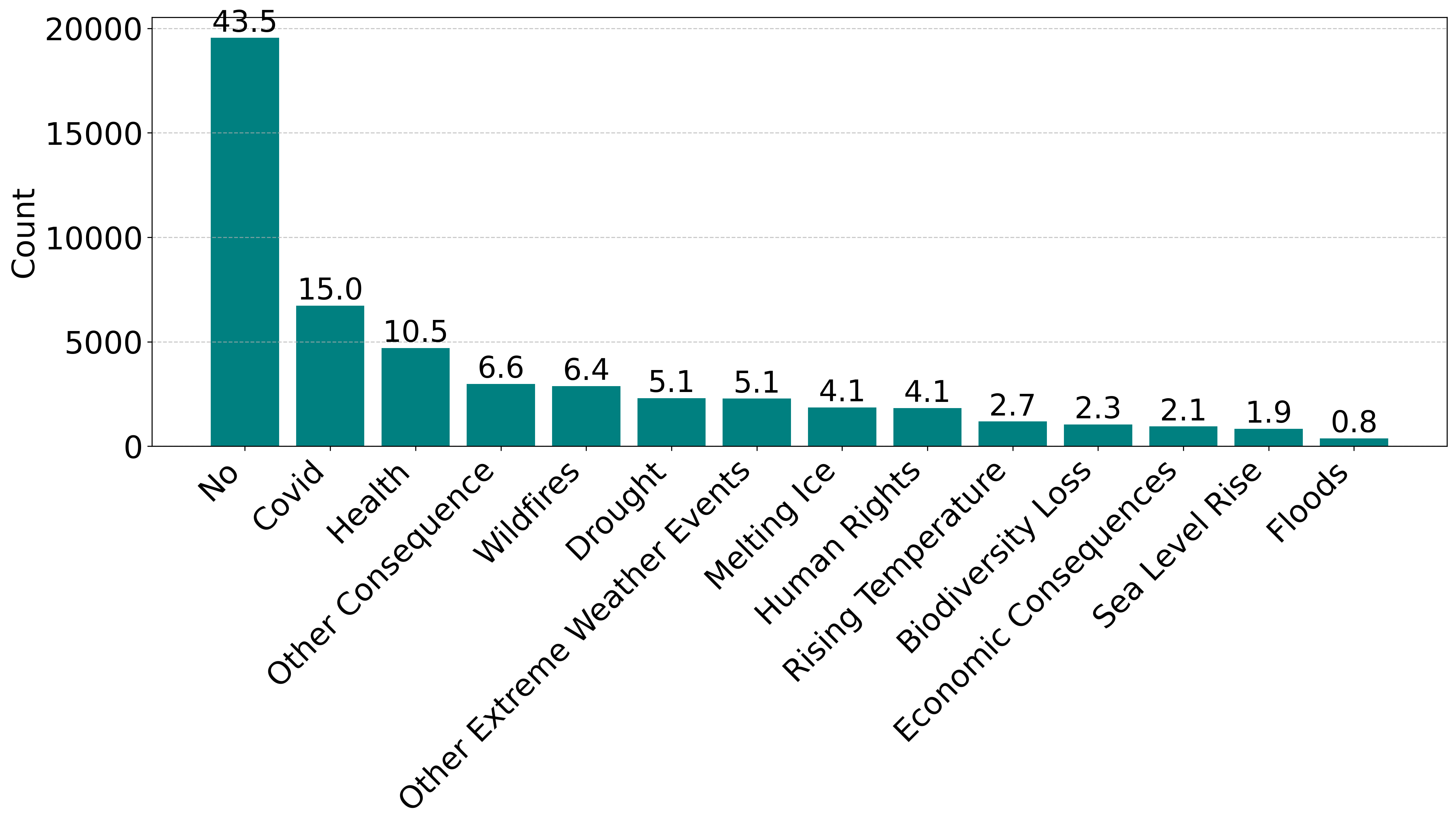}
    \caption{CLIP - Classification Results for the \textit{consequences} group. 
Bar values indicate the relative frequency of each assigned category, based on a total of 44,927 videos.}       
    \label{fig:clip_consequences}
\end{figure}

\begin{figure}[ht]
    \centering
    \includegraphics[width=0.7\textwidth]{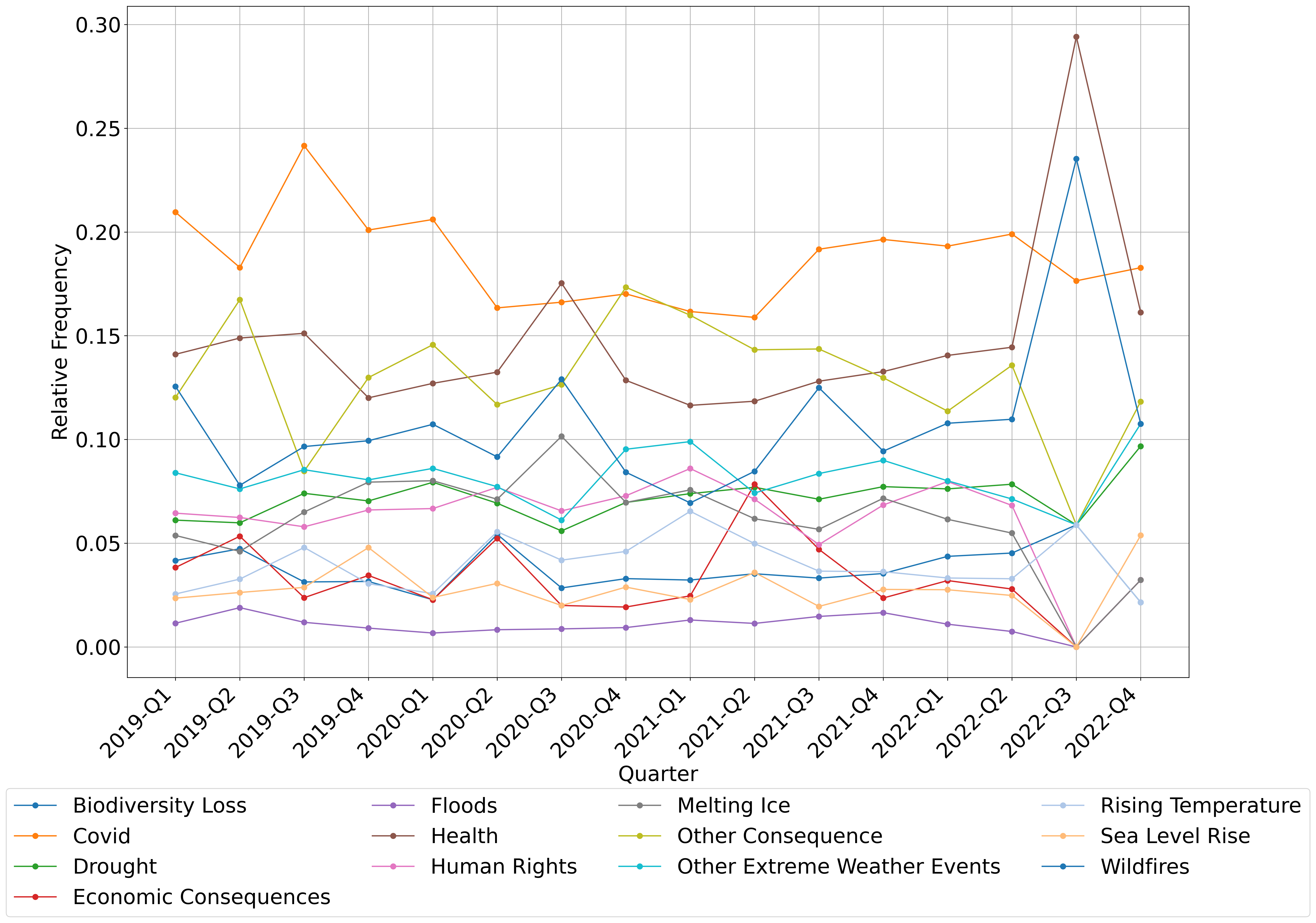}
    \caption{CLIP: Trend of \textit{consequences}}       
    \label{fig:clip_consequences_trend}
\end{figure}

As shown in \cref{fig:clip_consequences}, only 56.5\% of the videos were assigned a \textit{consequences} category. \textit{Covid} (15\%) and \textit{health} (10.5\%) were the most frequent assignments, while all other categories ranged from under 7\% to 0.8\%. Notably, no clear correlation was observed between potentially related extreme weather event categories such as \textit{wildfires}, \textit{droughts}, \textit{other extreme weather events} and \textit{floods}.
The trend graph for the \textit{consequences} categories (\cref{fig:clip_consequences_trend}) exhibit significant volatility across nearly all categories and relative frequencies. Some similar patterns can be observed among \textit{human rights}, \textit{melting ice}, \textit{drought} and \textit{other extreme weather events}. In 2022-Q3, many categories converge to similar relative frequencies, contributing to the high volatility observed in this quarter.

\newpage
\subsubsection{Category Group: Climate Action}
\begin{figure}[ht]
    \centering
    \includegraphics[width=0.7\textwidth]{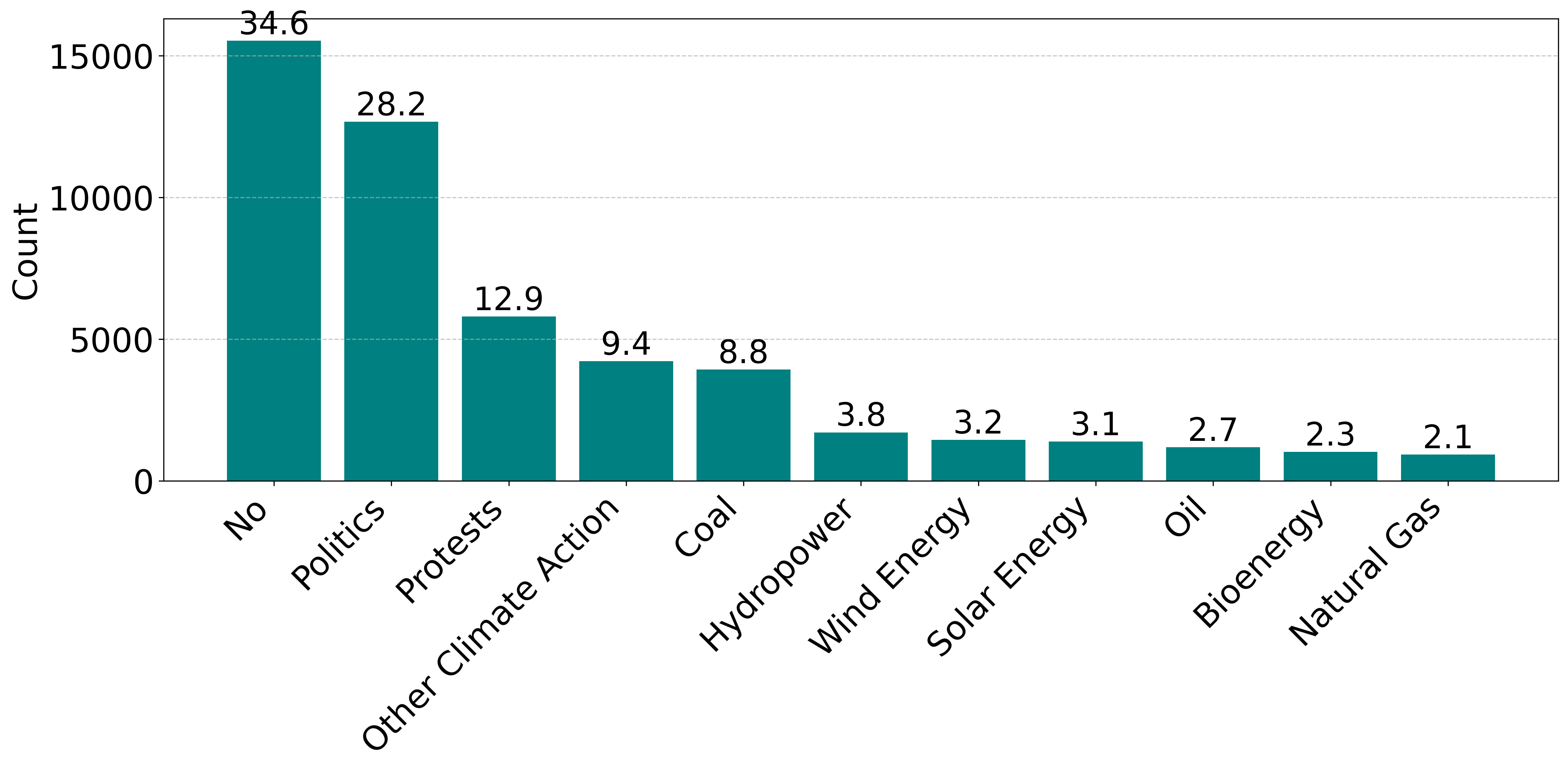}
    \caption{CLIP - Classification Results for the \textit{climate action} group. 
Bar values indicate the relative frequency of each assigned category, based on a total of 44,927 videos.}       
    \label{fig:clip_climateactions}
\end{figure}

\begin{figure}[ht]
    \vspace{-1cm}
    \centering
    \includegraphics[width=0.7\textwidth]{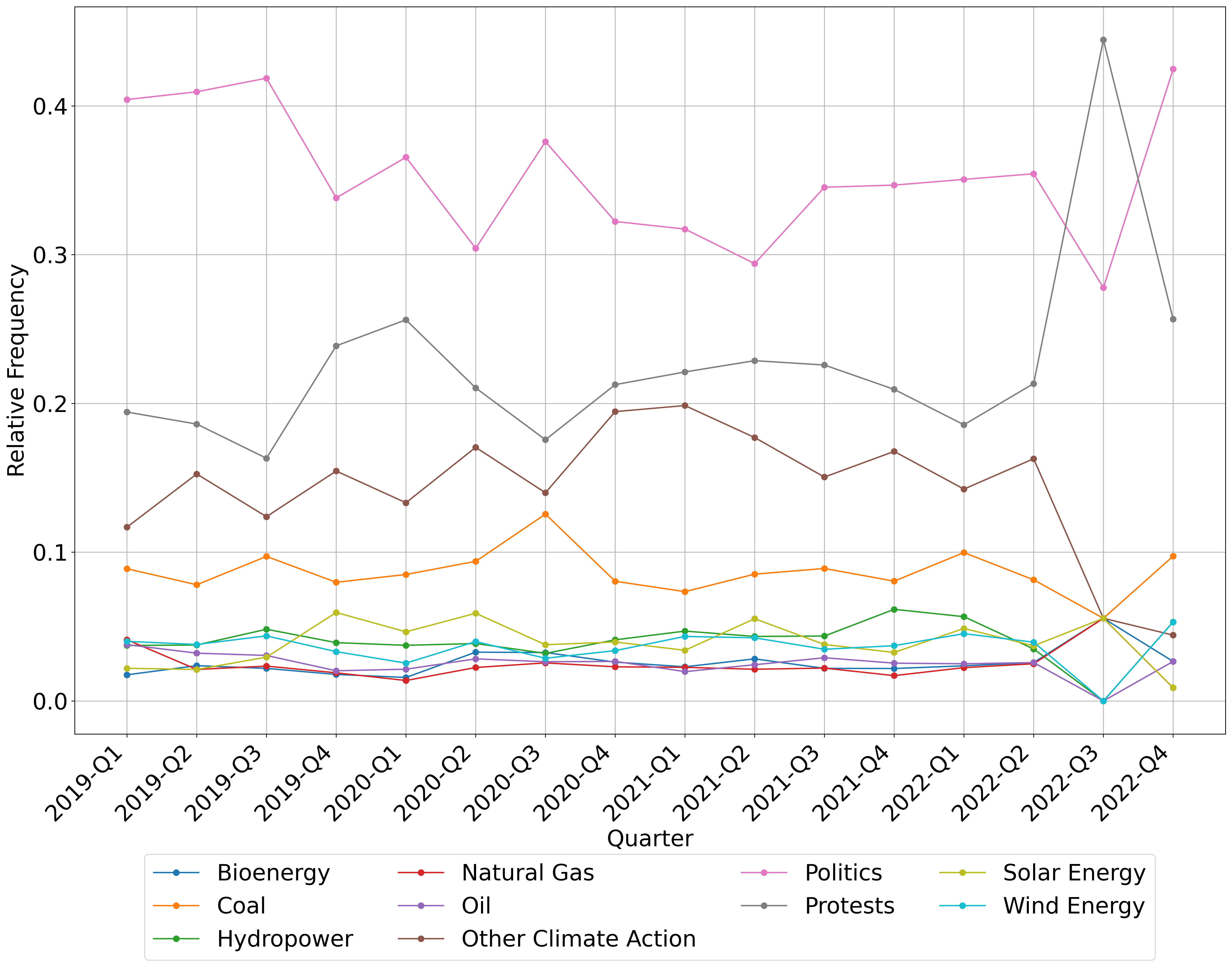}
    \caption{CLIP: Trend of \textit{climate action}}       
    \label{fig:clip_climateactions_trend}
\end{figure}

\cref{fig:clip_climateactions} reveals that CLIP assigned 28.2\% of videos to the \textit{politics} category, making it by far the most frequent \textit{climate action} related classification across all videos. 
Both renewable and fossil energy source categories fall between 2\% and 4\%, except for \textit{coal}, which accounts for 8.8\%.
The trend analysis for \textit{climate action} demonstrates that the categories with the highest count also exhibit the greatest volatility (\cref{fig:clip_climateactions_trend}). 
This pattern extends to the top four most frequent categories. 
In contrast, the remaining categories show little significant volatility. 
However, a close correlation in trends can be observed between \textit{oil}, \textit{natural gas} and \textit{bio energy}, while \textit{wind energy}, \textit{solar energy} and \textit{hydropower} show only some similarities in their movements.

\newpage
\subsubsection{Category Group: Setting}
\begin{figure}[ht]
    \centering
    \includegraphics[width=.7\textwidth]{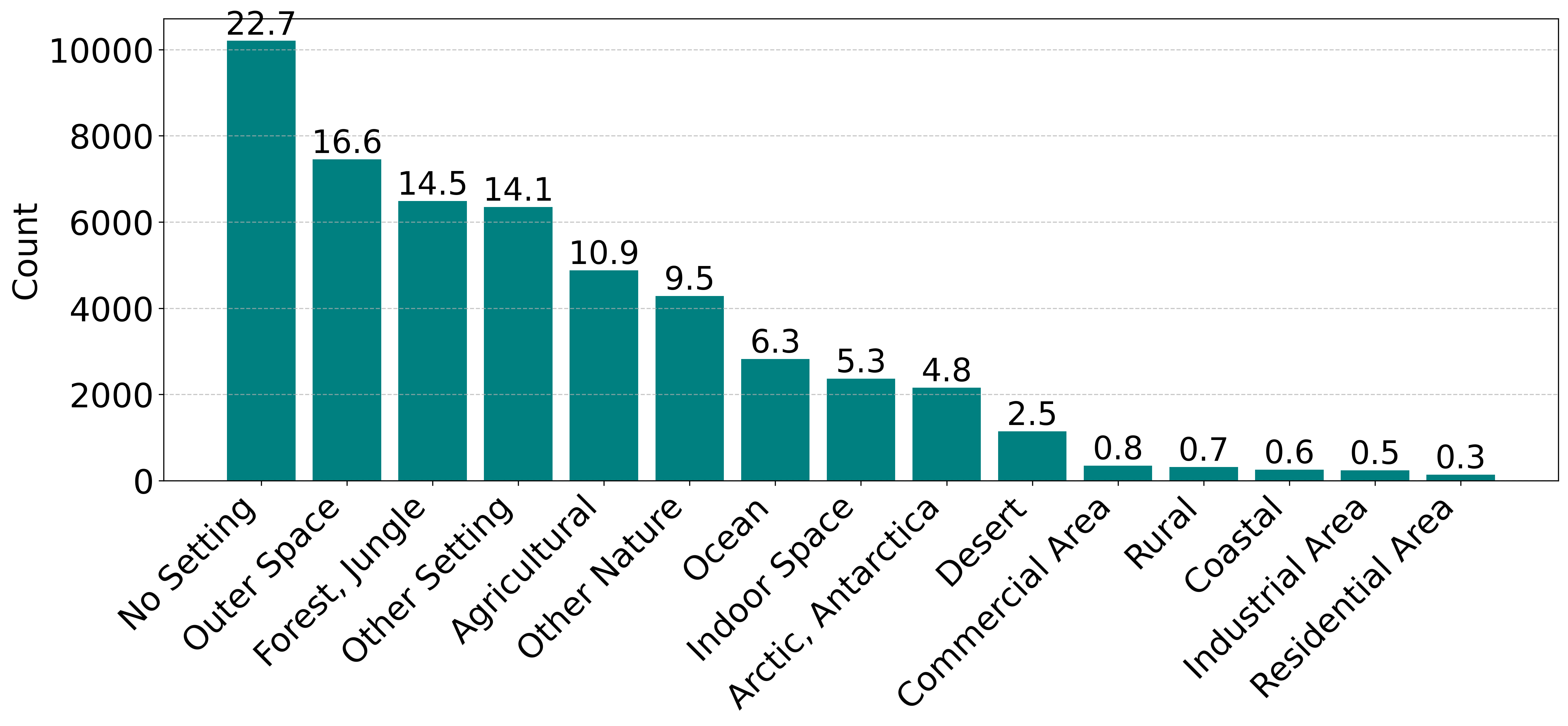}
    \caption{CLIP - Classification Results for the \textit{setting} group. 
Bar values indicate the relative frequency of each assigned category, based on a total of 44,927 videos.}       
    \label{fig:clip_setting}
\end{figure}

The display of the result of classifying in the \textit{setting} group is shown in \cref{fig:clip_setting}. It is indicated that \textit{no setting} was the most frequent assignment for videos. Also, noteworthy is the selection of \textit{outer space} at 16.6\%. \textit{forest or jungle}, \textit{other setting}, \textit{agricultural}, and \textit{other nature} all fall within the range of 14.5\% to 9.5\%. Categories that were rarely selected (under 1\%) include \textit{commercial area}, \textit{rural}, \textit{coastal}, \textit{industrial area} and \textit{residential area}.

\begin{figure}[ht]
    \centering
    \includegraphics[width=.7\textwidth]{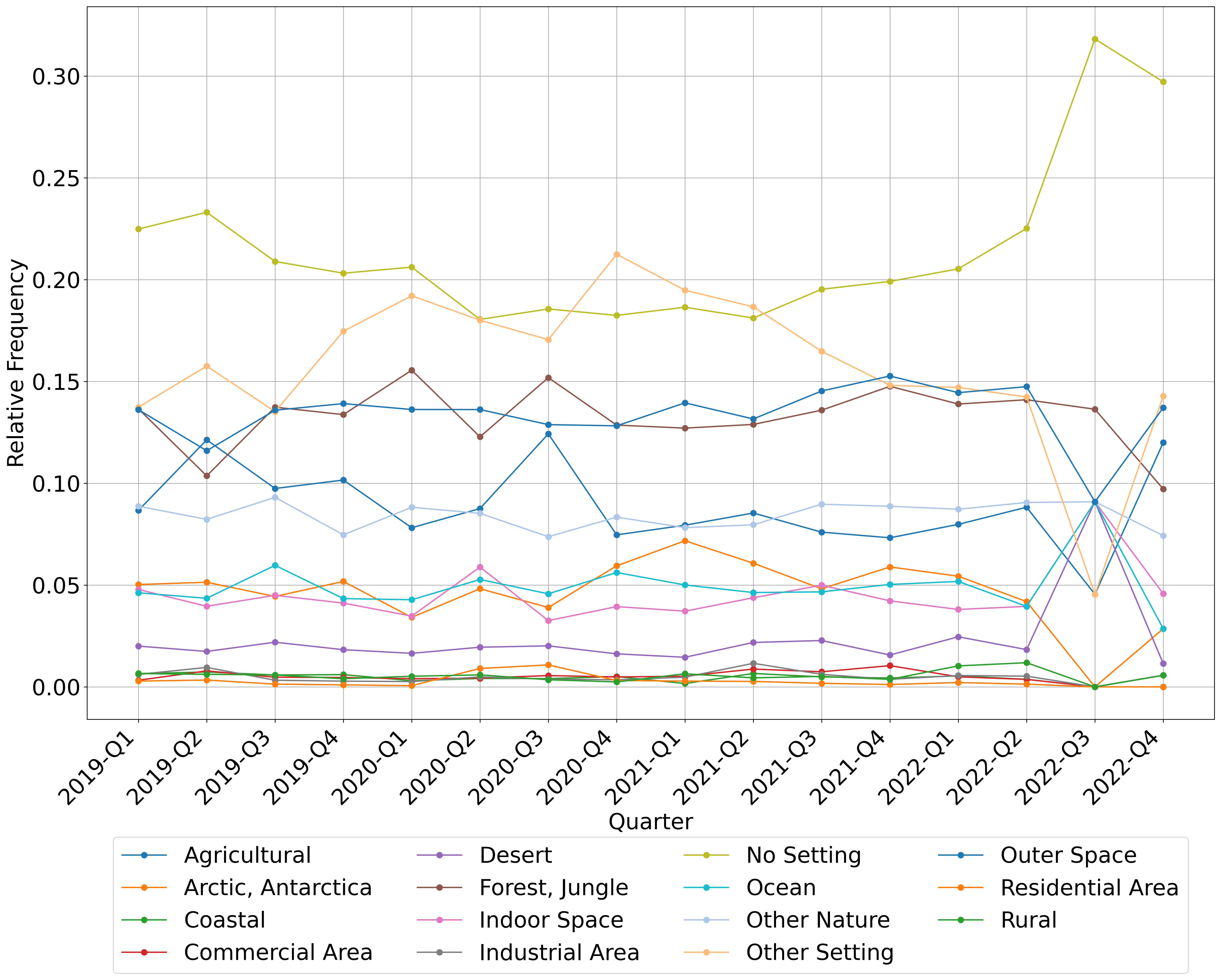}
    \caption{CLIP: Trend of \textit{setting}}       
    \label{fig:clip_setting_trend}
\end{figure}

The trend graph (\cref{fig:clip_setting_trend}) shows moderate volatility across most categories, with the exception of \textit{commercial area}, \textit{rural}, \textit{coastal}, \textit{industrial area}, \textit{residential area} and \textit{desert}, which exhibit minimal change. Notably, 2022-Q3 shows extreme volatility across almost every category due to many categories converging to similar relative frequencies during this period.

\newpage
\subsubsection{Category Group: Type} 
\begin{figure}[ht]
    \centering
    \includegraphics[width=0.7\textwidth]{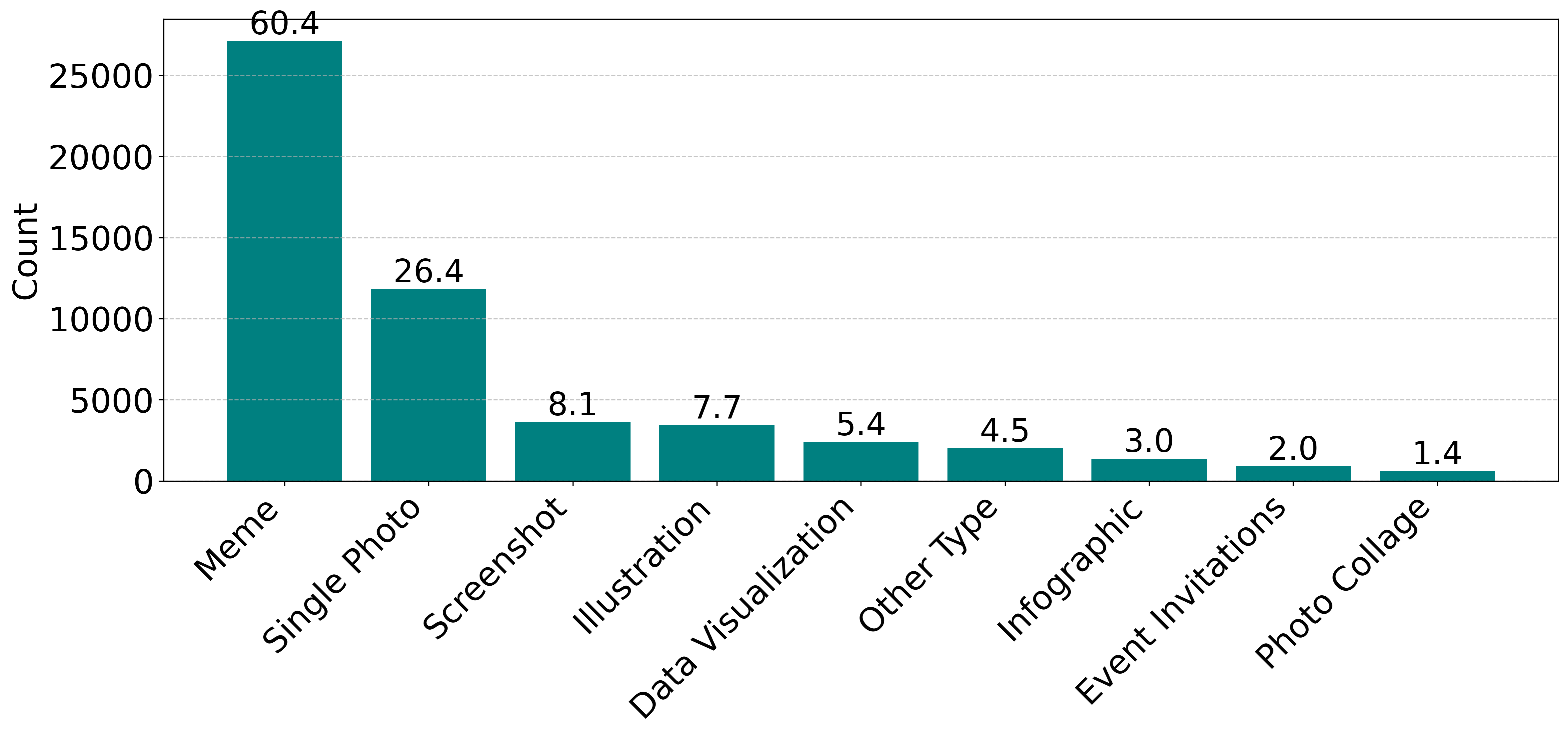}
    \caption{CLIP - Classification Results for the \textit{type} group.
Bar values indicate the relative frequency of each assigned category, based on a total of 44,927 videos.}       
    \label{fig:clip_type}
\end{figure}

\begin{figure}[ht]
    \centering
    \includegraphics[width=0.7\textwidth]{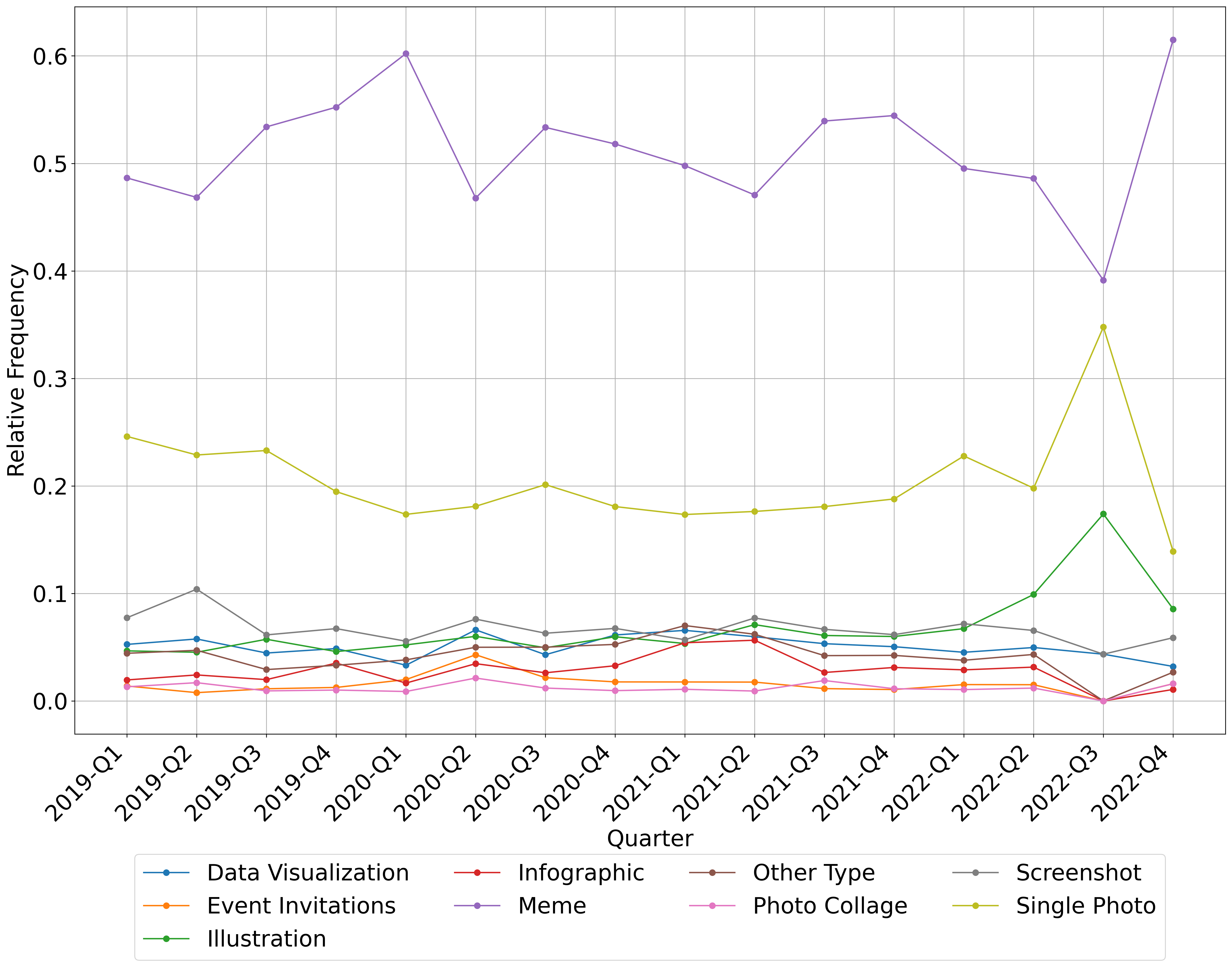}
    \caption{CLIP: Trend of \textit{type}}       
    \label{fig:clip_type_trend}
\end{figure}

\cref{fig:clip_type} reveals that CLIP assigned the label \textit{meme} to 60.4\% of all videos. \textit{single photo} was used for 26.4\% of the videos. All other video types were represented less than 10\%, ranging down to 1.4\%.
Except for the \textit{meme} category and the 2022-Q3 period, the trend graph (\cref{fig:clip_type_trend}) shows minimal volatility for every other category. The most volatility outside the 2022-Q3 period is shown by every category except \textit{single photo} in 2020-Q2 where \textit{meme} declined which was picked up most notable from all categories with a low relative frequency.

\newpage
\subsection{Comparison of Type Prediction}
\label{app:typepred}

\begin{figure}[ht]
    \centering
    \includegraphics[width=0.7\textwidth]{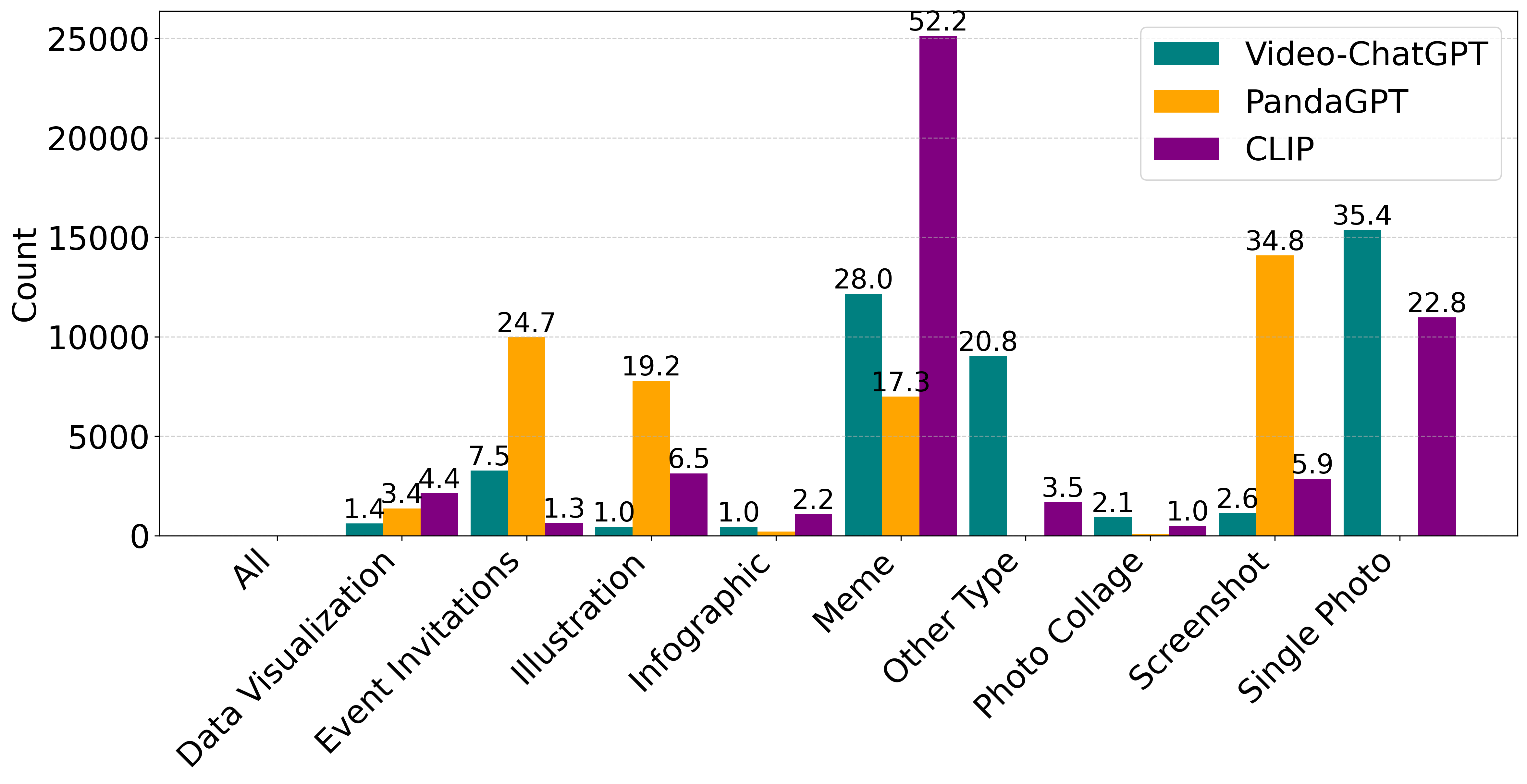}
    \caption{Combined Results of Classifying for Video-ChatGPT, PandaGPT \& CLIP - \textit{type}. Relative values from 1\% are shown on bars.}       
    \label{fig:combined_type}
\end{figure}

\cref{fig:combined_type} presents the combined classification results for the \textit{type} group from Video-ChatGPT, PandaGPT, and CLIP. 
It is important to note that Video-ChatGPT failed to classify 4466 videos. 
To ensure a fair comparison across all models in this comparison, these videos are excluded from the evaluation for every model.

Despite overall differences among the models, some similarities emerge in the categorization of \textit{All}, \textit{Infographic}, \textit{Data Visualization}, and \textit{Photo Collage}, since these categories were rarely selected by any of the models.

The most significant difference appears in the \textit{meme} category, which CLIP used far more frequently compared to the other models. Additionally, it is worth noting that \textit{other type} and \textit{single photo} were never chosen by PandaGPT, while Video-ChatGPT and CLIP assigned these categories to a certain extent. 

These differences are further proven by analyzing the Jaccard similarities between each pair of models (\cref{tab:jaccard_type}). Even under the best-case scenario where perfect agreement is inherently unachievable due to differing output distributions the observed Jaccard indices fall markedly short. \newpage
For instance, for the \textit{meme} category, the highest possible overlap between Video-ChatGPT (28.0\%) and CLIP (52.2\%) is 53.64\%, yet the actual Jaccard similarity is only 29.57\%.

These findings demonstrate that the models yield substantially different classification results. Given the absence of consistent labeling patterns, except in infrequently used categories, and the previously established reliability concerns for Video-ChatGPT and PandaGPT in other groups, no interpretative data can be derived for the \textit{type} group. Consequently, a combined trend analysis would be of limited value.

\begin{table}[ht]
  \centering
  \footnotesize
  \caption{Jaccard Similarity for the Combined Results of Classifying for Video-ChatGPT, PandaGPT \& CLIP - \textit{type}}
    \begin{tabularx}{\textwidth}{l||X|X|X}
    \toprule
    \textbf{Category} & \textbf{Video-ChatGPT} & \textbf{Video-ChatGPT} & \textbf{PandaGPT} \\
                      & \textbf{PandaGPT} & \textbf{CLIP} & \textbf{CLIP} \\ \hline
    All & 0.00\% & 0.00\% & 0.00\% \\
        & (0/10) & (0/10) & (0/0) \\
    Data Visualization & 7.92\% & 9.08\% & 29.23\% \\
        & (129/1629) & (171/1884) & (766/2621) \\
    Event Invitations & 7.79\% & 2.28\% & 2.38\% \\
        & (1035/13279) & (78/3426) & (276/11612) \\
    Illustration & 0.26\% & 0.92\% & 17.11\% \\
        & (22/8479) & (23/2508) & (1596/9326) \\
    Infographic & 0.00\% & 0.33\% & 5.40\% \\
        & (0/319) & (3/906) & (58/1074) \\
    Meme & 20.72\% & 29.57\% & 20.94\% \\
        & (3293/15893) & (7427/25116) & (4917/23486) \\
    Other Type & 0.00\% & 2.53\% & 0.00\% \\
        & (0/8631) & (246/9721) & (0/1336) \\
    Photo Collage & 0.46\% & 4.30\% & 0.76\% \\
        & (3/646) & (37/861) & (3/396) \\
    Screenshot & 2.08\% & 1.96\% & 6.50\% \\
        & (336/16126) & (53/2705) & (1081/16641) \\
    Single Photo & 0.02\% & 17.19\% & 0.02\% \\
        & (3/14889) & (3160/18388) & (1/6665) \\
 
    \bottomrule
  \end{tabularx}
\label{tab:jaccard_type}
\end{table}
\newpage
\subsection{Trend in 2022-Q3}
Something present in all shown trends is an increase in volatility in 2022-Q3. This can be easily explained due to the amount of videos present in this quarter. As indicated in \cref{tab:videos per quarter}, 2022-Q3 only contains 20 videos. This small sample size renders the observed trends in this quarter statistically unreliable, amplifying the impact of individual videos on the relative frequencies of the categories. A similar observation can be made about 2022-Q4, though to a lesser degree.

\begin{table}[ht]
  \centering
  \footnotesize
  \caption{Number of videos per quarter.}
    \begin{tabularx}{0.5\textwidth}{X|X}
      \toprule
      \textbf{Quarter} & \textbf{Videos} \\ \hline
      2019-Q1 & 2,196 \\
      2019-Q2 & 3,056 \\
      2019-Q3 & 6,735 \\
      2019-Q4 & 4,426 \\
      2020-Q1 & 4,522 \\
      2020-Q2 & 4,497 \\
      2020-Q3 & 6,159 \\
      2020-Q4 & 5,155 \\
      2021-Q1 & 3,691 \\
      2021-Q2 & 6,656 \\
      2021-Q3 & 7,300 \\
      2021-Q4 & 5,472 \\
      2022-Q1 & 8,446 \\
      2022-Q2 & 4,669 \\
      2022-Q3 & 20 \\
      2022-Q4 & 166 \\
      \bottomrule
  \end{tabularx}
\label{tab:videos per quarter}
\end{table}

\newpage
\section{Detailed Minimum Cost Multicut Clustering Results}
\label{app:clustering}
\subsection{Clustering Calibration Term Choice}
\label{app:calterm}
To determine the most suitable calibration term, we ablated the calibration term $cal = [0.1, .. , 0.9]$ on ConvNeXt V2 and DINOv2, both with the static frame selection and the simple averaging frame combination method. 
In contrast with \citet{prasse_i_2025}'s approach, where they assess similarity of image classes and clusters for calibration term validation, we report standard clustering metrics to determine the optimal calibration term.  

\cref{tab:calibration_metrics_performance_convnextv2} and \ref{tab:calibration_metrics_performance_dinov2} report the ranking of the calibration terms based on the weighted overall score.
The lower calibration terms are ranked as optimal for both models ($cal=0.1, 0.2, 0.3$ for ConvNeXt V2, and $cal=0.1$ for DINOv2), however, these configurations result in extremely fragmented clustering (more than 80\% of singletons), which offers limited practical value for visual frame analysis.

Therefore, taking practical interpretability into account, we choose the calibration term, $cal = 0.7$, for ConvNeXt~V2 (Table~\ref{tab:calibration_conv}), which results in 79 clusters with efficient coverage (88.7\% by the top 10 clusters) and low fragmentation (13.9\% singletons). 
For DINOv2 (Table~\ref{tab:calibration_dinov2}), the calibration term $cal = 0.6$ is selected, as it results in 44 clusters with strong coverage efficiency (97.1\% by the top 10 clusters) and moderate fragmentation (29.6\% singletons). 
These calibration choices balance quantitative cluster quality with semantic interpretability, and are consistent with the calibration validation results reported by \citet{prasse_i_2025}.

\begin{table}[ht]
\centering
\footnotesize
\caption{ConvNeXt V2 clustering statistics across calibration terms}
\label{tab:calibration_conv}
\resizebox{\textwidth}{!}{%
\begin{tabular}{cccccccccc}
\toprule
\textbf{Calibration} & \textbf{Clusters} & \textbf{Largest} & \textbf{Avg Size} & \textbf{Median} & \textbf{Singletons (\%)} & \textbf{Top 10\% (\%)} & \textbf{Top 20\% (\%)} & \textbf{\#Clusters 80\%} & \textbf{Gini} \\
\midrule
0.1 & 2703 & 9    & 1.10 & 1.0   & 93.12 & 1.95 & 3.20 & 2109 & 0.085 \\
0.2 & 2342 & 110  & 1.27 & 1.0   & 88.13 & 7.71 & 9.76 & 1748 & 0.199 \\
0.3 & 1893 & 217  & 1.57 & 1.0   & 82.57 & 17.31 & 20.13 & 1299 & 0.339 \\
0.4 & 1317 & 249  & 2.26 & 1.0   & 72.06 & 27.47 & 32.73 & 723  & 0.508 \\
0.5 & 719  & 488  & 4.13 & 1.0   & 54.24 & 44.31 & 51.35 & 227  & 0.674 \\
0.6 & 303  & 1074 & 9.80 & 2.0   & 31.35 & 64.04 & 70.37 & 50   & 0.789 \\
0.7 & 79   & 1558 & 37.59 & 4.0  & 13.92 & 88.65 & 92.83 & 3    & 0.882 \\
0.8 & 6    & 2917 & 495.0 & 6.0  & 16.67 & 100.0 & 100.0 & 1    & 0.824 \\
0.9 & 1    & 2970 & 2970.0 & 2970.0 & 0.0 & 100.0 & 100.0 & 1    & 0.000 \\
\bottomrule
\end{tabular}%
}
\end{table}

\begin{table}[ht]
\centering
\footnotesize
\caption{DINOv2 clustering statistics across calibration terms}
\label{tab:calibration_dinov2}
\resizebox{\textwidth}{!}{%
\begin{tabular}{cccccccccc}
\toprule
\textbf{Calibration} & \textbf{Clusters} & \textbf{Largest} & \textbf{Avg Size} & \textbf{Median} & \textbf{Singletons (\%)} & \textbf{Top 10\% (\%)} & \textbf{Top 20\% (\%)} & \textbf{\#Clusters 80\%} & \textbf{Gini} \\
\midrule
0.1 & 2470 & 30   & 1.20 & 1.0   & 89.07 & 4.41  & 6.23  & 1876 & 0.157 \\
0.2 & 1721 & 220  & 1.73 & 1.0   & 80.65 & 18.11 & 22.22 & 1127 & 0.390 \\
0.3 & 919  & 254  & 3.23 & 1.0   & 66.05 & 37.78 & 45.76 & 325  & 0.632 \\
0.4 & 387  & 515  & 7.67 & 1.0   & 51.94 & 65.59 & 72.83 & 47   & 0.803 \\
0.5 & 159  & 1198 & 18.68 & 2.0  & 34.59 & 85.15 & 89.26 & 6    & 0.886 \\
0.6 & 44   & 1668 & 67.50 & 3.0  & 29.55 & 97.07 & 98.72 & 2    & 0.923 \\
0.7 & 14   & 2790 & 212.14 & 2.5 & 35.71 & 99.87 & 100.0 & 1    & 0.913 \\
0.8 & 3    & 2968 & 990.00 & 1.0 & 66.67 & 100.0 & 100.0 & 1    & 0.666 \\
0.9 & 1    & 2970 & 2970.0 & 2970.0 & 0.0 & 100.0 & 100.0 & 1    & 0.000 \\
\bottomrule
\end{tabular}%
}
\end{table}

\begin{table}[htbp]
\centering
\footnotesize
\caption{ConvNeXt V2 clustering performance across calibration terms (ranked by overall score)}
\label{tab:calibration_metrics_performance_convnextv2}
\resizebox{0.7\textwidth}{!}{%
\begin{tabular}{ccccc}
\toprule
\textbf{Calibration} & \textbf{Silhouette} & \textbf{Davies-Bouldin} & \textbf{Calinski-Harabasz} & \textbf{Overall Score} \\
\midrule
0.1 & 0.636  & 0.504  & 43.747 & 0.602 \\
0.2 & 0.341  & 0.913  & 15.082 & 0.520 \\
0.3 & 0.210  & 1.155  & 9.839  & 0.484 \\
0.7 & -0.022 & 2.495  & 8.151  & 0.478 \\
0.6 & -0.018 & 2.030  & 5.582  & 0.458 \\
0.4 & 0.113  & 1.450  & 7.418  & 0.456 \\
0.5 & 0.041  & 1.734  & 6.247  & 0.449 \\
0.8 & 0.014  & 3.536  & 3.710  & 0.443 \\
0.9 & -1.000 & 10.000 & 0.000  & 0.300 \\
\bottomrule
\end{tabular}%
}
\end{table}

\begin{table}[htbp]
\centering
\footnotesize
\caption{DINOv2 clustering performance across calibration terms (ranked by overall score)}
\label{tab:calibration_metrics_performance_dinov2}
\resizebox{0.7\textwidth}{!}{%
\begin{tabular}{ccccc}
\toprule
\textbf{Calibration} & \textbf{Silhouette} & \textbf{Davies-Bouldin} & \textbf{Calinski-Harabasz} & \textbf{Overall Score} \\
\midrule
0.1 & 0.460  & 0.791  & 29.674 & 0.556 \\
0.6 & 0.081  & 2.096  & 25.296 & 0.519 \\
0.7 & 0.079  & 1.986  & 21.400 & 0.517 \\
0.5 & 0.031  & 1.923  & 12.656 & 0.492 \\
0.2 & 0.207  & 1.259  & 15.088 & 0.485 \\
0.4 & 0.074  & 1.746  & 13.490 & 0.478 \\
0.3 & 0.110  & 1.566  & 12.079 & 0.464 \\
0.9 & -1.000 & 10.000 & 0.000  & 0.300 \\
0.8 & -1.000 & 10.000 & 0.000  & 0.167 \\
\bottomrule
\end{tabular}%
}
\end{table}

\newpage
\subsection{Qualitative Cluster Evaluation}
\label{app:clusterquali}

\begin{figure}[htbp]
    \centering
        \includegraphics[width=0.22\textwidth]{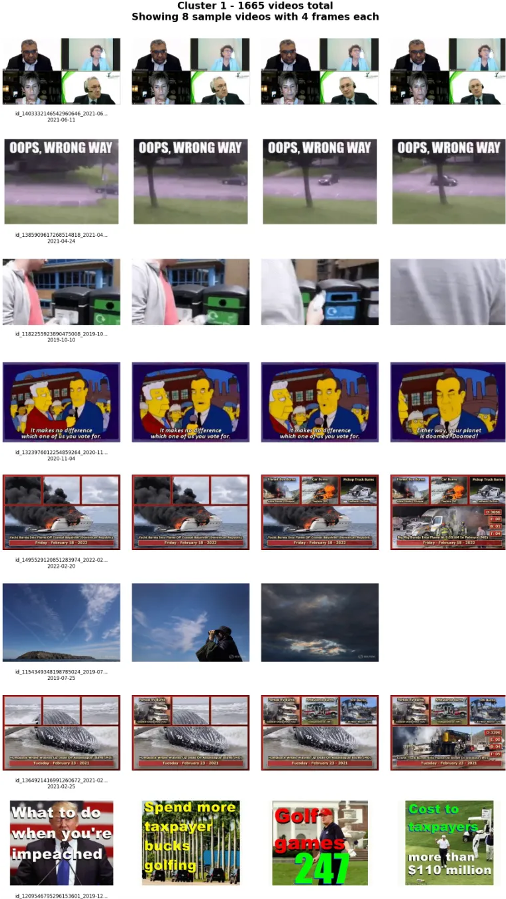}
        \includegraphics[width=0.22\textwidth]{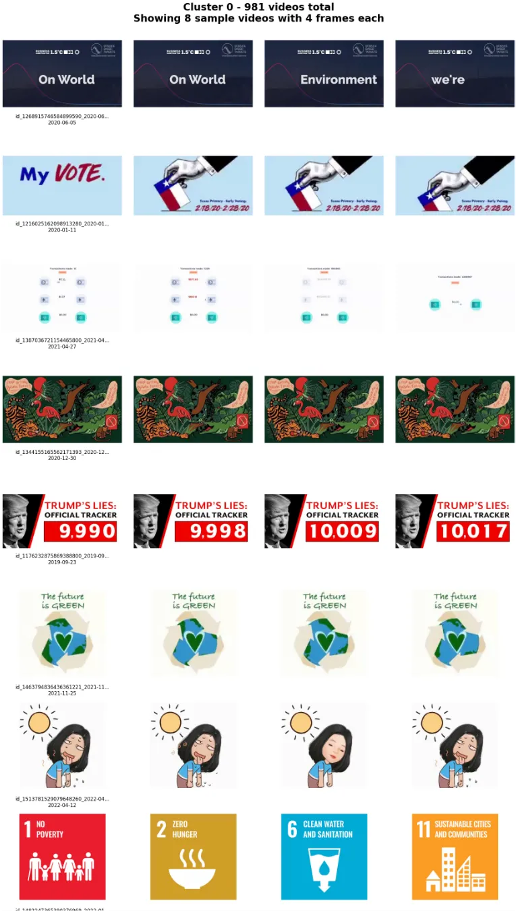}
        \includegraphics[width=0.22\textwidth]{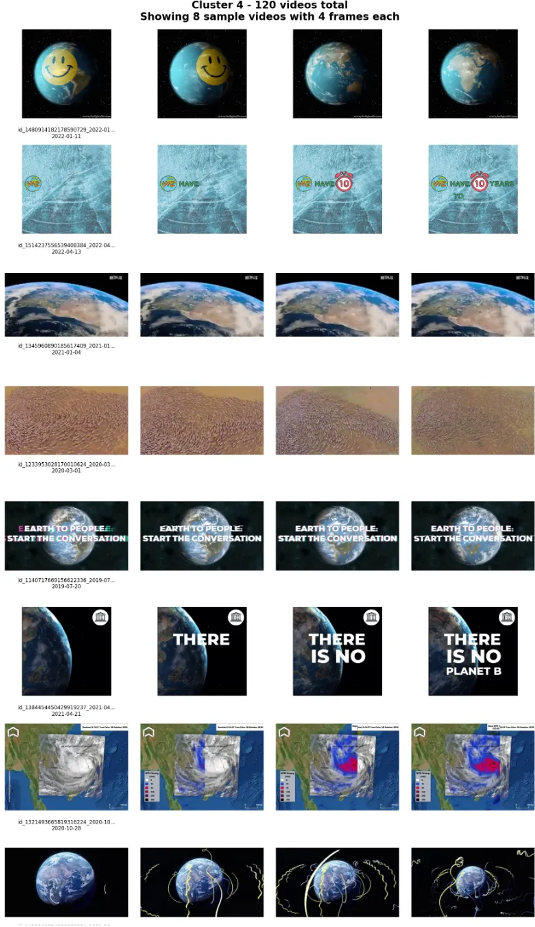}
        \includegraphics[width=0.22\textwidth]{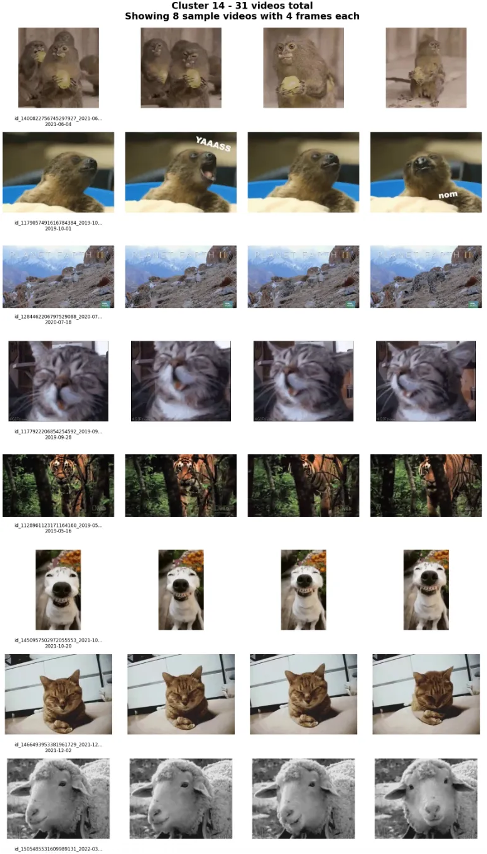}
        \includegraphics[width=0.22\textwidth]{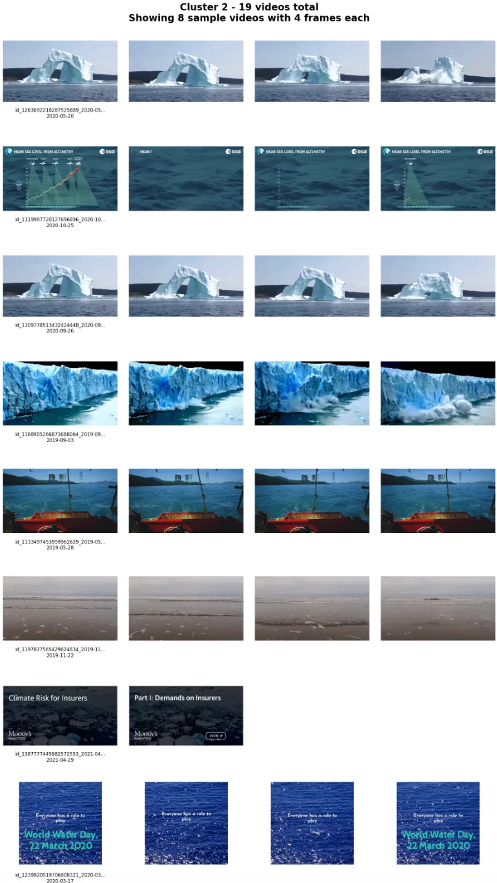}
        \includegraphics[width=0.22\textwidth]{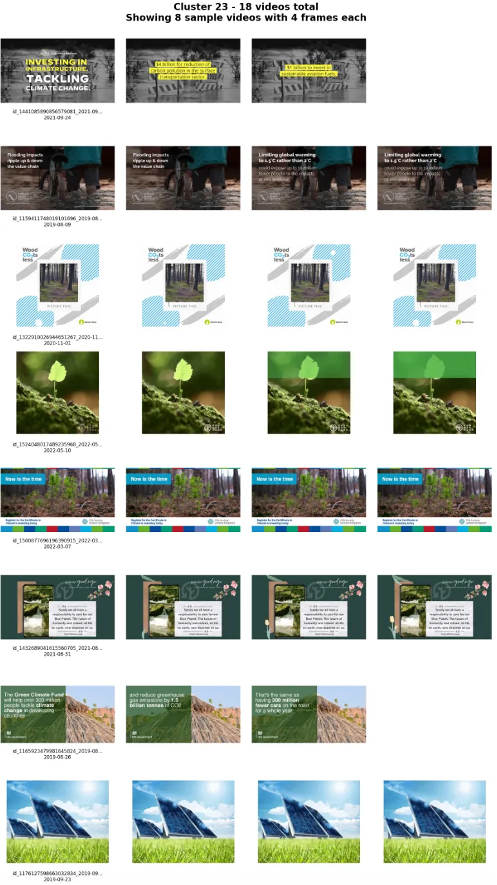}
        \includegraphics[width=0.22\textwidth]{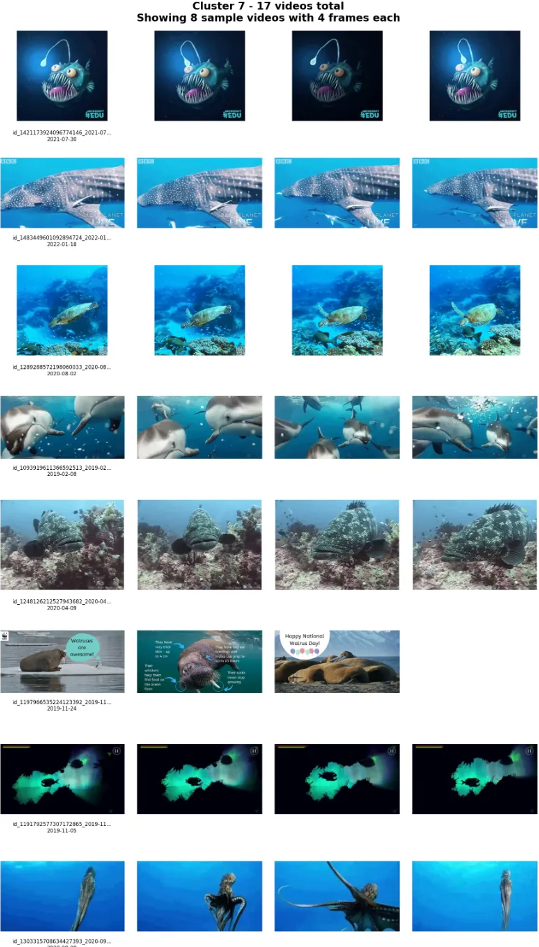}
        \includegraphics[width=0.22\textwidth]{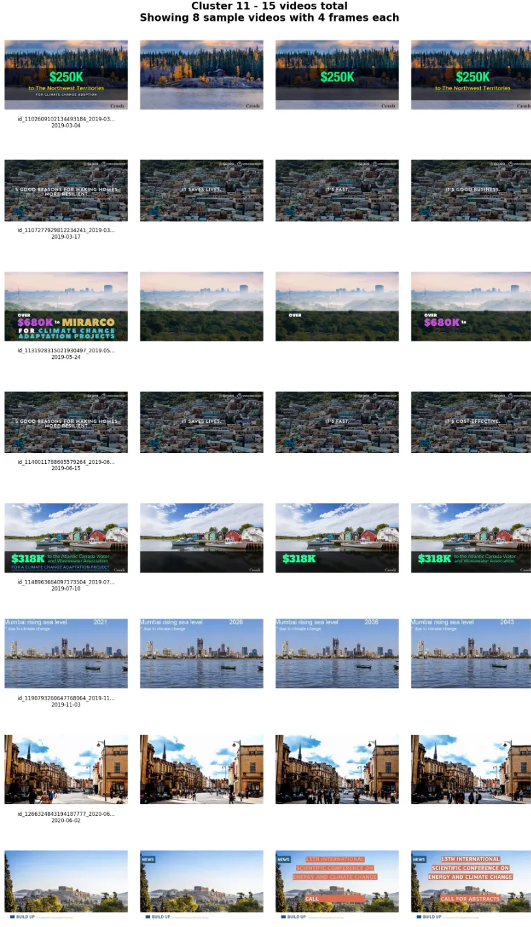}
        \includegraphics[width=0.22\textwidth]{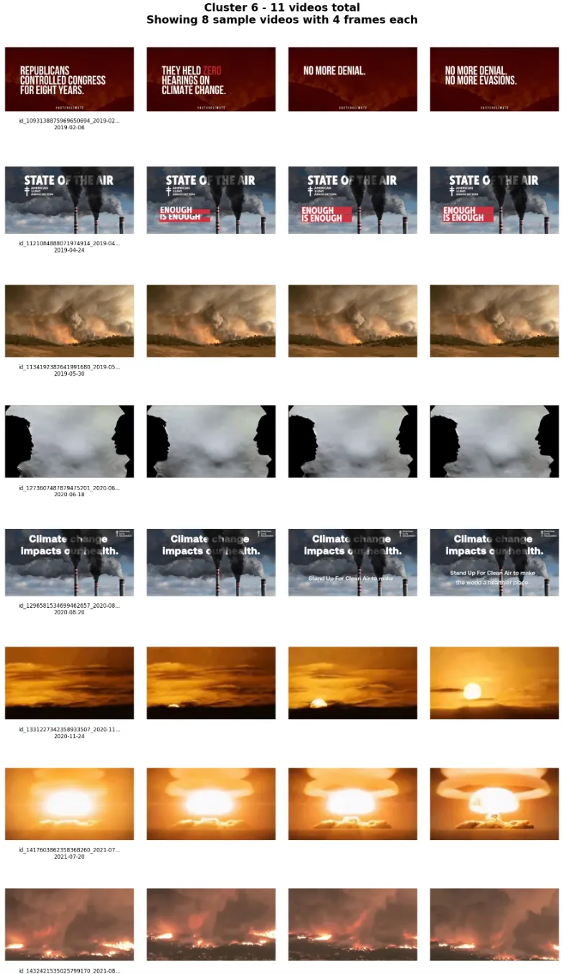}
        \includegraphics[width=0.22\textwidth]{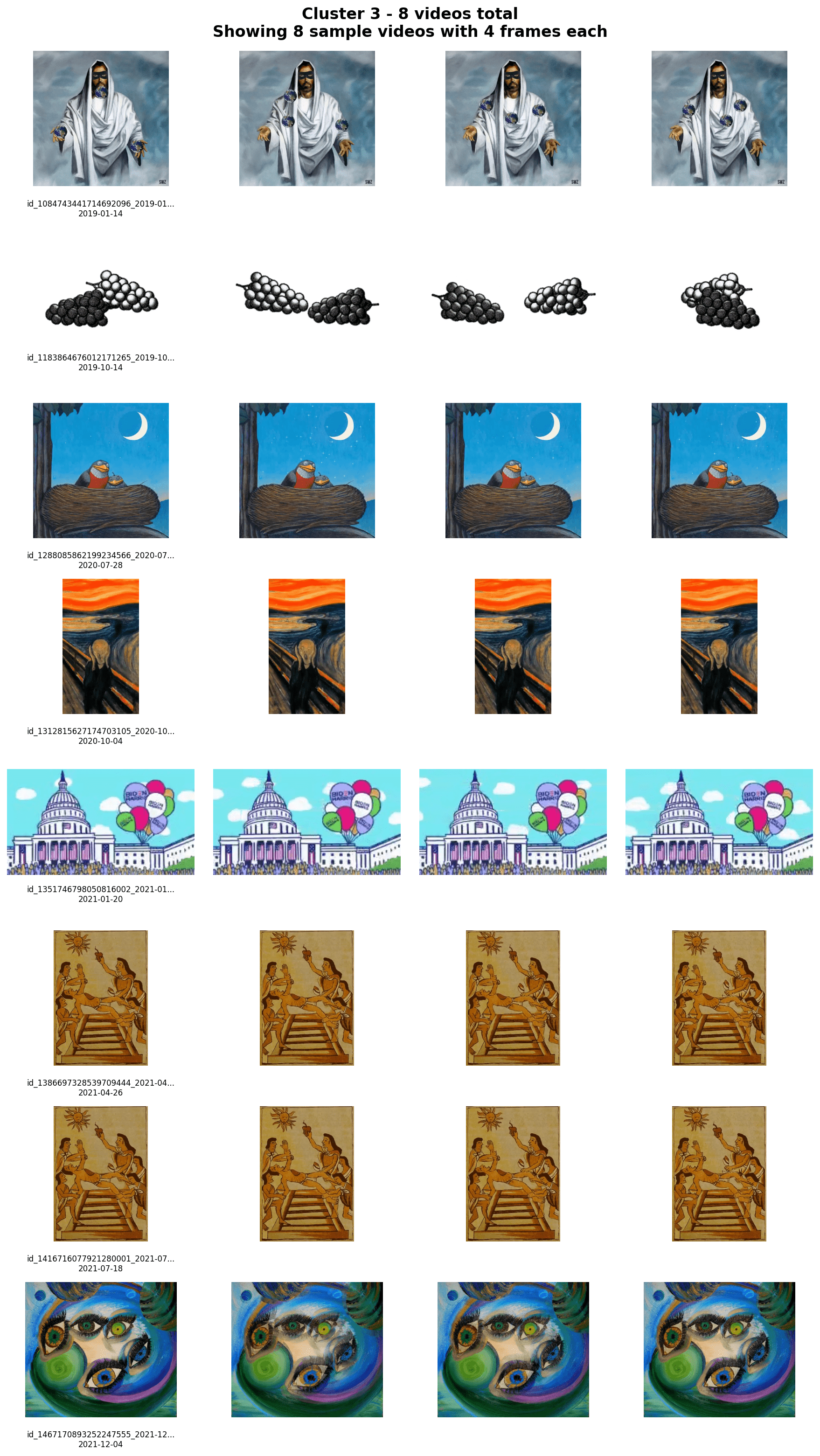}
        \caption{DINOv2 top 10 clusters: 
(1) Media and Digital Communication (1,665 videos, 56.1\%), 
(2) Information Graphics and Educational Content (981, 33.0\%), 
(3) Space/Cosmic Perspective (120, 4.0\%), 
(4) Wildlife and Animals (31 videos, 1.0\%), 
(5) Disappearing Arctic/Sea Level Rise (19, 0.6\%), 
(6) Solutions and Sustainability (18, 0.6\%), 
(7) Marine Ecosystem (17, 0.6\%), 
(8) Urban/Economic Impact (15, 0.5\%), 
(9) Climate Emergency/Disaster (11, 0.4\%), 
(10) Artwork (6, 0.2\%).}
 \label{fig:qualitative_cluster_comparison_dino}
    \end{figure}
    
    \begin{figure}[t]
        \centering
        \includegraphics[width=0.22\textwidth]{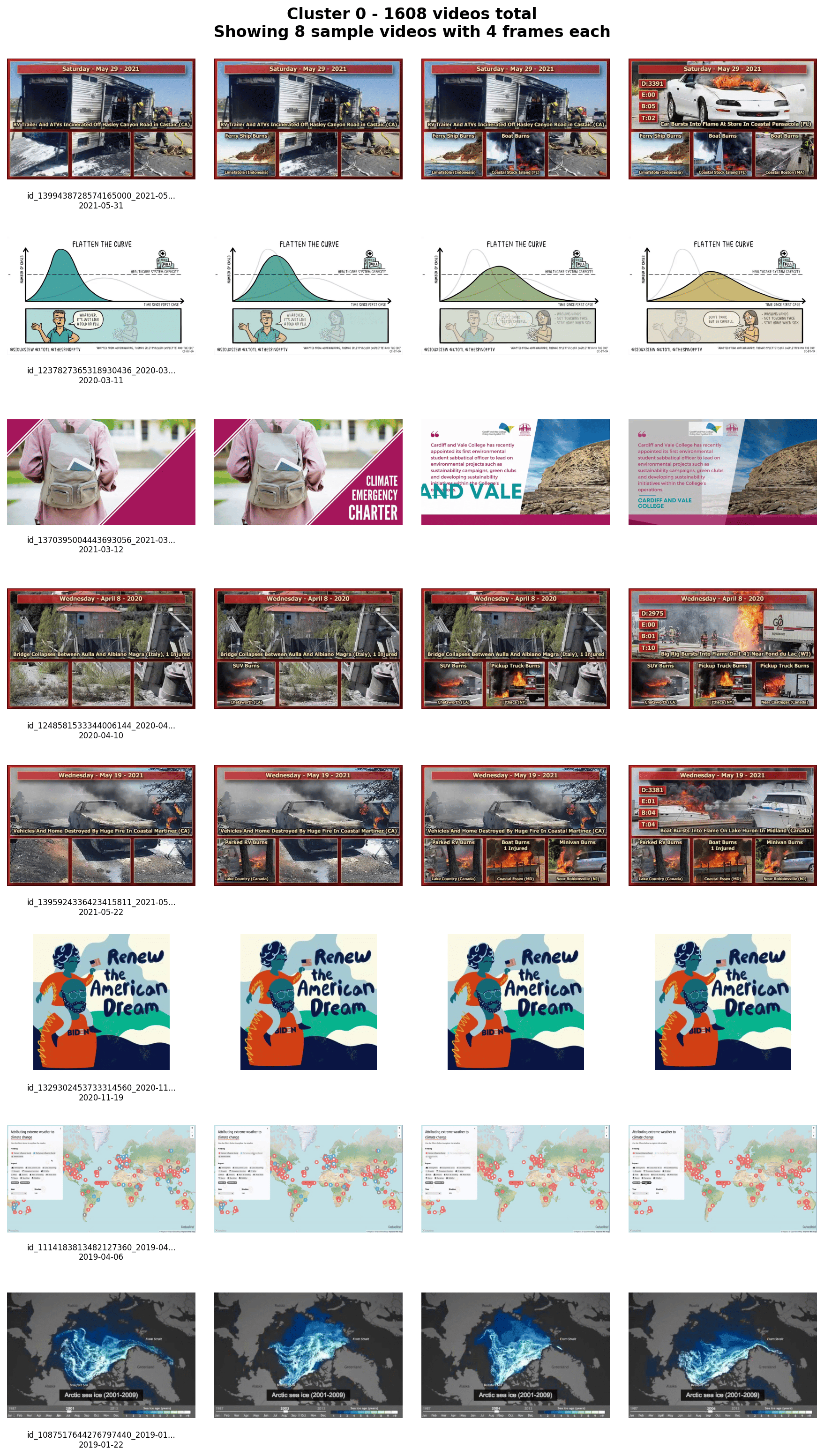}
        \includegraphics[width=0.22\textwidth]{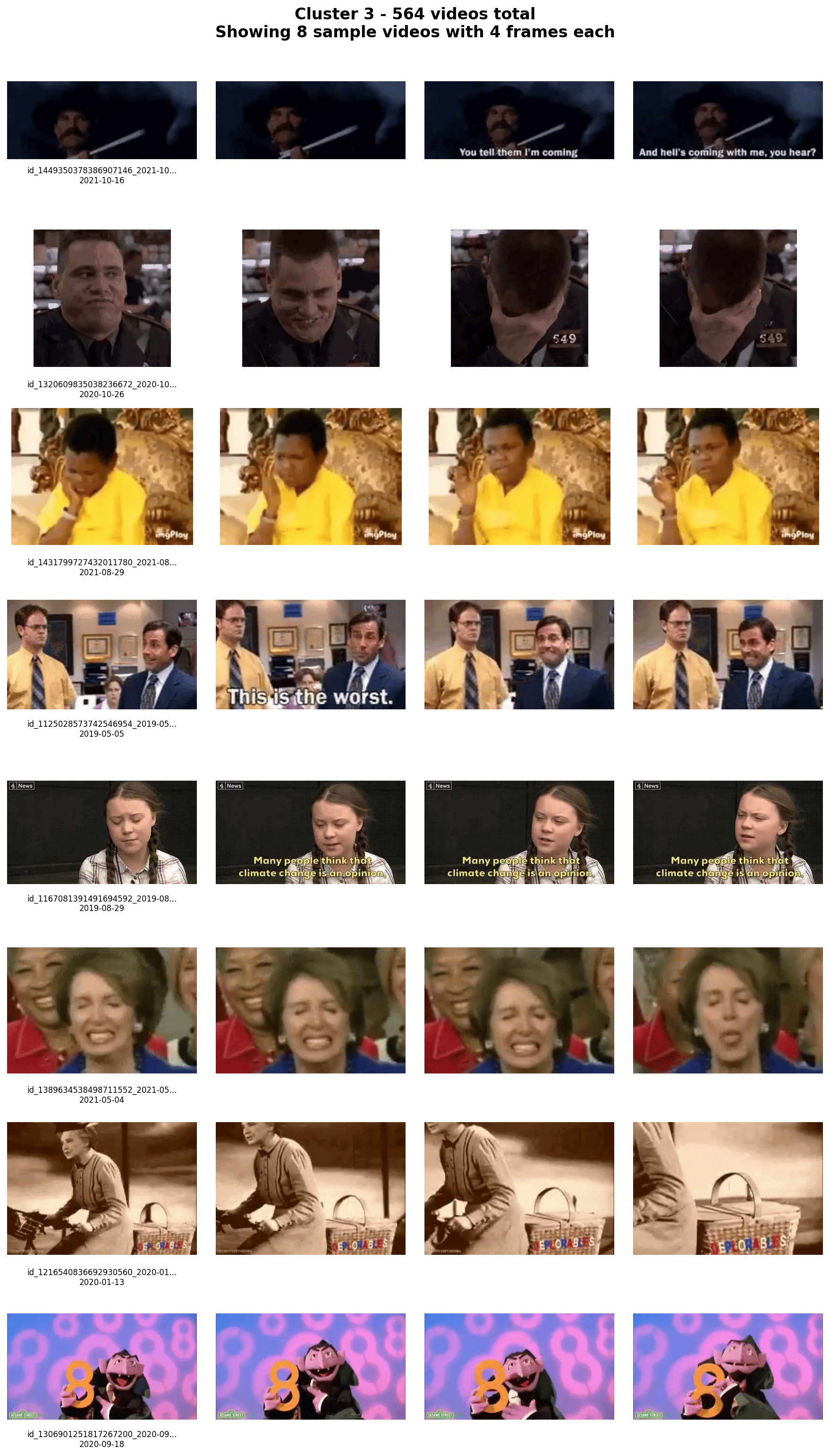}
        \includegraphics[width=0.22\textwidth]{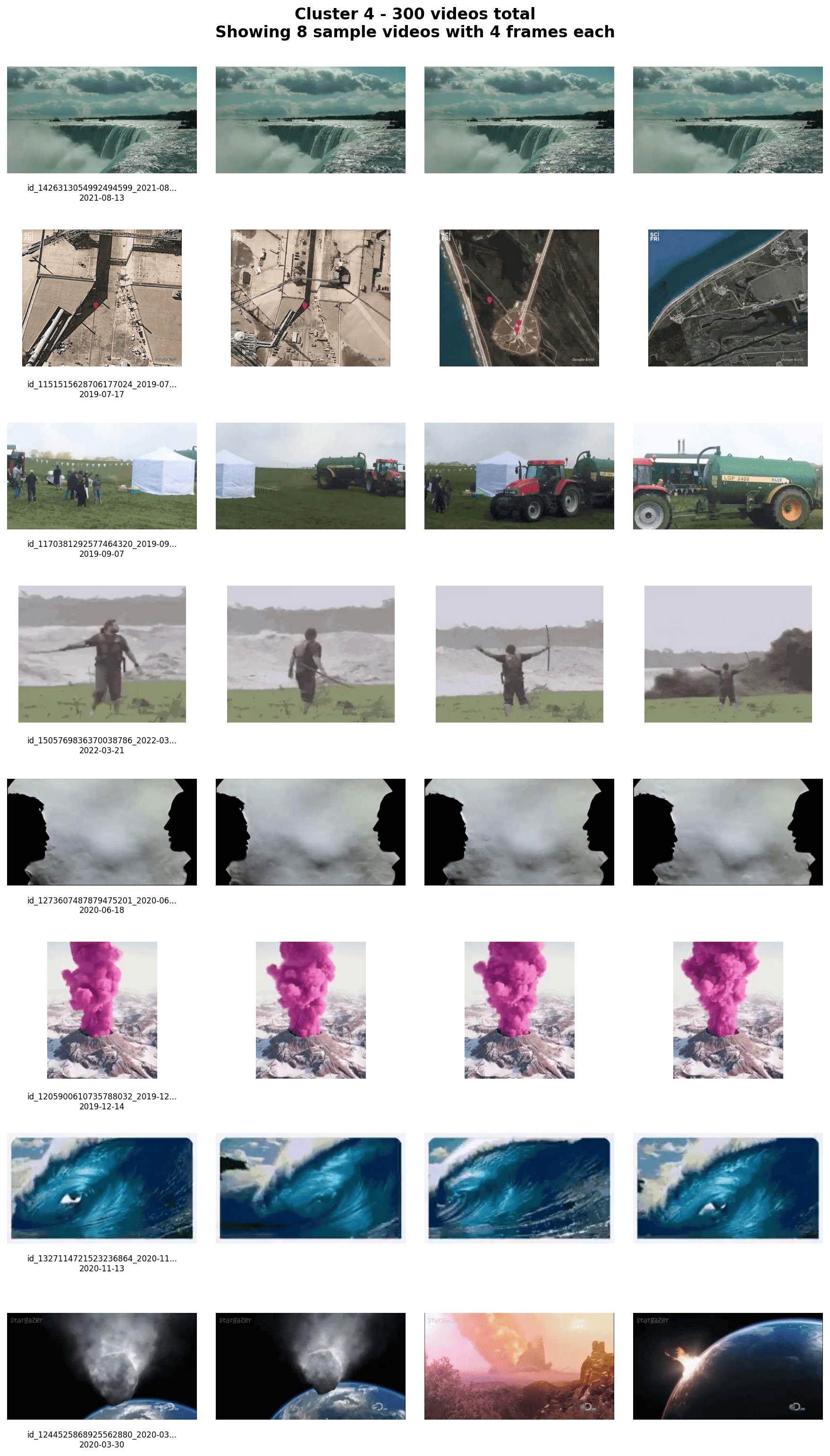}
        \includegraphics[width=0.22\textwidth]{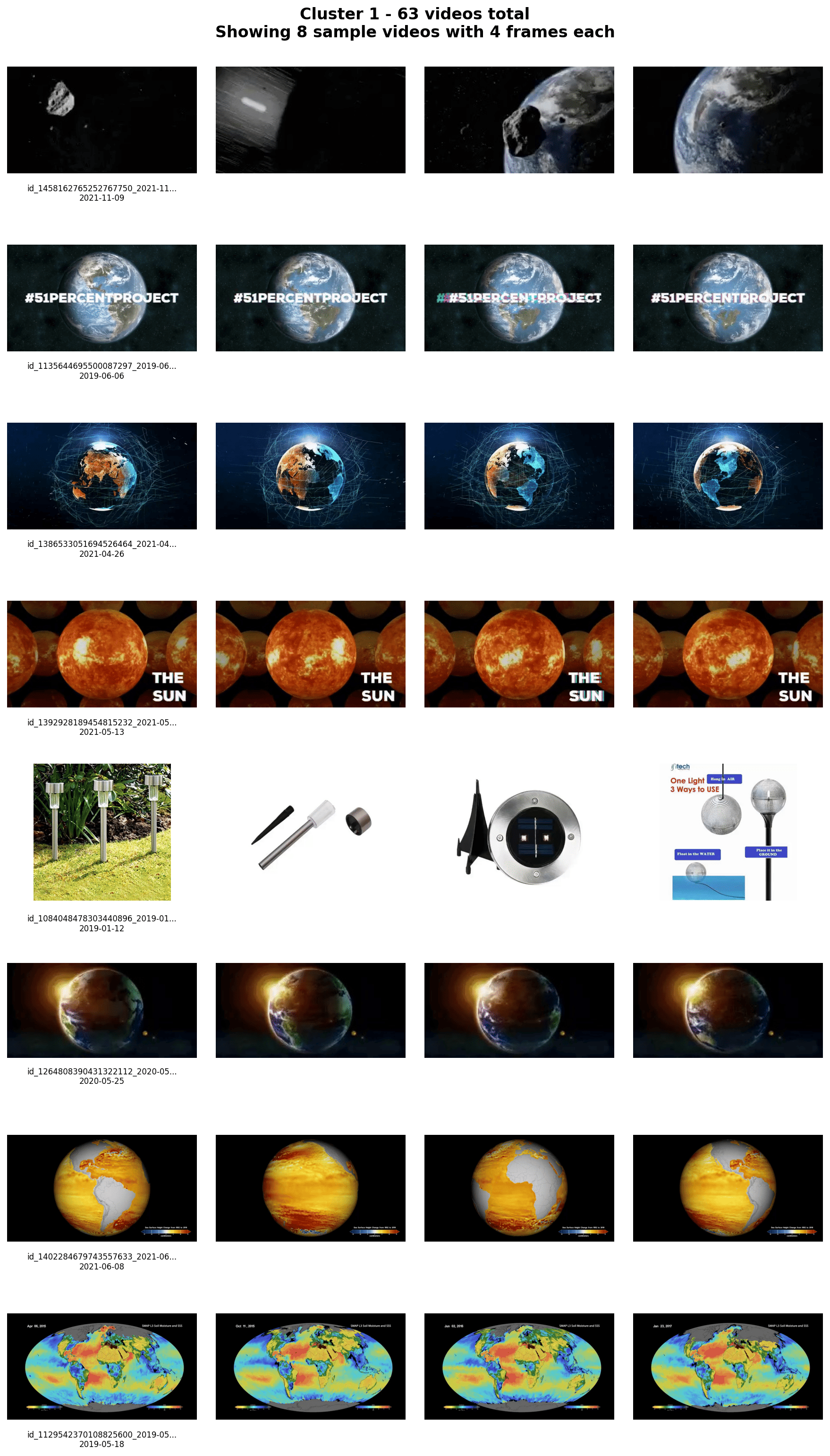}
        \includegraphics[width=0.22\textwidth]{vis/clustering/ConvNeXtV2vs.DINOv2Comparison/visualPatterns/convnextv2/cluster_9_overview.png}
        \includegraphics[width=0.22\textwidth]{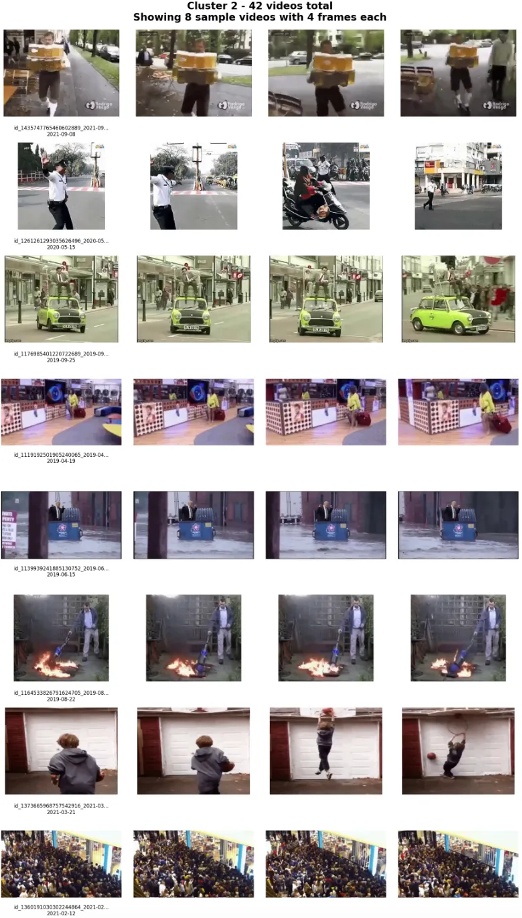}
        \includegraphics[width=0.22\textwidth]{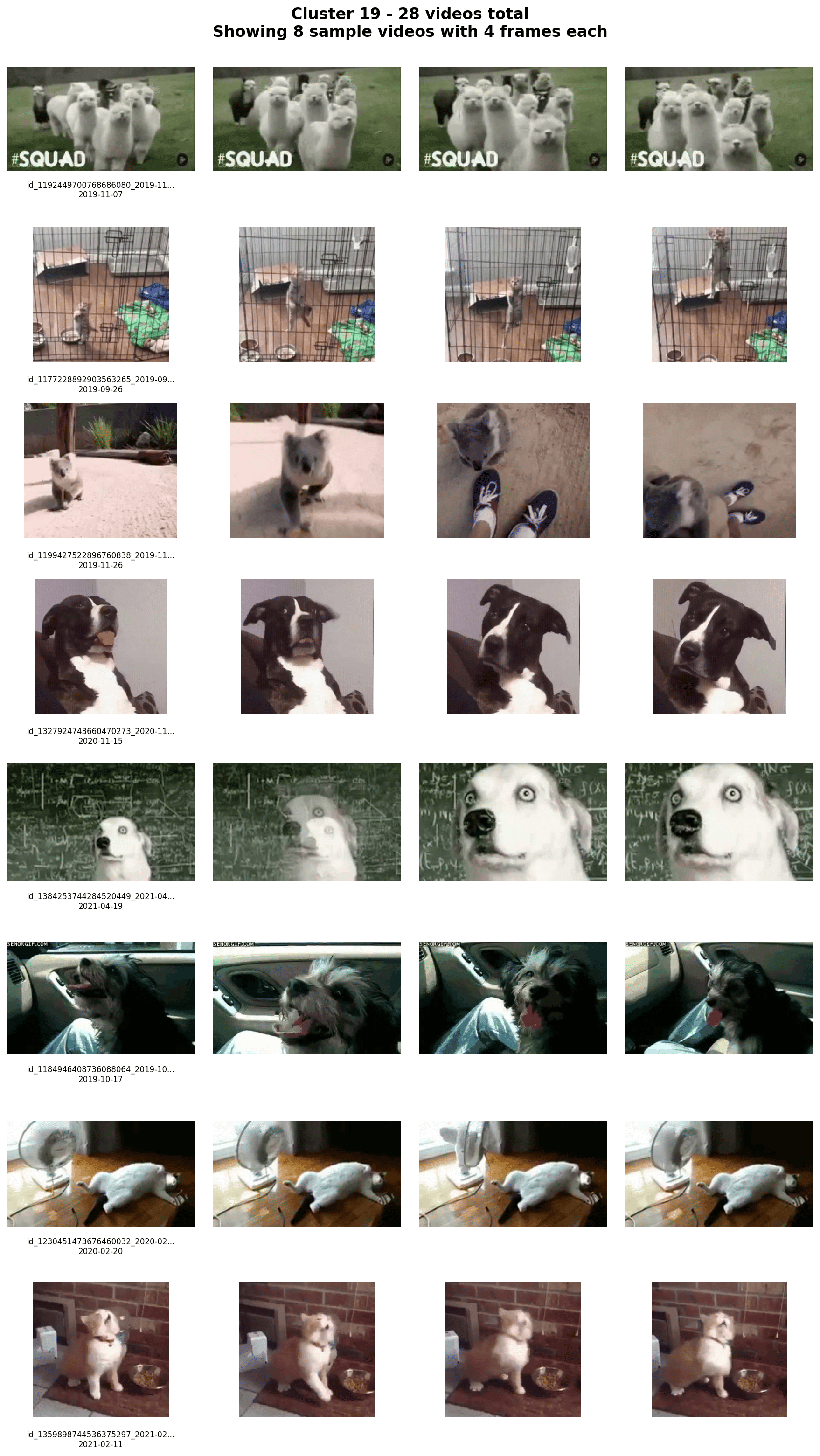}
        \includegraphics[width=0.22\textwidth]{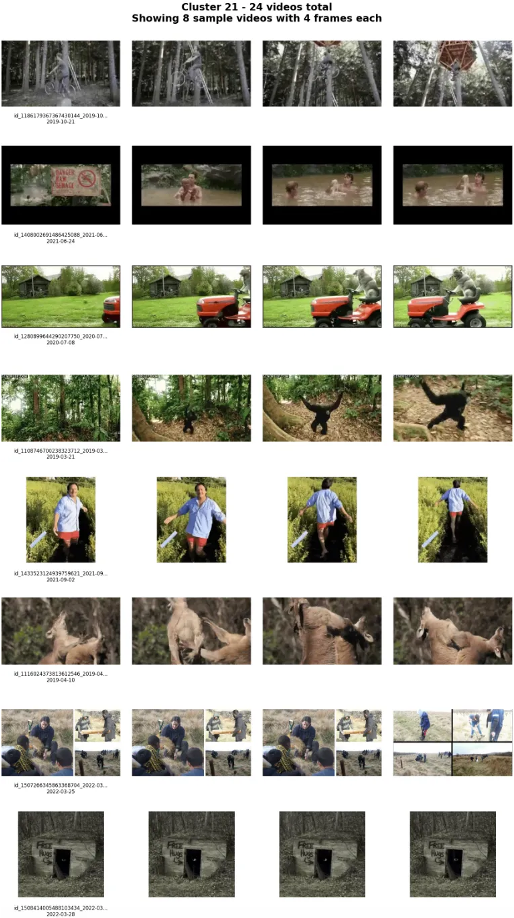}
        \includegraphics[width=0.22\textwidth]{vis/clustering/ConvNeXtV2vs.DINOv2Comparison/visualPatterns/convnextv2/cluster_8_overview.png}
        \includegraphics[width=0.22\textwidth]{vis/clustering/ConvNeXtV2vs.DINOv2Comparison/visualPatterns/convnextv2/cluster_13_overview.png}
        
        \caption{ConvNeXt~V2 top 10 clusters: 
(1) Multi-Panel News/Documentary (1,608 videos, 54.1\%), 
(2) Entertainment Media/Pop Culture (564, 19.0\%), 
(3) Natural Landscapes/Phenomena (300, 10.1\%), 
(4) Space/Cosmic Perspective (63, 2.1\%), 
(5) Digital Animation/Graphics (46, 1.5\%), 
(6) Human Activity/Urban Life (42, 1.4\%), 
(7) Animal Content (28, 0.9\%), 
(8) Rural Environment/Wildlife (24, 0.8\%), 
(9) Scientific Data Visualization (19, 0.6\%), 
(10) Marine Ecosystem (18, 0.6\%).}
\label{fig:qualitative_cluster_comparison_convNext}
\end{figure}

\begin{figure}[htbp]
    \centering
    \includegraphics[width=0.19\textwidth]{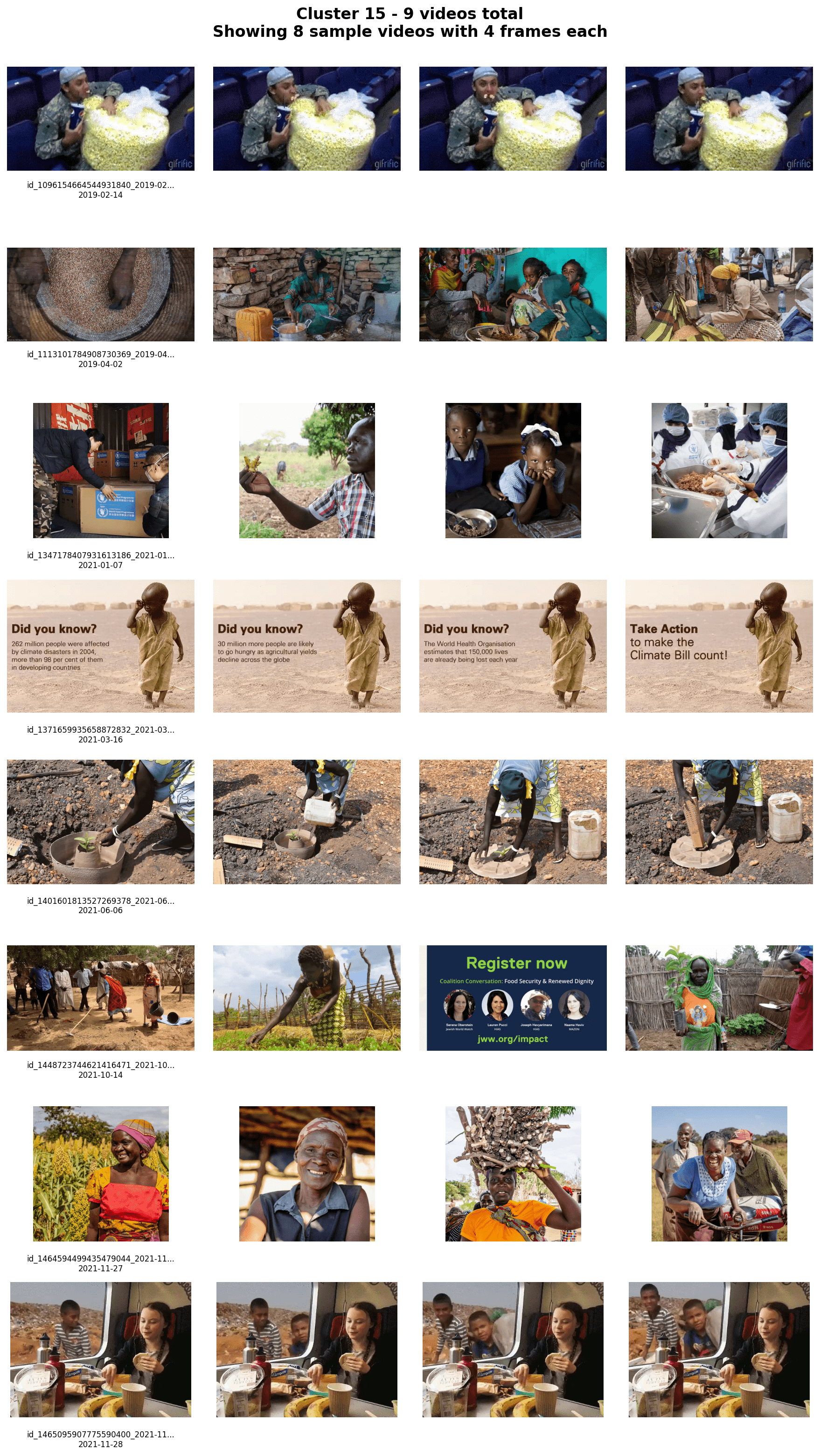}
    \includegraphics[width=0.19\textwidth]{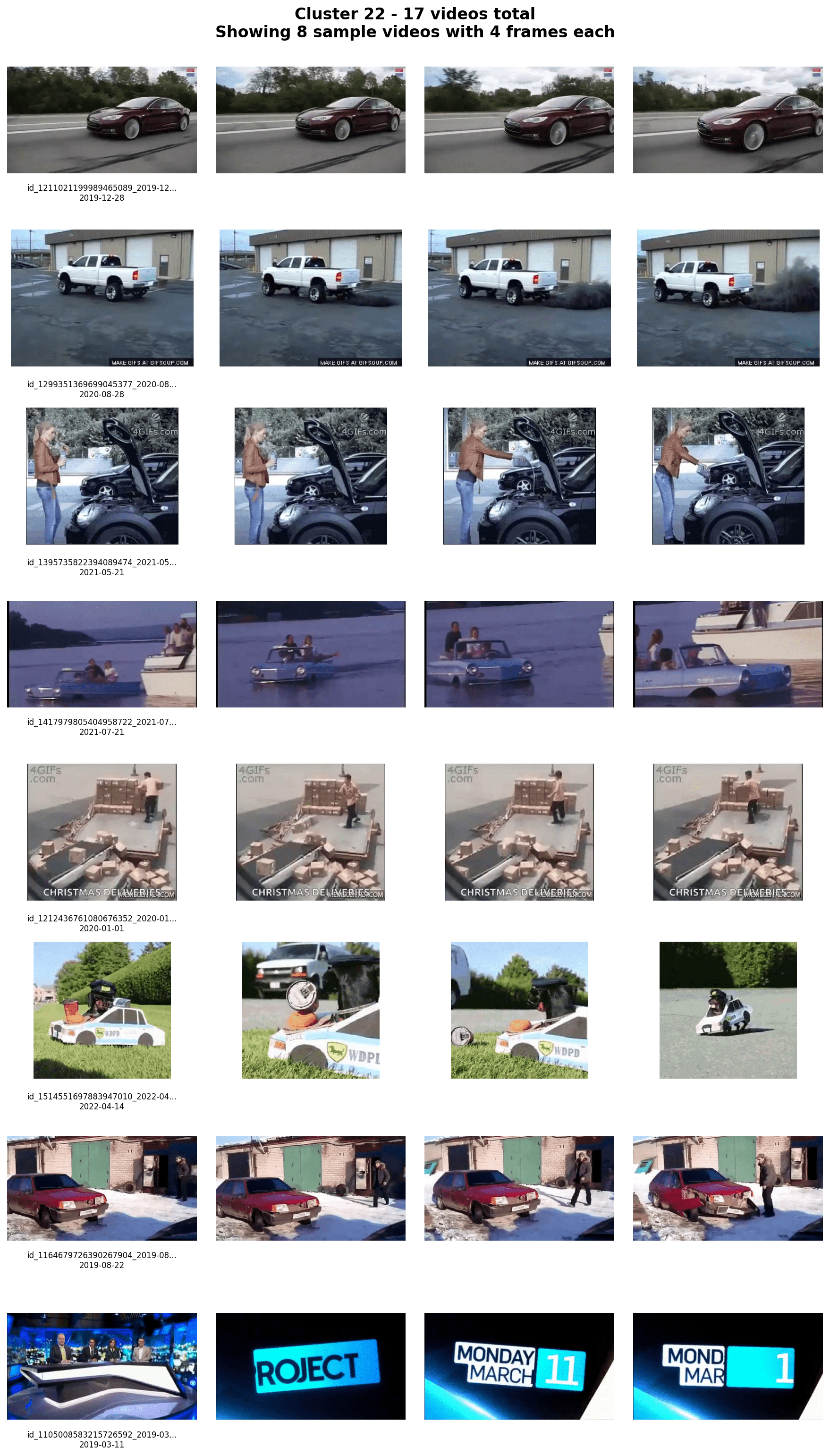}
    \includegraphics[width=0.19\textwidth]{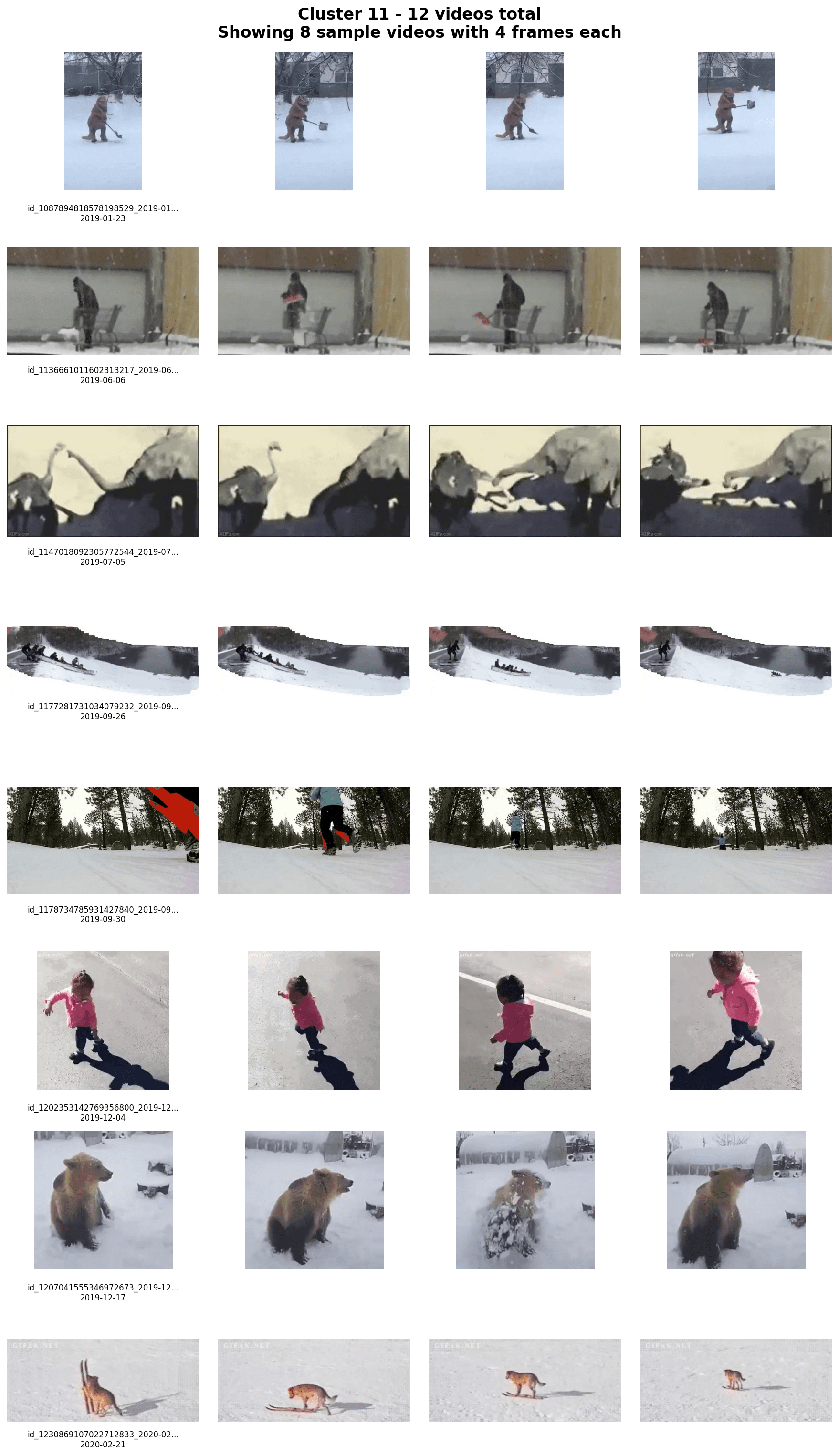}
    \includegraphics[width=0.19\textwidth]{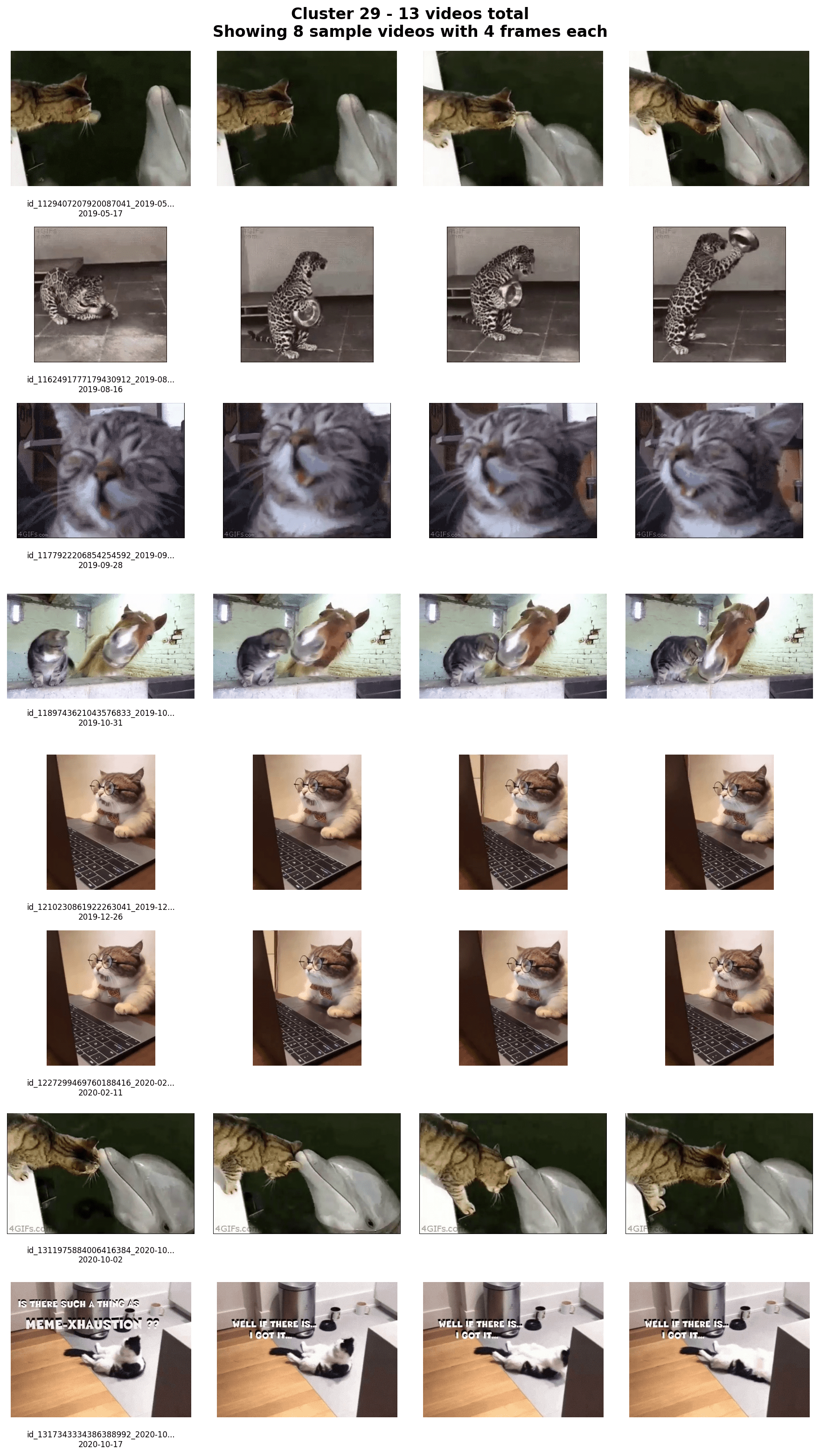}
    \includegraphics[width=0.19\textwidth]{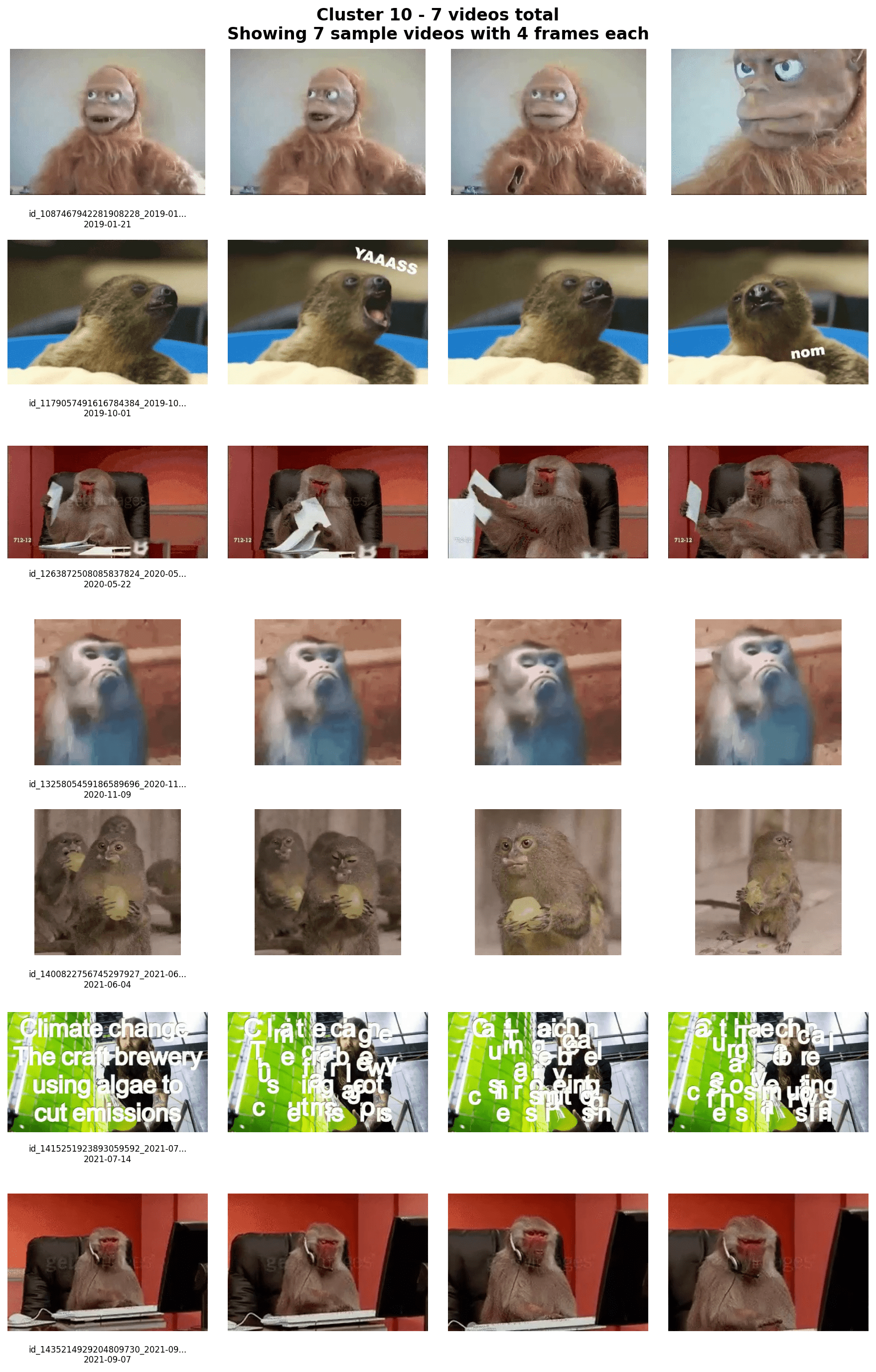}
    \caption{ConvNeXt V2 clusters with animal- or human-related content: 
    (1) Food and humanitarian content with human, 
    (2) Vehicles and TV/media content with human, 
    (3) Human or animals in winter/snow settings, 
    (4) Cats and felines, 
    (5) Primates/monkeys in various settings.}
    \label{fig:convnextv2animalandhuman}
\end{figure}

Fine-grained ConvNeXt V2 clusters are shown in \cref{fig:convnextv2animalandhuman}

\subsection{Cluster Overlap}
\label{app:clusteroverlap}

\begin{figure}[ht]
    \centering
    \includegraphics[width=1\textwidth]{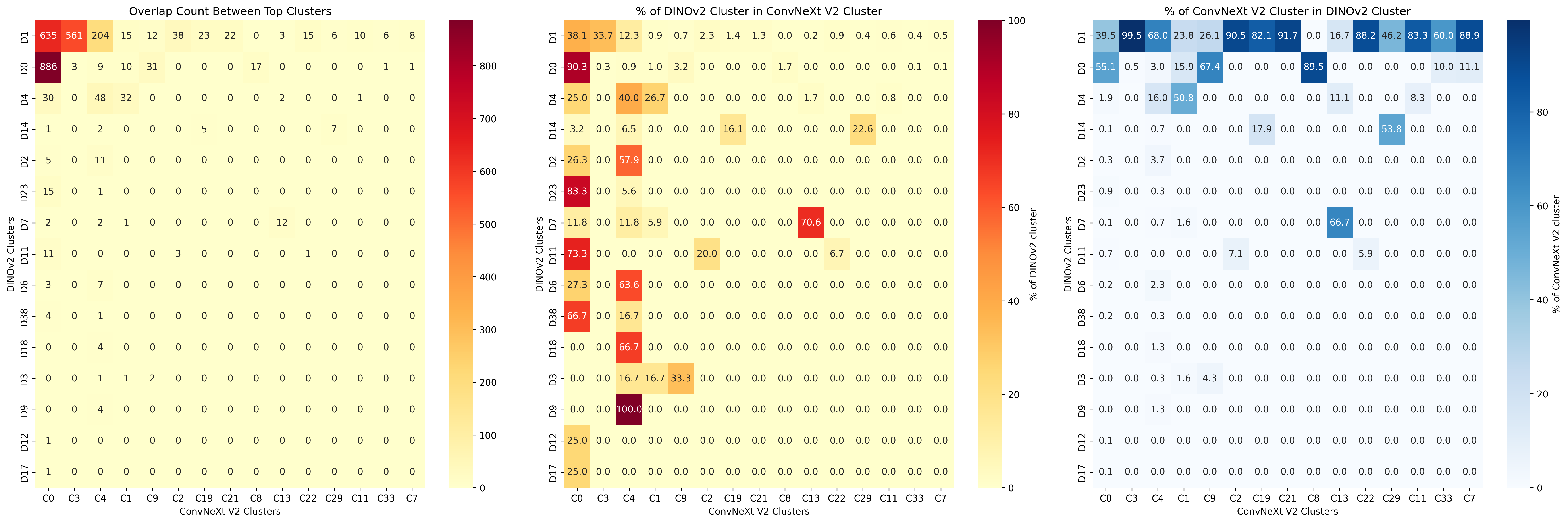}
    \caption{Top 15 Clusters overlap heatmap between DINOv2 and ConvNeXt~V2. 
    Darker cells indicate higher overlap percentages.}
    \label{fig:overlap_heatmap}
\end{figure}

\subsection{Clustering Metrics}
\label{app:metrics}

No configuration achieves well-separated clusters with regard to the evaluation metrics. The low silhouette scores ($<0.1$) could mean that climate videos from the sampled dataset might not naturally form discrete visual categories. Therefore, manual inspection is essential to validate whether the quantitative metrics align with the actual semantic insights from the clusters specific to climate communication patterns.

\begin{table}[htbp]
\centering
\footnotesize
\caption{Clustering performance for DINOv2}
\label{tab:dinov2_clustering_performance}
\resizebox{\textwidth}{!}{%
\begin{tabular}{lllrrrrrrrrrr}
\toprule
Selection & Combination & \#Clusters & Largest & Avg Size & Median & Gini & Top10\% & Singleton\% & Silhouette & DB Score & CH Score & Overall \\
\midrule
static & temporal coherence & 45 & 1647 & 66.0 & 3.0 & 0.922 & 96.90 & 24.44 & 0.077 & 2.056 & 23.37 & 0.523 \\
static & average & 44 & 1668 & 67.5 & 3.0 & 0.923 & 97.07 & 29.55 & 0.081 & 2.096 & 25.30 & 0.519 \\
diverse & temporal coherence & 44 & 1651 & 67.5 & 3.0 & 0.921 & 97.04 & 29.55 & 0.091 & 2.145 & 25.58 & 0.519 \\
static & weighted diversity & 45 & 1666 & 66.0 & 3.0 & 0.924 & 97.14 & 31.11 & 0.081 & 2.083 & 25.37 & 0.518 \\
diverse & average & 50 & 1665 & 59.4 & 2.0 & 0.928 & 97.07 & 34.00 & 0.090 & 2.036 & 24.05 & 0.517 \\
diverse & weighted diversity & 50 & 1665 & 59.4 & 2.0 & 0.928 & 97.07 & 36.00 & 0.089 & 2.057 & 24.75 & 0.514 \\
\bottomrule
\end{tabular}%
}
\end{table}

\begin{table}[htbp]
\centering
\footnotesize
\caption{Clustering performance for ConvNeXt V2}
\label{tab:convnextv2_clustering_performance}
\resizebox{\textwidth}{!}{%
\begin{tabular}{lllrrrrrrrrrr}
\toprule
Selection & Combination & \#Clusters & Largest & Avg Size & Median & Gini & Top10\% & Singleton\% & Silhouette & DB Score & CH Score & Overall \\
\midrule
diverse & average & 71 & 1608 & 41.83 & 4.0 & 0.895 & 91.31 & 14.08 & -0.033 & 2.442 & 8.40 & 0.480 \\
diverse & weighted diversity & 71 & 1607 & 41.83 & 4.0 & 0.894 & 91.25 & 14.08 & -0.033 & 2.462 & 8.37 & 0.479 \\
diverse & weighted confidence & 74 & 1606 & 40.14 & 4.0 & 0.895 & 91.08 & 14.86 & -0.034 & 2.436 & 8.18 & 0.478 \\
static & average & 79 & 1558 & 37.59 & 4.0 & 0.882 & 88.65 & 13.92 & -0.022 & 2.495 & 8.15 & 0.478 \\
static & weighted confidence & 83 & 1556 & 35.78 & 5.0 & 0.882 & 88.35 & 15.66 & -0.015 & 2.466 & 7.99 & 0.478 \\
static & weighted diversity & 84 & 1552 & 35.36 & 4.0 & 0.886 & 88.79 & 17.86 & -0.021 & 2.466 & 8.09 & 0.475 \\
static & max confidence & 178 & 1130 & 16.69 & 4.0 & 0.799 & 70.20 & 12.92 & -0.030 & 2.393 & 4.98 & 0.463 \\
diverse & max confidence & 160 & 1186 & 18.56 & 4.0 & 0.807 & 72.49 & 14.37 & -0.024 & 2.491 & 5.15 & 0.460 \\
\bottomrule
\end{tabular}%
}
\end{table}

\end{document}